\documentclass[sigconf]{acmart}

\AtBeginDocument{%
  }

\copyrightyear{2026}
\acmYear{2026}
\setcopyright{cc}
\setcctype{by}

\acmConference[MM '26]
{Proceedings of the 34th ACM International Conference on Multimedia}
{November 10--14, 2026}
{Rio de Janeiro, Brazil}

\acmBooktitle{Proceedings of the 34th ACM International Conference on Multimedia
(MM '26), November 10--14, 2026, Rio de Janeiro, Brazil}

\acmDOI{10.1145/3767308.3835083}
\acmISBN{979-8-4007-2213-4/2026/11}

\usepackage[utf8]{inputenc} 
\usepackage{nicefrac}      
\usepackage{enumitem}

\usepackage{wrapfig}
\usepackage{subcaption}
\usepackage{multirow}
\usepackage{makecell}

\usepackage{mathtools}
\usepackage{mathrsfs}
\usepackage{algorithm}
\usepackage{algpseudocode}
\usepackage{amsthm}

\usepackage{verbatim}
\usepackage{pifont}
\usepackage{bm}
\usepackage{titletoc}

\usepackage{colortbl}
\usepackage[dvipsnames, svgnames, x11names]{xcolor}
\definecolor{aliceblue}{rgb}{0.94, 0.97, 1.0}
\definecolor{deeppink}{RGB}{255,20,147}
\definecolor{mygray}{gray}{0.9}
\definecolor{myred}{RGB}{255, 0, 0}
\definecolor{myblue}{RGB}{0, 0, 255}   
\definecolor{mygreen}{RGB}{0, 150, 0}

\usepackage{amsmath,amsfonts,bm,xspace}

\def\eqref#1{equation~\ref{#1}}

\def\1{\bm{1}}

\def\mF{{\bm{F}}}

\def\mW{{\bm{W}}}
\def\mX{{\bm{X}}}

\def\mZ{{\bm{Z}}}

\DeclareMathAlphabet{\mathsfit}{\encodingdefault}{\sfdefault}{m}{sl}
\SetMathAlphabet{\mathsfit}{bold}{\encodingdefault}{\sfdefault}{bx}{n}

\newcommand{\eg}{e.g.\xspace}

\begin{document}

\title{SmartMage: Dynamic Modality Orchestration for 3D Scene Understanding}

\author{Yue Zhang}
\affiliation{
  \institution{Zhejiang University}
  \city{Hangzhou}
  \state{Zhejiang}
  \country{China}
}
\email{zhangyue1228@zju.edu.cn}
\orcid{0000-0002-0431-6390}

\author{Yingzhao Jian}
\affiliation{
  \institution{Zhejiang University}
  \city{Hangzhou}
  \state{Zhejiang}
  \country{China}
}
\email{jian.ying.zhao@qq.com}
\orcid{0009-0002-4559-6912}

\author{Yunqiu Xu}
\affiliation{
 \institution{Zhejiang University}
  \city{Hangzhou}
  \state{Zhejiang}
  \country{China}
}
\email{imyunqiuxu@gmail.com}
\orcid{0000-0002-2940-4805}

\author{Xiaoxiao Sun}
\affiliation{
  \institution{Stanford University}
  \city{Stanford}
  \state{California}
  \country{USA}
}
\email{xxsunzrt@gmail.com}
\orcid{0000-0002-6944-7914}

\author{Hehe Fan}
\authornote{Corresponding author.}
\affiliation{
  \institution{Zhejiang University}
  \city{Hangzhou}
  \state{Zhejiang}
  \country{China}
}
\email{hehefan@zju.edu.cn}
\orcid{0000-0001-9572-2345}

\renewcommand{\shortauthors}{Yue Zhang, Yingzhao Jian, Yunqiu Xu, Xiaoxiao Sun, and Hehe Fan}

\begin{abstract}
Understanding 3D scenes is fundamental to embodied intelligence, requiring joint reasoning over heterogeneous information from multiple modalities, including visual and geometric cues. However, the relevance of these modalities often varies across queries. Existing Multimodal Large Language Models (MLLMs) typically rely on fixed modality combinations, overlooking query-dependent modality needs. Such a rigid design can introduce semantic noise from irrelevant modalities while underutilizing more informative ones, leading to wasted computation and diluted reasoning. To address these challenges, this paper proposes SmartMage, a unified MLLM that dynamically orchestrates heterogeneous modalities for semantic-aware 3D scene understanding. Specifically, SmartMage incorporates: (1) a Semantic-guided Modality Adaptive RouTing (SMART) module that selects task-relevant modalities using semantic priors, text–modality alignment, and modality quality; and (2) a Modality-Aware Gating Expert (MAGE) module that leverages modality priors to guide expert activation, fostering adaptive specialization in multimodal reasoning. Empirically, SmartMage achieves state-of-the-art performance across five 3D scene understanding benchmarks, and attains competitive results on RGB-only video understanding benchmarks. In our diagnostic benchmark ScanFacet, tasks are divided into fine-grained semantic categories, enabling analysis of modality combinations preferred by each semantic type. The observed modality–semantic patterns provide further evidence of SmartMage’s effectiveness. Project page: \textcolor{blue}{\url{https://yuecheong.github.io/SmartMage/}}.
\end{abstract}



\begin{teaserfigure}
    \centering
\includegraphics[width=1.0\linewidth]{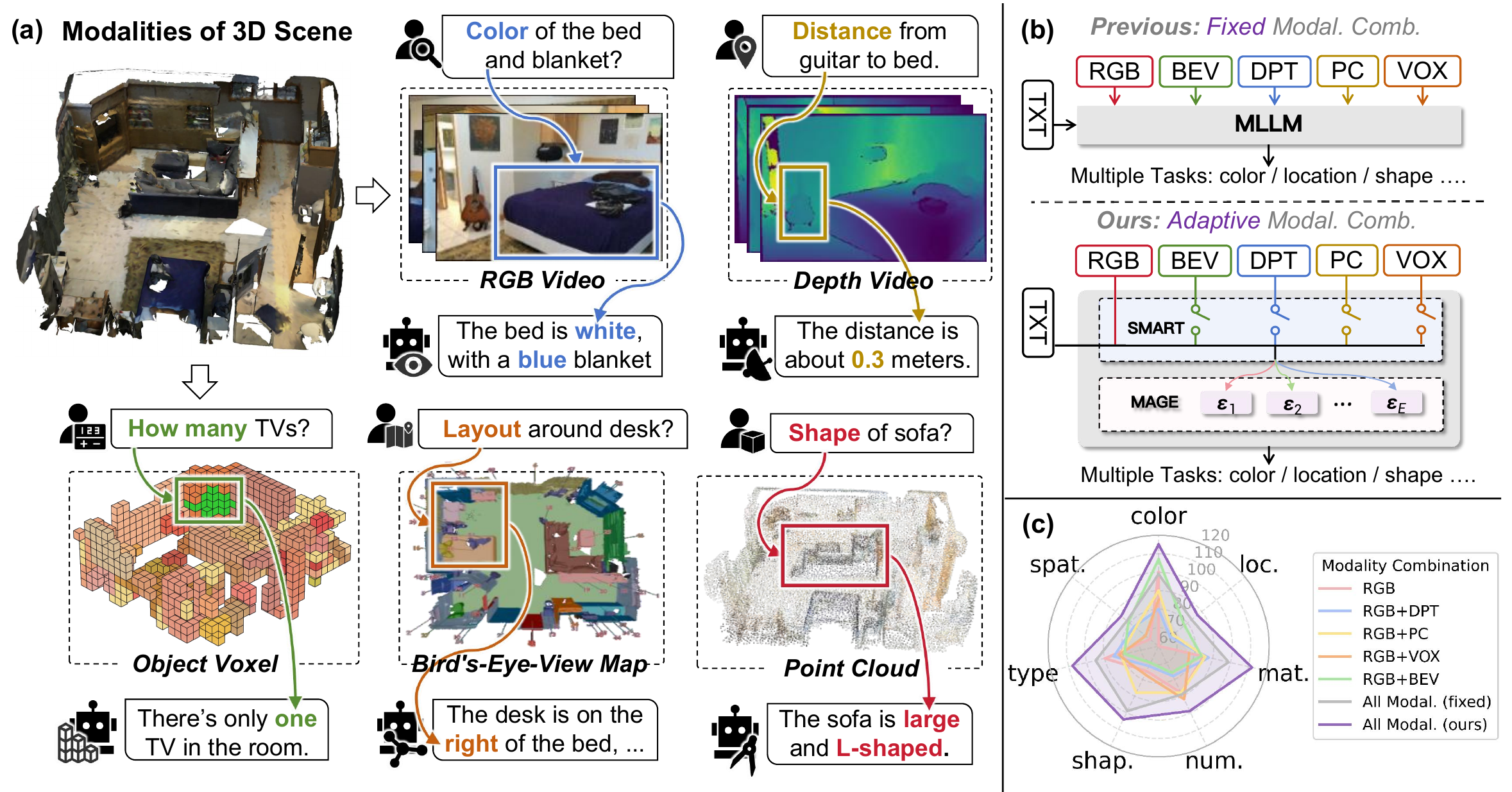}
    \vspace{-1em}
    \caption{Motivation of SmartMage. (a) Different question types exhibit inherent modality preferences. (b) Previous MLLMs fuse a fixed modality set, while SmartMage performs semantic-guided adaptive selection and assigns tokens to modality-specialized experts. (c) Our method outperforms fixed-modality approaches across various semantic categories.} 
    \label{fig:teaser}
\end{teaserfigure}

\maketitle

\section{Introduction}
Understanding 3D scenes forms the perceptual and reasoning backbone of embodied intelligence~\citep{werby2024hierarchical,yin2024sg,gu2024conceptgraphs}, empowering agents to navigate, manipulate, and interact purposefully within complex indoor environments~\cite{zha2025enable,huang2026wildtablebench,xu2024comp4d,yang2024llm}.
Real-world 3D scenes are captured by multiple complementary modalities, including RGB-D video~\cite{zhang2025prompt, zhang2025samcontrol}, Point Cloud (PC)~\cite{fan2022point}, voxel, and Bird's-Eye-View (BEV) projections, which jointly encode appearance, geometry, structure, and spatial layout. 
Recent multimodal approaches~\citep{zhang2025uni3d, chen2026cogflow, fu2024scene, xu2024gg} have therefore begun to integrate multiple 3D modalities~\cite{zhou2025its3d} in pursuit of a more complete scene perception~\citep{hou2026seeing}. Despite incorporating more modalities, existing approaches~\cite{zhu2024unifying, huang2024chat, zhang2025uni3d} typically rely on fixed modality combinations, implicitly treating all modalities as equally useful for every query.

However, such designs overlook the fact that \textbf{different questions favor distinct modalities and sensory cues, with varying importance across tasks.}
As illustrated in Fig.~\ref{fig:teaser}(a), geometric questions (\textit{e.g.}, \textit{``What shape is the sofa?''}) benefit more from point cloud representations, whereas appearance-related questions (\textit{e.g.}, \textit{``What color is the blanket?''}) primarily rely on RGB inputs.
When all modalities are indiscriminately fed into the model, as depicted in Fig.~\ref{fig:teaser}(b), irrelevant inputs may introduce semantic noise, while informative ones fail to receive sufficient attention~\citep{wu2025mitigating, wei2025improving, cai2025diagnosing}, leading to wasted computation and diluted reasoning.
Consequently, as shown in Fig.~\ref{fig:teaser}(c), simply stacking more modalities under a fixed fusion scheme can be counter-productive, potentially leading to performance degradation in 3D scene understanding.

To address the limitation of fixed multimodal fusion, we propose \textbf{SmartMage}, a unified model that dynamically orchestrates heterogeneous modalities for 3D scene understanding. 
The key idea is to decompose multimodal reasoning into two steps: (1) \textbf{selecting relevant modalities} and (2) \textbf{assigning modality tokens to specialized experts for processing}.
As shown in Fig.~\ref{fig:teaser}(b), SmartMage realizes these two steps through the Semantic-guided Modality Adaptive RouTing (\textbf{SMART}) module and the Modality-Aware Gating Expert (\textbf{MAGE}) module, respectively.

Specifically, the SMART module serves as the global modality scheduler, determining which modalities should participate in reasoning for each query.
It reaches this decision by jointly considering three cues:
a semantic prior estimator that infers the query’s expected modality usage,
a semantic similarity scorer that measures text–modality semantic consistency,
and a modality quality evaluator that checks the completeness and reliability of each modality.
Combining these signals yields a relevance distribution that retains RGB as the fixed primary modality while adaptively selecting a semantic-dependent relevant subset of complementary modalities.

Given the selected modalities, we focus on how to utilize them effectively during reasoning.
Accordingly, we introduce MAGE, a modality-aware expert allocation mechanism that assigns modality tokens to appropriate experts within the LLM.
It comprises:
(1) a modality-aware expert speculation module that predicts token-level modality attribution and provides a soft modality–expert affinity prior;
and (2) a sparse Mixture-of-Experts (MoE)-based LLM that incorporates this prior into its gating function, biasing token-level routing and promoting expert specialization across modalities.

We evaluate SmartMage across three complementary settings: standard 3D scene understanding benchmarks~\citep{azuma2022scanqa, chen2021scan2cap, ma2022sqa3d, chen2020scanrefer, zhang2023multi3drefer}, RGB-only video understanding benchmarks~\citep{yang2025thinking, yang2025cambrian, yang2025mmsi}, and a newly proposed diagnostic benchmark.
SmartMage achieves State-Of-The-Art (SOTA) performance on five 3D benchmarks, surpassing Ross3D~\citep{wang2025ross3d} by +5.1 Acc@0.5 on ScanRefer~\citep{chen2020scanrefer} and +6.4 F1@0.5 on Multi3DRefer~\citep{zhang2023multi3drefer}.
On three RGB-only video benchmarks, it remains highly competitive despite the absence of 3D inputs, demonstrating robustness to missing modalities.
Furthermore, existing benchmarks mix diverse question types and obscure modality-specific behaviors.
We therefore introduce ScanFacet, a diagnostic benchmark that organizes questions into eight semantic facets, enabling fine-grained analysis of modality usage.
SmartMage shows consistent improvements across all facets, with particularly large gains of +27.1 and +15.9 CIDEr on material and color understanding.
Main contributions of this paper can be summarized as follows:

\begin{itemize}[leftmargin=0pt]
    \item 
    \textbf{Unified 3D scene understanding MLLM.}
    We introduce SmartMage, a unified framework that integrates heterogeneous 3D modalities into a single multimodal reasoning pipeline. SmartMage performs global-to-local semantic adaptation, bridging high-level semantic intent and low-level multimodal fusion within a cohesive architecture.
    \item  
    \textbf{Semantic-guided global modality routing.}
    We propose SMART, a global modality scheduler that dynamically selects informative modalities based on query semantics. It transforms multimodal fusion from static combination into semantic-aware, context-adaptive routing for efficient and interpretable reasoning.
    \item 
    \textbf{Modality-aware local expert specialization.}
    We introduce MAGE, a modality-aware expert allocation mechanism that leverages modality priors to guide expert activation, fostering adaptive specialization in multimodal reasoning.
\end{itemize}

\begin{figure*}[t!]
    \centering
\includegraphics[width=1\linewidth]{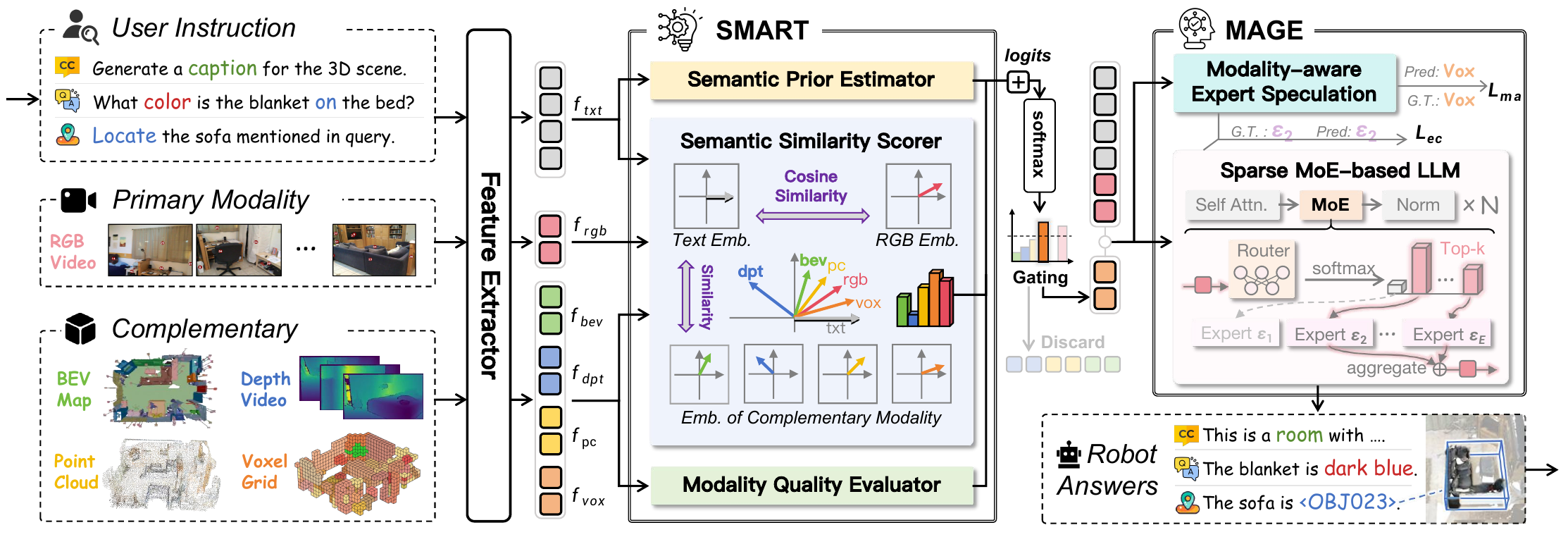}
\vspace{-2.2em}
    \caption{Overview of SmartMage. The omni-modal feature extractor first encodes text, RGB, depth, BEV, point cloud, and voxel inputs into a unified embedding space. The SMART module serves as a semantics-aware modality scheduler. It evaluates semantic priors, semantic similarity, and modality quality to choose an instruction-dependent set of auxiliary modalities. The selected modalities are combined with the primary RGB and text features and passed into the MAGE module. Within MAGE, the modality-aware expert speculation component injects modality cues into the sparse MoE-based LLM, promoting expert specialization in handling modality-specific information.
    }
    \label{fig:overview}
\vspace{-1.2em}
\end{figure*}

\section{Related Work}
\noindent \textit{\textbf{3D Scene Understanding with Multimodal Inputs.}}
Early works mainly use point clouds~\citep{xu2024pointllm, guo2023point, liu2024uni3d, cao2025objvariantensemble, qi2024gpt4point} or voxels~\citep{fu2024scene, zhu2024unifying, yang2025lidar}, enabling 3D grounding~\citep{yang2024llm, chen2024grounded, ahmed2025kestrel, wang2025liba, xu2025mc} and question answering~\citep{zhu20254d, szymanska2024space3d, li2025embodied} with LLMs. Later, inspired by strong 2D encoders, methods~\citep{zhu2025llava, yang2025lidar} project 2D features into 3D space to improve fine-grained understanding. 
Recently, 2D video-based methods~\citep{fu2024scene, xiong20253ur, qi2025gpt4scene, zheng2025video} achieve competitive performance compared with methods using explicit 3D representations, showing that spatial reasoning can emerge from temporal visual inputs~\citep{yang2025thinking, yang2025cambrian, yang2025mmsi}.
However, existing methods typically rely on fixed modality combinations, overlooking the task-dependent nature of modality contributions.

\noindent\textit{\textbf{Expert-based Multimodal Modeling.}}
The MoE framework offers a modular paradigm for multimodal reasoning by allocating computation to specialized experts~\citep{zhou2022mixture, cai2024survey, du2024revisiting, song2024promoe, xue2024openmoe}.
Routing strategies in MoE are typically divided into hard and soft forms.
Hard routing~\citep{bao2022vlmo, shen2023scaling} deterministically assigns tokens to fixed experts~\citep{wang2023chat, wang2023image}, providing separation but lacking token-level and cross-modal adaptability.
Soft routing~\citep{yue2024ada, liu2026clear, jia2024mos2, lin2026moe, li2025uni} learns dynamic token-to-expert mappings, offering greater flexibility yet often suffering from unguided expert behaviors and weak interpretability. 
Existing MoE routing lacks modality-aware signals, resulting in unstructured expert allocation and suboptimal multimodal reasoning.

\section{SmartMage}
SmartMage achieves effective 3D scene understanding through global-to-local adaptive modality selection and processing.
The pipeline is depicted in Fig.~\ref{fig:overview}.
An omni-modal 3D feature extractor handles heterogeneous 3D inputs (\textit{e.g.}, RGB, depth, BEV, PC and voxel representations) and projects them into a unified embedding space (\S\ref{sec:extractor}).
Then, SMART performs semantic-guided global modality selection (\S\ref{sec:smart}), and MAGE enables modality-aware local expert specialization within the LLM (\S\ref{sec:mage}).

\subsection{Omni-modal 3D Scene Feature Extractor}\label{sec:extractor}
SmartMage exploits RGB-D video, BEV, PC and voxel to construct 3D scene representations.
For RGB-D video, we introduce a fast keyframe selection algorithm, \texttt{FoVSR}~\citep{zheng2025video}, which selects spatially diverse frames by estimating voxel coverage directly from camera poses, achieving over 100× speedup compared with depth-based methods (see Appendix for details).
To supply global context beyond egocentric views, we render semantic BEV maps from the scene mesh with orthographic projection.
RGB-D and BEV are encoded by a vision encoder~\citep{bai2025qwen3} with an adapter, yielding $\mathbf{f}_{\text{rgb}}$, $\mathbf{f}_{\text{dpt}}$, $\mathbf{f}_{\text{bev}}$. 
The point cloud branch applies farthest point sampling and \texttt{PointNet++}~\cite{qi2017pointnet++} to obtain $\mathbf{f}_{\text{pc}}$, while the voxel branch voxelizes the scene and uses \texttt{Mask3D}’s~\cite{schult2023mask3d} sparse U-Net with object-level aggregation to produce $\mathbf{f}_{\text{vox}}$. 
Finally, with modality set $\mathcal{M}=\{\mathrm{rgb},\mathrm{dpt},\mathrm{bev},\mathrm{pc},\mathrm{vox}\}$, we collect features $\{\mathbf{f}_m\in\mathbb{R}^{N_m\times d_v}\}_{m\in\mathcal{M}}$, where $N_m$ is the token count for modality $m$ and $d_v$ is the visual embedding dimension.

\subsection{Semantic-guided Modality Adaptive Routing}\label{sec:smart}
We propose SMART, a modality router that selects the primary RGB modality and adaptively supplements it with query-relevant complementary modalities.
Given the instruction embedding $\mathbf{f}_{\text{txt}}$ and multimodal visual features $\{\mathbf{f}_m\}_{m\in\mathcal{M}}$, SMART predicts a modality-level routing distribution.
SMART integrates three complementary signals: the \emph{Semantic Prior Estimator} (SPE) infers modality preference from instruction semantics, the \emph{Semantic Similarity Scorer} (SSS) evaluates text–visual similarity, and the \emph{Modality Quality Evaluator} (MQE) estimates visual reliability.  
Overall, SMART defines a unified paradigm for instruction-conditioned multimodal routing, achieving dynamic and robust modality coordination.

\noindent\textit{\textbf{Semantic Prior Estimator.}}
Textual instructions naturally reveal modality preference.
For example, \textit{``What color is the blanket?''} primarily relies on RGB cues, while \textit{``Where is the sofa located?''} depends more on geometric modalities such as BEV or voxel.
The SPE module captures this instruction-conditioned prior through a lightweight prediction head applied to the text embedding:
\begin{equation}\label{eq:txt_prior}
\mathbf{p} = \operatorname{Softmax}\big(\mW_p\operatorname{LN}(\mathbf{f}_{\text{txt}})\big),
\end{equation}
where LN($\cdot$) is layer normalization applied to the text embedding $\mathbf{f}_{\text{txt}}$, and $\mW_p$ projects it into the modality space.
Overall, SPE predicts an instruction-driven modality relevance distribution that serves as a prior for modality selection.

\noindent
\textit{\textbf{Semantic Similarity Scorer.}}
To estimate the relevance of each modality to the given instruction, the SSS module constructs text-guided modality summaries and measures their semantic alignment with the textual intent.
A lightweight text summarizer decomposes the text embedding $\mathbf{f}_{\text{txt}}$ into two complementary representations: (i) a global instruction embedding $\mathbf{g}_{\text{txt}}$ that captures the overall semantic intent, and (ii) a set of modality-specific query vectors $\{\boldsymbol{\phi}_m\}_{m=1}^M$ that guide attention over the corresponding visual modalities.
Each modality feature $\mathbf{f}_m$ interacts with its query $\boldsymbol{\phi}_m$ through cross-attention, yielding a text-conditioned representation $\hat{\mathbf{f}}_m = \text{CrossAttn}(\boldsymbol{\phi}_m, \mathbf{f}_m)$ that selectively aggregates instruction-relevant evidence.
We then project $\mathbf{g}_{\text{txt}}$ and $\{\hat{\mathbf{f}}_m\}_{m=1}^M$ into a shared semantic space and compute their similarity as:
\begin{equation}
s_m = \gamma_m \cdot 
\mathrm{Norm}\bigl(\langle \mW_m \hat{\mathbf{f}}_m,\; \mW_t \mathbf{g}_{\text{txt}} \rangle\bigr)
+ \beta_m,
\end{equation}
where $\mW_m$ and $\mW_t$ are learnable projection matrices, $\langle \cdot,\cdot\rangle$ denotes cosine similarity, and $\mathrm{Norm}(\cdot)$ performs per-sample normalization across modalities. The parameters $\gamma_m$ and $\beta_m$ are learnable modality-specific scale and bias terms for affine calibration.
Collecting all $s_m$ yields $\mathbf{s} \in \mathbb{R}^M$, a modality-wise similarity vector.
Overall, the SSS module extracts instruction-relevant representations from each visual modality and evaluates their alignment with the textual intent to produce modality relevance scores.

\begin{figure}[t!]
    \centering
    \includegraphics[width=1.0\linewidth]{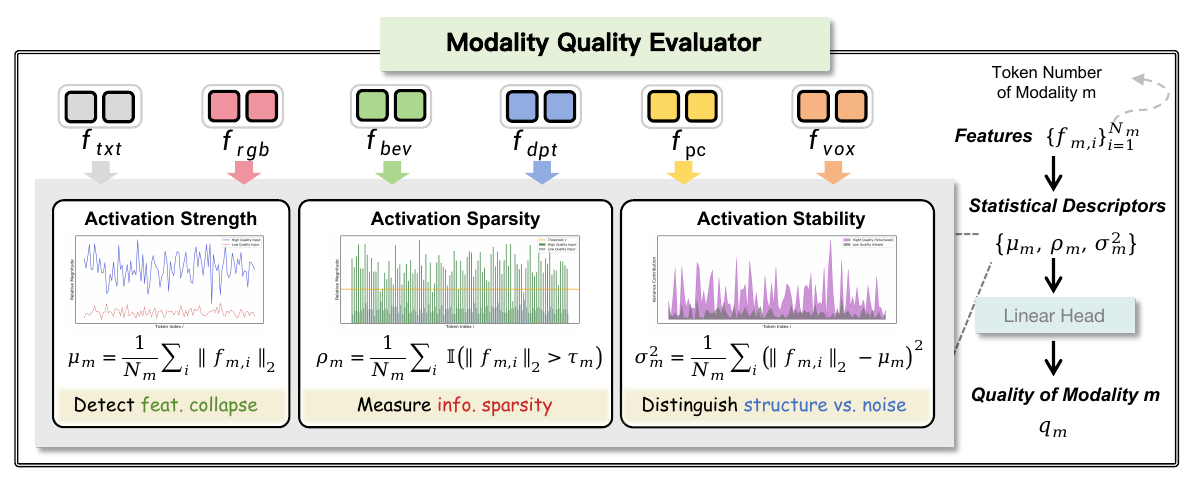}
    \vspace{-2em}
    \caption{Details of the Modality Quality Evaluator. We model modality quality using activation statistics that capture strength, sparsity, and stability, and convert them into a reliability score for routing.}
    \label{fig:mqe}
    \vspace{-1em}
\end{figure}

\noindent
\textit{\textbf{Modality Quality Evaluator.}}
As shown in Fig.~\ref{fig:mqe}, the MQE module estimates a quality score for each modality from its feature activations.
For modality $m$, we derive activation statistics from feature norms, including the activation strength $\mu_m$, sparsity $\rho_m$, and stability $\sigma_m^2$, whose detailed formulations are given in the Appendix.
These statistics are concatenated into a descriptor and mapped to a scalar quality score by a lightweight linear head:
\begin{equation}
q_m = \mathbf{w}_q^\top [\mu_m,\rho_m,\sigma_m^2] + b_q,
\end{equation}
where $\mathbf{w}_q$ and $b_q$ are learnable parameters.
Collecting all modality scores yields a quality vector $\mathbf{q} \in \mathbb{R}^M$.
This design allows routing to down-weight degraded modalities and favor more reliable ones.

\noindent\textit{\textbf{Modality Adaptive Routing.}}
After obtaining the modality-wise prior $\mathbf{p}$, semantic similarity $\mathbf{s}$, and quality score $\mathbf{q}$, the router integrates them into unified routing logits:
\begin{equation}\label{eq:logits}
  \mathbf{z} = \alpha_p\,\mathbf{p} + \alpha_s\,\mathbf{s} + \alpha_q\,\mathbf{q} + \mathbf{b}, 
\end{equation}
where $\alpha_s$, $\alpha_p$, $\alpha_q$ weight each cue, and $\mathbf{b}$ is a learnable bias.
To further adapt routing, we introduce an \emph{RGB evidence gate} that modulates $\mathbf{z}$ based on text–RGB similarity: it strengthens RGB when the views provide clear semantic evidence and suppresses it when instruction-relevant content is absent from the selected views.
Overall, the SMART module jointly captures (1) \emph{what the instruction demands}, (2) \emph{how well each modality aligns}, and (3) \emph{how reliable each modality is}, enabling dynamic and interpretable modality selection.

\subsection{Modality-Aware Gating Expert}\label{sec:mage}
While SMART performs input-level modality selection, the decoder still processes multimodal tokens uniformly, lacking modality-specific inductive bias.
To address this, we introduce MAGE, which aligns expert routing with modality-aware representations for structured multimodal cooperation.

\noindent\textit{\textbf{Soft Routing for Expert Selection.}}
The MoE module employs a learnable soft routing mechanism to achieve adaptive token-to-expert assignment.
For each token feature $\mathbf{h}_i^{(\ell)}$ (token $i$ at layer $\ell$), the routing network computes a probability distribution $\pi_{i,e}^{(\ell)}$ over $E$ experts as:
\begin{equation}
\pi_{i,e}^{(\ell)}=
\frac{\exp\!\left(\mathbf{w}_e^{(\ell)\top}\mathbf{h}_i^{(\ell)} / \tau\right)}
{\sum_{j=1}^{E}\exp\!\left(\mathbf{w}_j^{(\ell)\top}\mathbf{h}_i^{(\ell)} / \tau\right)},
\end{equation}
where $\mathbf{w}_e^{(\ell)}$ is the gating weight of expert $e$ at layer $\ell$ and $\tau$ is the gating temperature. 
Each token $\mathbf{h}_i^{(\ell)}$ is then routed to its top-$k$ experts and aggregated as:
\begin{equation}
\hat{\mathbf{h}}_i^{(\ell)}=
\sum_{e\in \operatorname{TopK}(\boldsymbol{\pi}_i^{(\ell)})}
\pi_{i,e}^{(\ell)}\,\mathcal{E}_e\!\left(\mathbf{h}_i^{(\ell)}\right),
\end{equation}
where $\boldsymbol{\pi}_i^{(\ell)}=[\pi_{i,1}^{(\ell)},\dots,\pi_{i,E}^{(\ell)}]$, $\operatorname{TopK}(\cdot)$ returns the indices of the top-$k$ experts, and $\mathcal{E}_e(\cdot)$ denotes the transformation implemented by expert $e$.

\noindent\textit{\textbf{Modality-aware Expert Specialization.}}
Given a token $\mathbf{h}_i^{(\ell)}$, a Modality-aware Expert Speculation (MES) module predicts a modality distribution $\mathbf{r}_i = [r_{i,1}, \dots, r_{i,M}] \in \Delta^{M-1}$ over $M$ modalities, where $r_{i,m}$ indicates the probability that token $i$ is associated with modality $m$.
To model modality–expert relationships, we introduce a learnable affinity matrix $\mathbf{A}$, where each entry $a_{m,e}$ measures the compatibility between modality $m$ and expert $e$.
We then derive a modality-aware expert prior for each token:
\begin{equation}
\tilde{\pi}_{i,e} = \sum_{m=1}^{M} r_{i,m} \cdot 
\frac{\exp(a_{m,e}/\tau)}{\sum_{j=1}^{E}\exp(a_{m,j}/\tau)}.
\end{equation}

During training, we regularize the routing distribution $\pi_{i,e}^{(\ell)}$ toward the modality-aware prior $\tilde{\pi}_{i,e}$ via the expert calibration loss $\mathcal{L}_{\text{ec}}$, encouraging modality-consistent expert specialization (detailed in \S\ref{sec:obj}).
Through such adaptive routing, MAGE realizes fine-grained, modality-consistent fusion within each MoE layer.

\begin{figure}[t!]
    \centering
\includegraphics[width=1.0\linewidth]{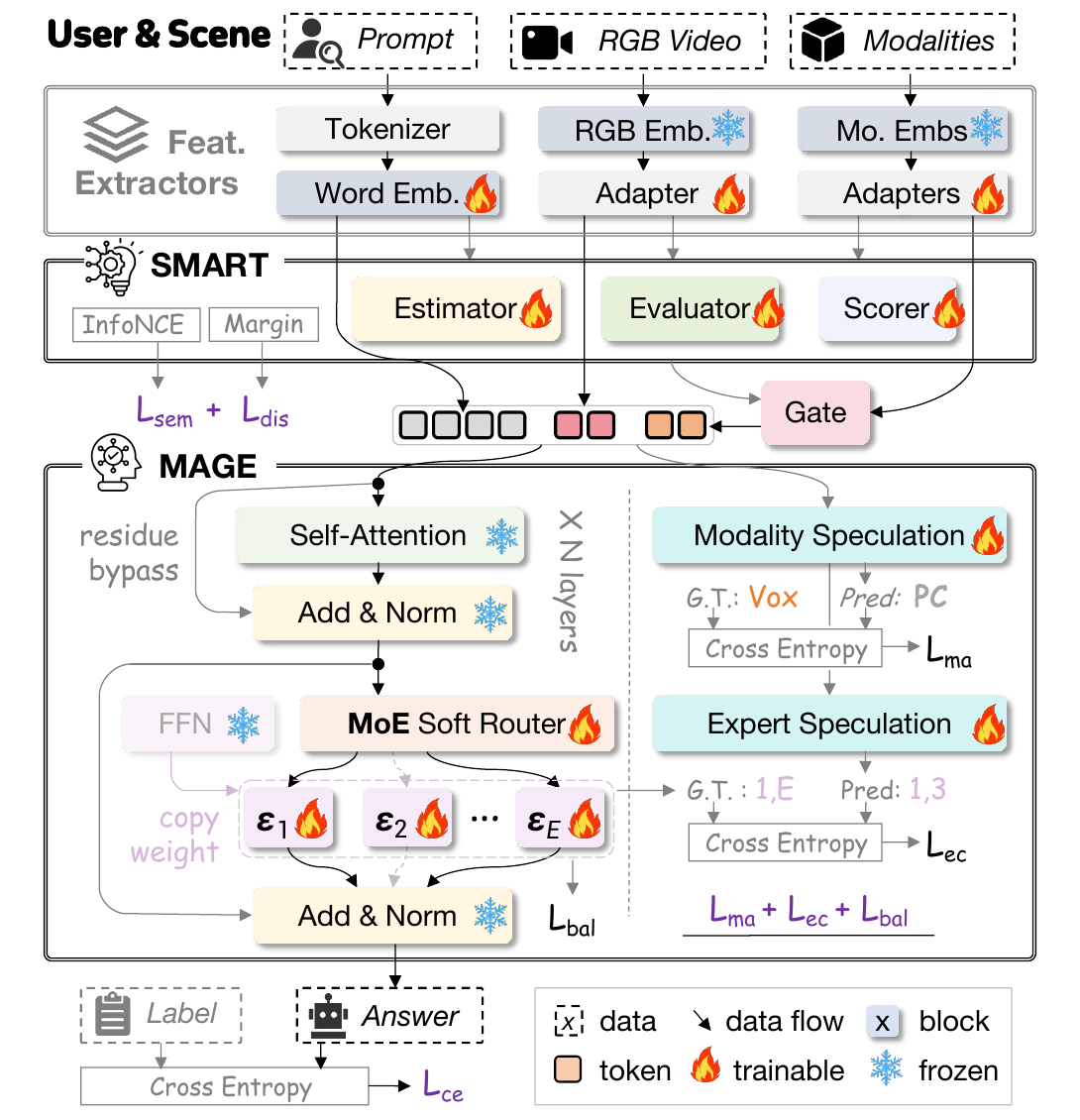}
\vspace{-2em}
    \caption{Overview of the end-to-end training strategy.} 
    \label{fig:train-stage}
\vspace{-1em}
\end{figure}

\subsection{Network Optimization}\label{sec:obj}
As shown in Fig.~\ref{fig:train-stage}, we adopt end-to-end selective fine-tuning.
Visual encoders are frozen, while lightweight adapters remain trainable.
SMART is jointly optimized with its estimator, evaluator, and scorer.
For MAGE, selected LLM layers use router--expert blocks with 8 experts and top-2 routing.
Experts are initialized from pretrained LLM FFNs~\cite{bai2025qwen3}, while only router--expert branches are updated.

The overall training objective is:
\begin{equation}
\mathcal{L} = 
\mathcal{L}_{\text{ce}} + 
    \underbrace{
\lambda_{\text{sem}} \mathcal{L}_{\text{sem}} + \lambda_{\text{dis}} \mathcal{L}_{\text{dis}}
   }_{\text{semantic alignment}} \\
    +   
    \underbrace{
    \mathcal{L}_{\text{ma}} +  \mathcal{L}_{\text{ec}} + \lambda_{\text{bal}} \mathcal{L}_{\text{bal}} 
    }_{\text{expert assignment}},
\end{equation}
where $\mathcal{L}_{\text{ce}}$ is the standard cross-entropy loss~\citep{qi2025gpt4scene} for language modeling.
Semantic alignment terms encourage the router to prioritize modalities that are semantically aligned with the input query.
Expert assignment terms regularize expert routing to ensure modality-consistent activation and balanced expert utilization.

\noindent
\textit{\textbf{Semantic Alignment Objectives.}}
We employ complementary objectives to encourage semantically consistent modality routing.
The semantic correlation loss $\mathcal{L}_{\text{sem}}$ increases the similarity between the query and relevant modalities~\citep{oord2018representation}.
The discrimination loss $\mathcal{L}_{\text{dis}}$ enlarges the gap between relevant and irrelevant ones~\citep{schroff2015facenet}.

\noindent
\textit{\textbf{Expert Assignment Objectives.}}
We regularize expert routing to achieve modality-consistent and balanced specialization.
The modality attribution loss $\mathcal{L}_{\text{ma}}$ supervises token-level modality distributions, and the expert calibration loss $\mathcal{L}_{\text{ec}}$ aligns routing decisions with modality-aware priors~\citep{hinton2015distilling}.
Meanwhile, the balancing loss $\mathcal{L}_{\text{bal}}$ encourages uniform expert utilization, preventing collapse and stabilizing training~\citep{shazeer2017outrageously}.

Together, these objectives enable semantically consistent modality selection and stable expert specialization during training.
Detailed formulations of all loss terms are provided in the Appendix.

\begin{table*}[t!]
    \centering
    \caption{Performance comparison on 3D scene understanding benchmarks, including ScanQA~\cite{azuma2022scanqa}, SQA3D~\cite{ma2022sqa3d}, Scan2Cap~\cite{chen2021scan2cap}, ScanRefer~\cite{chen2020scanrefer} and Multi3DRefer~\cite{zhang2023multi3drefer}. The symbol {\textcolor{mygreen}{\ding{51}}} indicates the modality used by each method. EM@1 denotes top-1 exact match. C@0.25/0.5, Acc@0.25/0.5, and F1@0.25/0.5 correspond to CIDEr, grounding accuracy, and F1 scores under IoU thresholds of 0.25/0.5.  $\dagger$ denotes results without high-resolution settings. The best and the second-best results are colored in \textcolor{myred}{red} and \textcolor{myblue}{blue}.}
    \vspace{-0.5em}
\resizebox{\textwidth}{!}{
\begin{tabular}{lcccccccccccccc}
    \toprule
    \multirow{3}{*}{\textbf{Method}} 
    & \multicolumn{5}{c}{\textbf{Modality Coverage}} 
    & \multicolumn{2}{c}{\textbf{3D Question Answering}} 
    & \multicolumn{2}{c}{\textbf{3D Dense Captioning}} 
    & \multicolumn{4}{c}{\textbf{3D Visual Grounding}} \\
    
    \cmidrule(lr){2-6} \cmidrule(lr){7-8} \cmidrule(lr){9-10} \cmidrule(lr){11-14}
    
    & \multirow{2}{*}{\textbf{RGB}} 
    & \multirow{2}{*}{\textbf{BEV}} 
    & \multirow{2}{*}{\textbf{Depth}} 
    & \multirow{2}{*}{\textbf{PC}} 
    & \multirow{2}{*}{\textbf{Voxel}} 
    & \textbf{ScanQA} 
    & \textbf{SQA3D} 
    & \multicolumn{2}{c}{\textbf{Scan2Cap}} 
    & \multicolumn{2}{c}{\textbf{ScanRefer}} 
    & \multicolumn{2}{c}{\textbf{Multi3DRefer}} \\

    &  &  &  &  &  
    & EM@1 $\uparrow$
    & EM@1 $\uparrow$
    & C@0.25 $\uparrow$
    & C@0.5 $\uparrow$
    & Acc@0.25 $\uparrow$ 
    & Acc@0.5 $\uparrow$
    & F1@0.25 $\uparrow$ 
    & F1@0.5 $\uparrow$ \\

\midrule
3D-VLP ~\citep{jin2023context}              &           &           &           & \textcolor{mygreen}{\ding{51}} &           & -      & 54.9  & 70.7       & 55.0      & 51.7      & 40.5      & -            & -            \\
3D-VisTA~\citep{zhu20233d}                  &           &           &           & \textcolor{mygreen}{\ding{51}} &           & 22.4   & 48.5  & 71.0       & 66.9      & 50.6      & 45.8      & -            & -            \\
Qwen2-VL-7B~\citep{wang2024qwen2}           & \textcolor{mygreen}{\ding{51}} &           &           &           &           & -      & 40.7  & 0.0        & 0.0       & 5.4       & 5.1       & 21.1         & 19.9         \\
LLaVA-Video-7B~\citep{zhang2024video}          & \textcolor{mygreen}{\ding{51}} &           &           &           &           & -      & 48.5  & -          & -         &      -     &   -        &      -        &      -        \\
PQ3D~\citep{zhu2024unifying}                & \textcolor{mygreen}{\ding{51}} &           &           & \textcolor{mygreen}{\ding{51}} & \textcolor{mygreen}{\ding{51}} & 20.0   & 47.1  & \textcolor{myblue}{87.1}       & 80.3      & 57.0      & 51.2      & -            & 50.1         \\
LAMM~\citep{yin2023lamm}                    & \textcolor{mygreen}{\ding{51}} &           &           & \textcolor{mygreen}{\ding{51}} &           & -      & -     & -          & -         & -         & 3.38      & -            & -            \\
3D-LLM~\citep{hong20233d}                   & \textcolor{mygreen}{\ding{51}} &           & \textcolor{mygreen}{\ding{51}} & \textcolor{mygreen}{\ding{51}} & \textcolor{mygreen}{\ding{51}} & 20.5   & -     & -          & -         & 30.3      & -         & -            & -            \\
Grounded 3D-LLM~\cite{chen2024grounded}                     &           &           &           & \textcolor{mygreen}{\ding{51}} &           &        & -     & 74.6       & 70.4      & 47.9      & 44.1      & 45.2         & 40.6         \\
Spatial 3D-LLM~\cite{wang2025spatial} & & & & \textcolor{mygreen}{\ding{51}} & & - & 46.2 & - & 72.2 &  44.3 & 37.2 & 48.3 & 41.2\\
Chat-3D~\citep{wang2023chat}                &           &           &           & \textcolor{mygreen}{\ding{51}} &           & -      & -     & -          & -         & -         & -         & -            & -            \\
Chat-3D V2~\cite{huang2023chat}                          &           &           &           & \textcolor{mygreen}{\ding{51}} &           & 22.9   & 54.7  & -          & -         & 42.5      & 38.4      & 45.1         & 41.6         \\
Chat-Scene~\citep{huang2024chat}            & \textcolor{mygreen}{\ding{51}} &           &           & \textcolor{mygreen}{\ding{51}} &           & 21.6   & 54.6  & 81.9       & 77.1      & 55.5      & 50.2      & 57.1         & 52.4         \\
LL3DA~\citep{chen2024ll3da}                 &           &           &           & \textcolor{mygreen}{\ding{51}} &           & -      & -     & 74.2       & 65.2      & -         & -         & -            & -            \\
LLaVA-3D~\citep{zhu2025llava}               & \textcolor{mygreen}{\ding{51}} & \textcolor{mygreen}{\ding{51}} &           &           &           & 27.0   & 55.6  & -          & 79.2      & 54.1      & 42.7      & -            & -            \\
LEO~\citep{huang2024embodied}               & \textcolor{mygreen}{\ding{51}} &           &           &           &           & -      & 50.0  & -          & 72.4      & -         & -         & -            & -            \\
Scene-LLM~\citep{fu2024scene}               & \textcolor{mygreen}{\ding{51}} &           & \textcolor{mygreen}{\ding{51}} & \textcolor{mygreen}{\ding{51}} & \textcolor{mygreen}{\ding{51}} & 27.2   & 54.2  & -          & 37.9      & -         & -         & -            & -            \\
GPT4Scene$^{\dagger}$~\citep{qi2025gpt4scene} & \textcolor{mygreen}{\ding{51}} & \textcolor{mygreen}{\ding{51}} &          &           &           & -      & -     & 63.8       & 60.6      & 40.5      & 36.7      & 45.4         & 42.1         \\
Video-3D LLM~\cite{zheng2025video}          & \textcolor{mygreen}{\ding{51}} &           & \textcolor{mygreen}{\ding{51}} &           &           & 30.1  & 58.6  & -          & \textcolor{myblue}{83.8}      & 58.1      & 51.7      & 58.0         & 52.7         \\
Ross3D~\cite{wang2025ross3d}                & \textcolor{mygreen}{\ding{51}} & \textcolor{mygreen}{\ding{51}} &           &           &           & \textcolor{myblue}{30.8}  & \textcolor{myblue}{63.0}  & -          & 81.3      & \textcolor{myblue}{61.1}      & \textcolor{myblue}{54.4}      & \textcolor{myblue}{59.6}         & \textcolor{myblue}{54.3}         \\
\textbf{SmartMage (ours)}                            & \textcolor{mygreen}{\ding{51}} & \textcolor{mygreen}{\ding{51}} & \textcolor{mygreen}{\ding{51}} & \textcolor{mygreen}{\ding{51}} & \textcolor{mygreen}{\ding{51}} & \textcolor{myred}{\textbf{32.6}}   & \textcolor{myred}{\textbf{66.8}}  & \textcolor{myred}{\textbf{93.8}}       & \textcolor{myred}{\textbf{88.7}}      & \textcolor{myred}{\textbf{65.9}}    & \textcolor{myred}{\textbf{59.5}}     &  \textcolor{myred}{\textbf{65.4}}         &  \textcolor{myred}{\textbf{60.7}}        \\ 
\bottomrule
\end{tabular}
}
\label{tab:overall_performance}
\end{table*}

\section{Experiments}
\subsection{Experiment Settings}\label{sec:exper-set}
\noindent\textbf{Datasets.}
For training, we construct a unified corpus from ScanNet~\citep{dai2017scannet}-based annotations that cover diverse 3D scene understanding tasks, enabling joint learning under a shared scene representation.
For evaluation, we consider three groups of benchmarks: standard 3D scene understanding benchmarks, RGB-only video understanding benchmarks, and a diagnostic benchmark constructed in this work.
Specifically, we evaluate on five widely used 3D benchmarks: ScanQA~\citep{azuma2022scanqa} and SQA3D~\citep{ma2022sqa3d} for 3D question answering, Scan2Cap~\citep{chen2021scan2cap} for dense captioning, and ScanRefer~\citep{chen2020scanrefer} together with Multi3DRefer~\citep{zhang2023multi3drefer} for 3D visual grounding.
To assess robustness when only RGB observations are available, we further evaluate on RGB-only video understanding benchmarks, including VSI-Bench~\citep{yang2025thinking}, VSI-SUPER~\citep{yang2025cambrian}, and MMSI-Bench~\citep{yang2025mmsi}.

To investigate semantic-modality dependence, we construct a diagnostic benchmark named ScanFacet.
ScanFacet reorganizes the question-answer pairs from ScanQA~\citep{azuma2022scanqa} and SQA3D~\citep{ma2022sqa3d} into eight semantic categories: \emph{color}, \emph{location}, \emph{material}, \emph{number}, \emph{object shape}, \emph{object type}, \emph{spatial relation}, and \emph{other}.
Each sample is categorized through an LLM-assisted taxonomy pipeline comprising semantic parsing, intent normalization, and self-consistency filtering, followed by light human verification to ensure label reliability.

\noindent\textit{\textbf{Evaluation Metrics.}}
Following~\citep{zhu2025llava, wang2025ross3d}, we report EM@1 for 3D question answering, CIDEr@0.25/0.5 for 3D dense captioning, and Acc@0.25/0.5 or F1@0.25/0.5 for 3D visual grounding. For RGB video benchmarks, we report answer accuracy.

\noindent\textit{\textbf{Implementation Details.}}
We initialize the model from Qwen3-VL-8B-Instruct~\cite{bai2025qwen3}.
SmartMage is optimized using AdamW with a learning rate of 2e-5, a warm-up ratio of 0.03, and cosine decay.
The loss weights are set to $\lambda_{\text{sem}}=0.5$, $\lambda_{\text{dis}}=1.0$, and $\lambda_{\text{bal}}=0.01$.
Training is conducted for 1 epoch with a batch size of 64 on 2$\times$H800 GPUs, using BF16 mixed precision and DeepSpeed ZeRO-2.
Each scene is sampled into 32 RGB-D frames with a resolution of 128$\times$123.
BEV images are rendered from meshes at the same resolution.
Point clouds are sampled to 8192 points, and voxel representations use a voxel size of 0.02.
More details can be found in the Appendix.

\subsection{Comparison with State-of-the-art Methods} 
As shown in Table~\ref{tab:overall_performance}, SmartMage consistently outperforms prior state-of-the-art methods across five benchmarks covering 3D question answering, dense captioning, and visual grounding.

\noindent
\textit{\textbf{3D Question Answering.}}
Our method achieves 32.6 EM@1 on ScanQA~\cite{azuma2022scanqa} and 66.8 EM@1 on SQA3D~\cite{ma2022sqa3d}, outperforming the previous SOTA Ross3D~\cite{wang2025ross3d} by +1.8 and +3.8, respectively.
These consistent gains indicate that dynamic modality orchestration improves both general spatial reasoning (ScanQA~\citep{azuma2022scanqa}) and embodied, context-aware understanding (SQA3D~\citep{ma2022sqa3d}).

\noindent
\textit{\textbf{3D Dense Captioning.}}
Following~\cite{wang2023chat, huang2024embodied}, we generate captions for each detected object proposal and evaluate under IoU thresholds of 0.25 / 0.5.
Our method achieves 93.8 CIDEr@0.25 and 88.7 CIDEr@0.5 on Scan2Cap~\cite{chen2021scan2cap}, surpassing PQ3D~\cite{zhu2024unifying} by +6.7 at CIDEr@0.25 and Video-3D LLM~\cite{zheng2025video} by +4.9 at CIDEr@0.5.
These improvements demonstrate improved alignment between visual structure and language generation, leading to stronger spatial grounding and more coherent object-level descriptions.

\noindent
\textit{\textbf{3D Visual Grounding.}}
Following~\cite{huang2024chat, qi2025gpt4scene}, we detect all objects and perform grounding over the generated proposals.
Our method achieves 65.9 / 59.5 Acc@0.25/0.5 on ScanRefer~\cite{chen2020scanrefer}, and 65.4 / 60.7 F1@0.25/0.5 on Multi3DRefer~\cite{zhang2023multi3drefer}.
Compared to previous SOTA Ross3D~\cite{wang2025ross3d}, our model yields notable gains of +4.8 / +5.1 Acc on ScanRefer~\cite{chen2020scanrefer} and +5.8 / +6.4 F1 on Multi3DRefer~\cite{zhang2023multi3drefer}, demonstrating more precise localization across diverse grounding scenarios.

\begin{figure}[t!]
    \centering
    \includegraphics[width=1.0\linewidth]{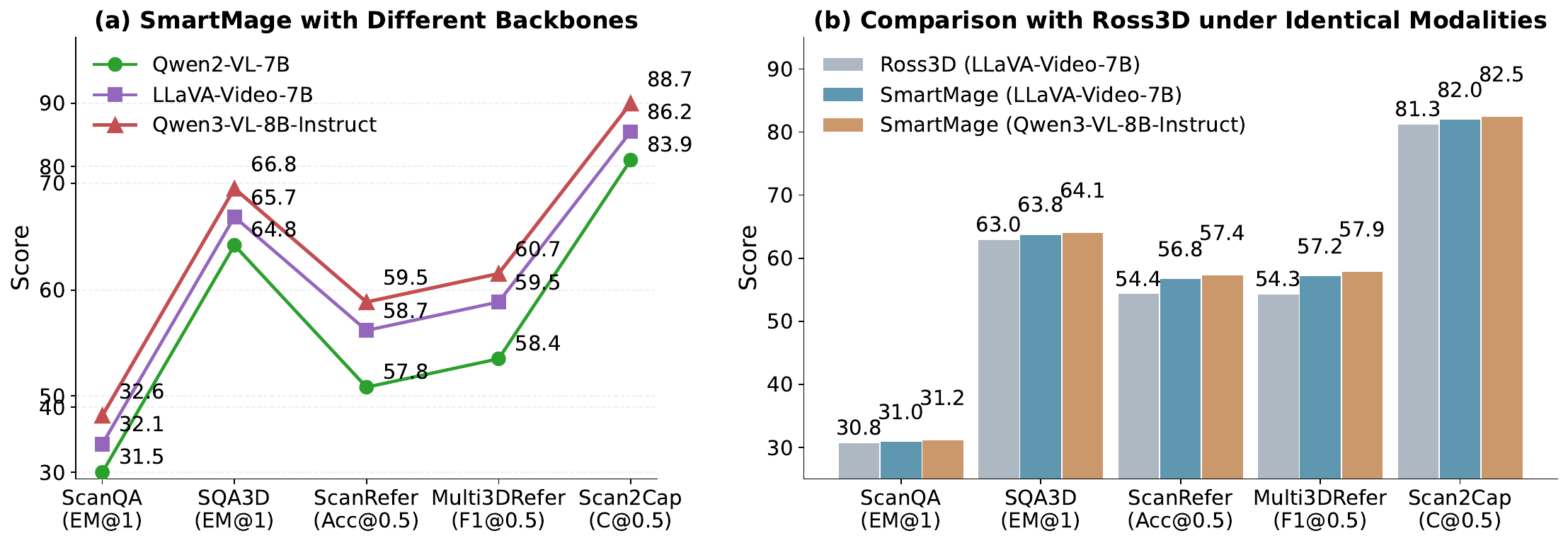}
    \vspace{-2em}
    \caption{Controlled comparison of SmartMage. (a) SmartMage serves as a flexible framework compatible with various backbones. (b) Under the same modalities, resolution, and backbone, SmartMage still outperforms Ross3D.}
    \label{fig:ross3d-comparision}
    \vspace{-1em}
\end{figure}

\noindent
\textit{\textbf{Architectural Gains Are Independent of Backbone Strength and Modality Scaling.}}
As shown in Fig.~\ref{fig:ross3d-comparision}(a), SmartMage serves as a flexible framework compatible with various backbones, including Qwen3-VL-8B-Instruct~\citep{bai2025qwen3}, Qwen2-VL-7B-Instruct~\citep{wang2024qwen2} and LLaVA-Video-7B~\citep{zhang2024video}. Across diverse backbones, our method consistently achieves competitive performance on five 3D scene understanding benchmarks.
As shown in Fig.~\ref{fig:ross3d-comparision}(b), under the same modalities (RGB, BEV, Depth), resolution ($432 \times 432$), and backbone (LLaVA-Video-7B~\citep{zhang2024video}), SmartMage consistently outperforms Ross3D~\citep{wang2025ross3d} across all five benchmarks.
This demonstrates that the performance improvements stem from our dynamic modality orchestration rather than more input modalities and stronger backbone.

\begin{table}[t!]
\footnotesize
    \centering
    \caption{Ablation study on SMART and MAGE modules.}
    \vspace{-1em}
 \setlength{\tabcolsep}{2.2pt}{
    \begin{tabular}{cccccc}
    \toprule
          \multirow{2}{*}{\textbf{Method}} & \textbf{ScanQA} & \textbf{SQA3D} & \textbf{Scan2Cap} & \textbf{ScanRefer} & \textbf{Multi3DRefer} \\
          \cmidrule(lr){2-2}  \cmidrule(lr){3-3} \cmidrule(lr){4-4} \cmidrule(lr){5-5} \cmidrule(lr){6-6}
         & EM@1 $\uparrow$ & EM@1 $\uparrow$ & C@0.5 $\uparrow$ & Acc@0.5 $\uparrow$ & F1@0.5 $\uparrow$ \\
       \midrule
       \rowcolor{gray!15!white}\multicolumn{6}{c}{\textit{w/ SMART (w/o MAGE)}} \\
        w/o SPE  & 28.2 & 62.1 & 78.4 & 55.2 & 54.1 \\
        w/o SSS  & 27.1 & 60.8 & 77.2 & 54.1 & 52.2 \\
        w/o MQE  & 28.6 & 62.5 & 79.1 & 55.8 & 54.6 \\
        \textbf{w/ SMART (full)}  & 29.8 & 63.4 & 82.5 & 57.1 & 56.2\\ 
        \midrule
     \rowcolor{gray!15!white}\multicolumn{6}{c}{\textit{w/ MAGE (w/o SMART)}} \\
       w/o MES & 29.4 & 63.1 & 79.8 & 56.5 & 55.8 \\
       \textbf{w/ MAGE (full)} & 30.7 & 64.5 & 83.8 & 57.8 & 57.2 \\
      \midrule
      \rowcolor{gray!15!white}\multicolumn{6}{c}{\textit{w/ SMART + MAGE}} \\
      \textbf{SmartMage (ours)}  & 32.6   & 66.8       & 88.7      & 59.5      &  60.7        \\ 
    \bottomrule
    \end{tabular}
    }
    \label{tab:abla_module}
\end{table}

\subsection{Ablation Studies}

\noindent
\textit{\textbf{Effect of SMART \& MAGE Modules.}}
Table~\ref{tab:abla_module} presents the effect of each component in SMART and MAGE.
Within SMART, removing SSS leads to the largest performance drop (\textit{e.g.}, from 29.8 to 27.1 EM@1 on ScanQA~\citep{azuma2022scanqa} and from 56.2 to 52.2 F1@0.5 on Multi3DRefer~\citep{zhang2023multi3drefer}), highlighting the importance of semantic similarity for adaptive routing.
Similarly, removing MES in MAGE decreases performance from 64.5 to 63.1 EM@1 on SQA3D~\cite{ma2022sqa3d} and from 57.2 to 55.8 F1@0.5 on Multi3DRefer~\cite{zhang2023multi3drefer}, demonstrating its effectiveness for modality-aware expert allocation.

\noindent
\textit{\textbf{Effect of Loss Design.}}
Table~\ref{tab:abla_loss} shows that different loss components play distinct and complementary roles.
Removing $\mathcal{L}_{\text{sem}}$ causes the most significant degradation, indicating that explicit query--modality alignment is essential for selecting relevant modalities.
In contrast, $\mathcal{L}_{\text{dis}}$ brings only marginal gains, suggesting it mainly refines the alignment.
For expert assignment, $\mathcal{L}_{\text{ma}}$ alone provides limited improvement, while further removing $\mathcal{L}_{\text{ec}}$ and $\mathcal{L}_{\text{bal}}$ leads to substantial drops, revealing that effective routing requires both modality-aware priors and balanced expert utilization.
Overall, the results highlight that both semantic alignment and structured routing are critical for robust multimodal reasoning.

\begin{table}[t!]
\footnotesize
\centering
\caption{Ablation study on the effect of loss design.}
\vspace{-1em}
\setlength{\tabcolsep}{1.2pt}{
\begin{tabular}{cccccc}
\toprule
\multirow{2}{*}{\textbf{Method}} & \textbf{ScanQA} & \textbf{SQA3D} & \textbf{Scan2Cap} & \textbf{ScanRefer} & \textbf{Multi3DRefer} \\
\cmidrule(lr){2-2} \cmidrule(lr){3-3} \cmidrule(lr){4-4} \cmidrule(lr){5-5} \cmidrule(lr){6-6}
& EM@1$\uparrow$ & EM@1 $\uparrow$ & C@0.5 $\uparrow$ & Acc@0.5 $\uparrow$ & F1@0.5 $\uparrow$ \\
\midrule
w/o $\mathcal{L}_{\text{sem}}$ & 28.9 & 62.6 & 84.7 & 57.1 & 59.3 \\
w/o $\mathcal{L}_{\text{dis}}$ & 30.4 & 64.2 & 85.9 & 58.0 & 59.9 \\
\midrule
w/o $\mathcal{L}_{\text{ma}}$ & 30.8 & 64.6 & 86.3 & 58.2 & 60.1 \\
w/o $\mathcal{L}_{\text{ma}} + \mathcal{L}_{\text{ec}}$ & 29.9 & 63.5 & 85.1 & 57.4 & 59.3 \\
w/o $\mathcal{L}_{\text{ma}} + \mathcal{L}_{\text{ec}} + \mathcal{L}_{\text{bal}}$ & 28.7 & 62.1 & 83.4 & 56.5 & 58.2 \\
\midrule
w/ $\mathcal{L}$ & 32.6 & 66.8 & 88.7 & 59.5 & 60.7 \\
\bottomrule
\end{tabular}
}
\label{tab:abla_loss}
\vspace{-1em}
\end{table}

\begin{table}[t!]
\footnotesize
\centering
\caption{Ablation on modality selection (fixed \textit{vs.} adaptive). Best results are in \textcolor{myred}{red}, and second-best results are in \textcolor{myblue}{blue}.}
\vspace{-1em}
\setlength{\tabcolsep}{2pt}{
\begin{tabular}{c|ccccc|ccccc}
\toprule
\textbf{No.} & \multicolumn{5}{c|}{~~\textbf{Modality}~~} & \textbf{ScanQA} & \textbf{SQA3D} & \textbf{Scan2Cap} & \textbf{ScanRefer} & \textbf{Multi3DRefer} \\
\cmidrule(lr){1-1} \cmidrule(lr){2-6} \cmidrule(lr){7-7} \cmidrule(lr){8-8} \cmidrule(lr){9-9} \cmidrule(lr){10-10} \cmidrule(lr){11-11}
 & \rotatebox{90}{RGB} & \rotatebox{90}{BEV} & \rotatebox{90}{Depth} & \rotatebox{90}{PC} & \rotatebox{90}{Voxel}
& EM@1$\uparrow$ & EM@1$\uparrow$ & C@0.5$\uparrow$ & Acc@0.5$\uparrow$ & F1@0.5$\uparrow$ \\
\midrule

 \rowcolor{gray!15!white}
 \multicolumn{11}{c}{\textit{Fixed Modality Combination}} \\ 

1 & \textcolor{mygreen}{\ding{51}} &  &  &  &  
& 28.5 & 59.2 & 78.4 & 54.1 & 55.6 \\

2 & \textcolor{mygreen}{\ding{51}} & \textcolor{mygreen}{\ding{51}} &  &  &  
& 28.9 & 60.5 & 79.0 & 54.4 & 55.9 \\

3 & \textcolor{mygreen}{\ding{51}} &  & \textcolor{mygreen}{\ding{51}} &  &  
& 29.6 & 61.7 & 80.6 & 55.4 & 56.3 \\

4 & \textcolor{mygreen}{\ding{51}} &  &  & \textcolor{mygreen}{\ding{51}} &  
& 29.3 & 61.4 & 81.4 & 56.0 & 56.7 \\

5 & \textcolor{mygreen}{\ding{51}} &  &  &  & \textcolor{mygreen}{\ding{51}} 
& 29.8 & 61.9 & 81.8 & 56.9 & 56.9 \\

\midrule

6 & \textcolor{mygreen}{\ding{51}} & \textcolor{mygreen}{\ding{51}} & \textcolor{mygreen}{\ding{51}} &  & 
& 29.4 & 61.7 & 80.2 & 56.1 & 56.4 \\

7 & \textcolor{mygreen}{\ding{51}} & \textcolor{mygreen}{\ding{51}} &  & \textcolor{mygreen}{\ding{51}} &  
& 30.1 & 62.6 & 82.4 & 57.3 & 57.5 \\

8 & \textcolor{mygreen}{\ding{51}} & \textcolor{mygreen}{\ding{51}} &  &  & \textcolor{mygreen}{\ding{51}} 
& \textcolor{myblue}{30.9} & 63.4 & 83.6 & \textcolor{myblue}{58.2} & \textcolor{myblue}{58.0} \\

\midrule

9 & \textcolor{mygreen}{\ding{51}} & \textcolor{mygreen}{\ding{51}} &  & \textcolor{mygreen}{\ding{51}} &  \textcolor{mygreen}{\ding{51}}
& 30.3 &  \textcolor{myblue}{64.7} & \textcolor{myblue}{84.2} & 58.0  & 57.7  \\

10 & \textcolor{mygreen}{\ding{51}} & \textcolor{mygreen}{\ding{51}} & \textcolor{mygreen}{\ding{51}} & \textcolor{mygreen}{\ding{51}} & \textcolor{mygreen}{\ding{51}} 
& 30.7 & 64.5 & 83.8 & 57.8 & 57.2 \\

\rowcolor{gray!15!white}
\multicolumn{11}{c}{\textit{Adaptive Modality Selection (ours)}} \\ 

11 & \textcolor{mygreen}{\ding{51}} & \textcolor{mygreen}{\ding{51}} & \textcolor{mygreen}{\ding{51}} & \textcolor{mygreen}{\ding{51}} & \textcolor{mygreen}{\ding{51}} 
& \textcolor{myred}{\textbf{32.6}} 
& \textcolor{myred}{\textbf{66.8}} 
& \textcolor{myred}{\textbf{88.7}} 
& \textcolor{myred}{\textbf{59.5}} 
& \textcolor{myred}{\textbf{60.7}} \\

\bottomrule
\end{tabular}
}
\label{tab:abla_modaltiy}
\vspace{-1em}
\end{table}

\begin{figure}[t!]
    \centering
    \includegraphics[width=\linewidth]{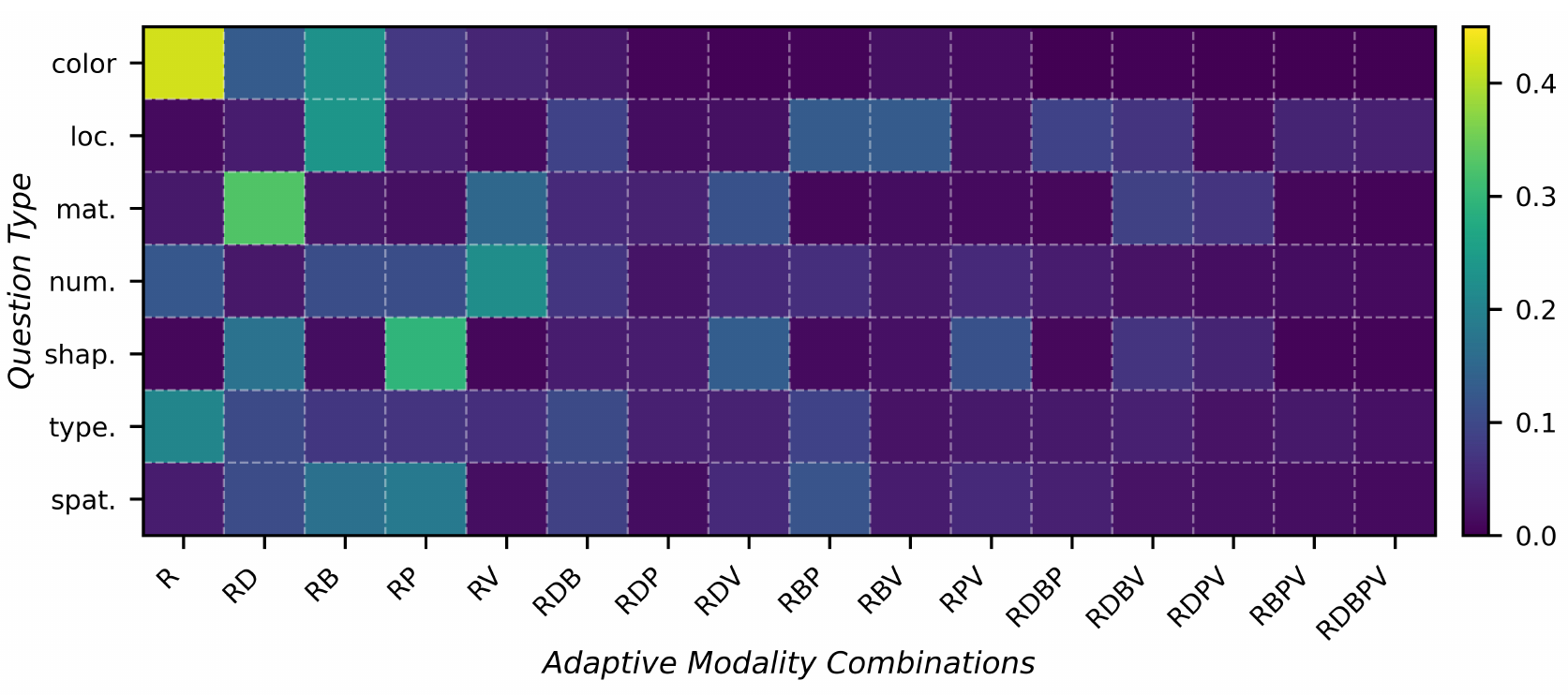}
    \vspace{-2em}
    \caption{Modality preference across seven question types on ScanFacet. Each cell shows the normalized frequency of a modality combination selected for a given question type. 
    Question types are color, location, material, number, shape, type and spatial relation. R, B, D, P, and V denote RGB, BEV, Depth, Point Cloud, and Voxel, respectively.}
    \label{fig:modality_selection}
    \vspace{-1.5em}
\end{figure}

\noindent
\textit{\textbf{Effect of Adaptive Modality Selection.}}
Table~\ref{tab:abla_modaltiy} compares fixed-modality combination with adaptive modality selection.
Using only RGB (setting 1) yields limited performance, while adding one complementary modality generally improves the results (settings 2--5).
We observe that introducing 3D information into RGB often leads to performance gains, and voxel-based representations tend to perform slightly better than point clouds and depth in several settings.
However, adding more modalities does not consistently improve performance; for example, setting 8 outperforms both settings 9 and 10 on several benchmarks despite using fewer modalities, indicating potential redundancy and interference.
Different fixed combinations show advantages on different benchmarks, suggesting task-dependent modality preferences.
This highlights the need for adaptive modality selection.
Accordingly, our adaptive strategy (setting 11) achieves the best performance across all benchmarks, demonstrating that dynamic routing better exploits complementary multimodal cues.

\begin{table}[t!]\footnotesize
\centering
\caption{Model Efficiency. Training cost reports parameters and time per iteration, and inference cost reports activated parameters, data preprocessing time (2D+3D Pre.), Time-to-First-Token (TTFT), and End-to-End (E2E) latency.}\label{tab:efficiency}
\vspace{-1em}
\setlength{\tabcolsep}{2.75pt}{
\begin{tabular}{l|cc|cccc}
\toprule
\multirow{2}{*}{\textbf{Method}}  & \multicolumn{2}{c|}{\textbf{Training Cost}} & \multicolumn{4}{c}{\textbf{Inference Cost}} \\
 \cmidrule(lr){2-3}  \cmidrule(lr){4-7}
& \makecell{Train / All \\ Params.} & \makecell{Time  \\(s/iter)} & \makecell{Act. \\ Params.} & \makecell{ 2D+3D Pre. \\ (ms)} & \makecell{TTFT \\ (ms)} & \makecell{E2E \\ (ms)} \\
\midrule
Video-3D LLM~\citep{zheng2025video}  & $\sim$ 7.7B / 8.1B & 91.61 &  $\sim$ 8.1B & 82.0 + 15.0 & 115.0 & 	527.0 \\ 
Ross3D~\citep{wang2025ross3d}  &  $\sim$ 7.7B / 8.1B  & 125.2   &   $\sim$ 8.1B &  57.6 + 25.0	& 135.0  & 551.6 \\ 
\textbf{SmartMage (ours)}   &  $\sim$ 6.3B / 12.6B & 47.44  & $\sim$8.8B & 48.5 + 64.8 & 112.5 & 538.2 \\
\bottomrule
\end{tabular}
}
\vspace{-1em}
\end{table}

\begin{figure}[t!]
    \centering
\includegraphics[width=\linewidth]{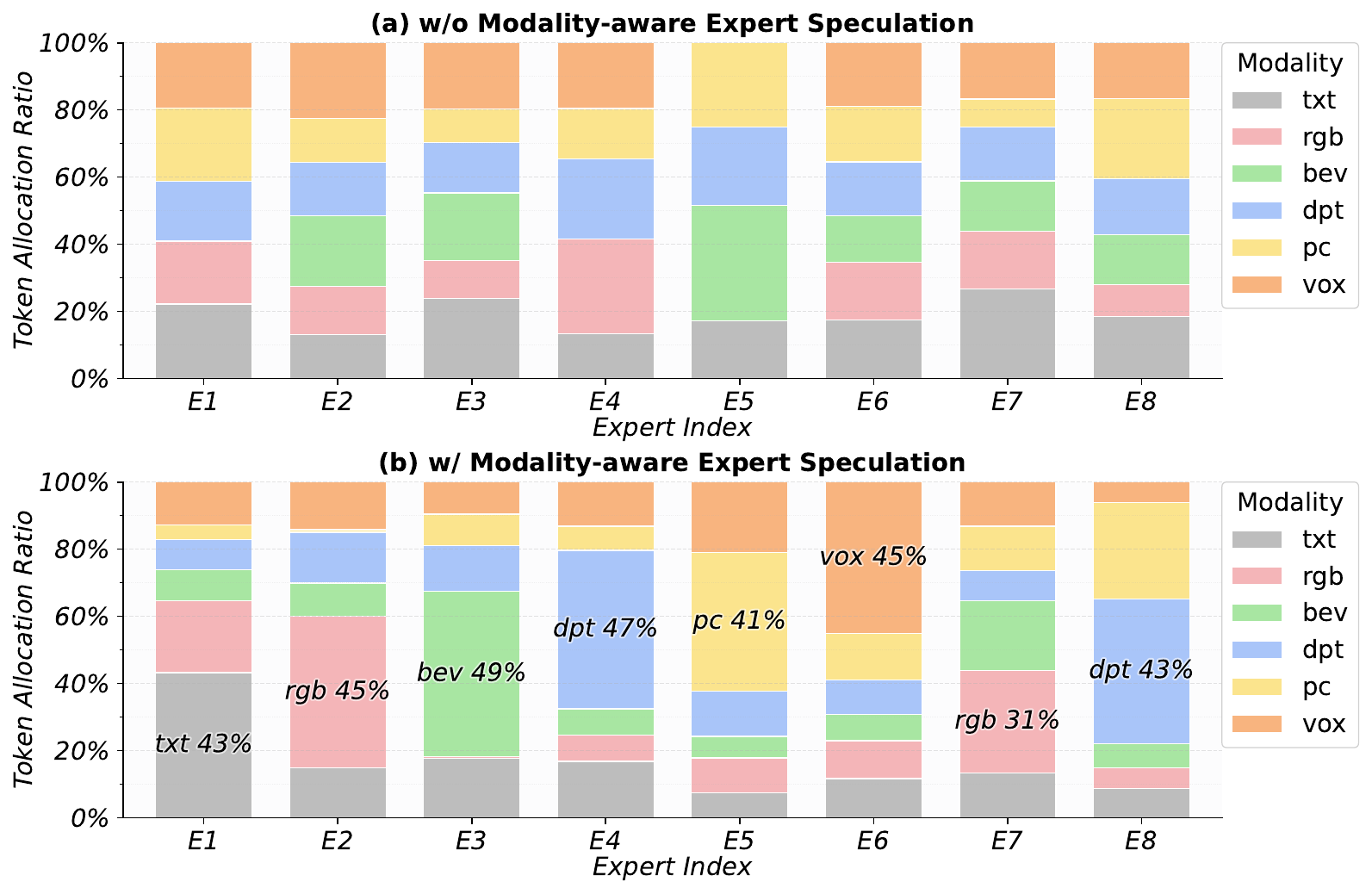} 
\vspace{-2em}
    \caption{Modality-aware expert speculation. Comparison of token-to-expert allocation ratios without (a) and with (b) the Modality-aware Expert Speculation (MES) mechanism.}
    \label{fig:expert_selection}
\vspace{-1em}
\end{figure}

\begin{figure}[t!]
    \centering
    \includegraphics[width=\linewidth]{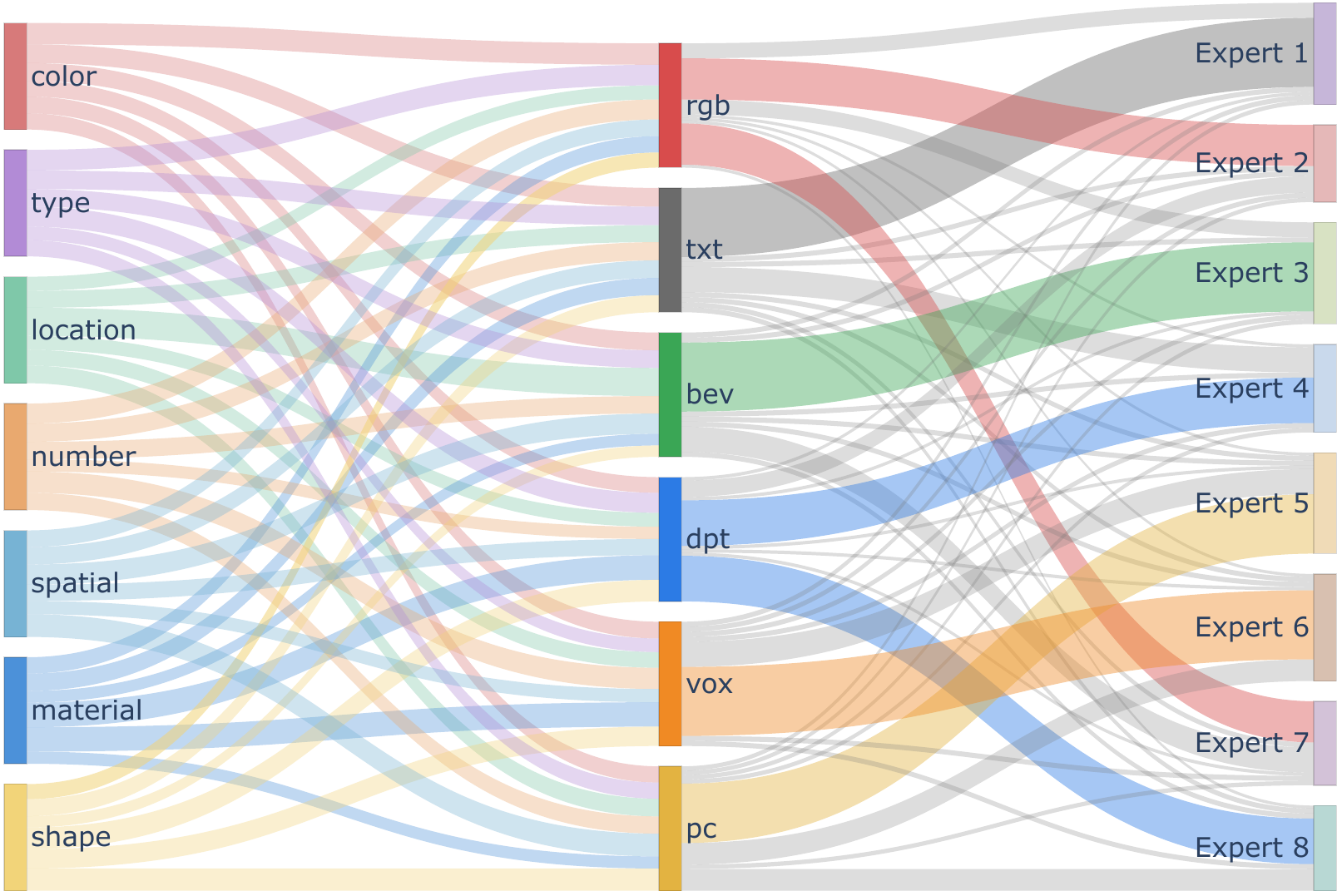} 
    \vspace{-1.5em}
    \caption{{Hierarchical routing from question types to modalities and experts. Line thickness represents normalized routing frequency, showing that different question types prefer distinct modality–expert pathways.}}
    \label{fig:q_m_e}
    \vspace{-1em}
\end{figure}

\begin{figure*}[t!]
    \centering
\includegraphics[width=1\linewidth]{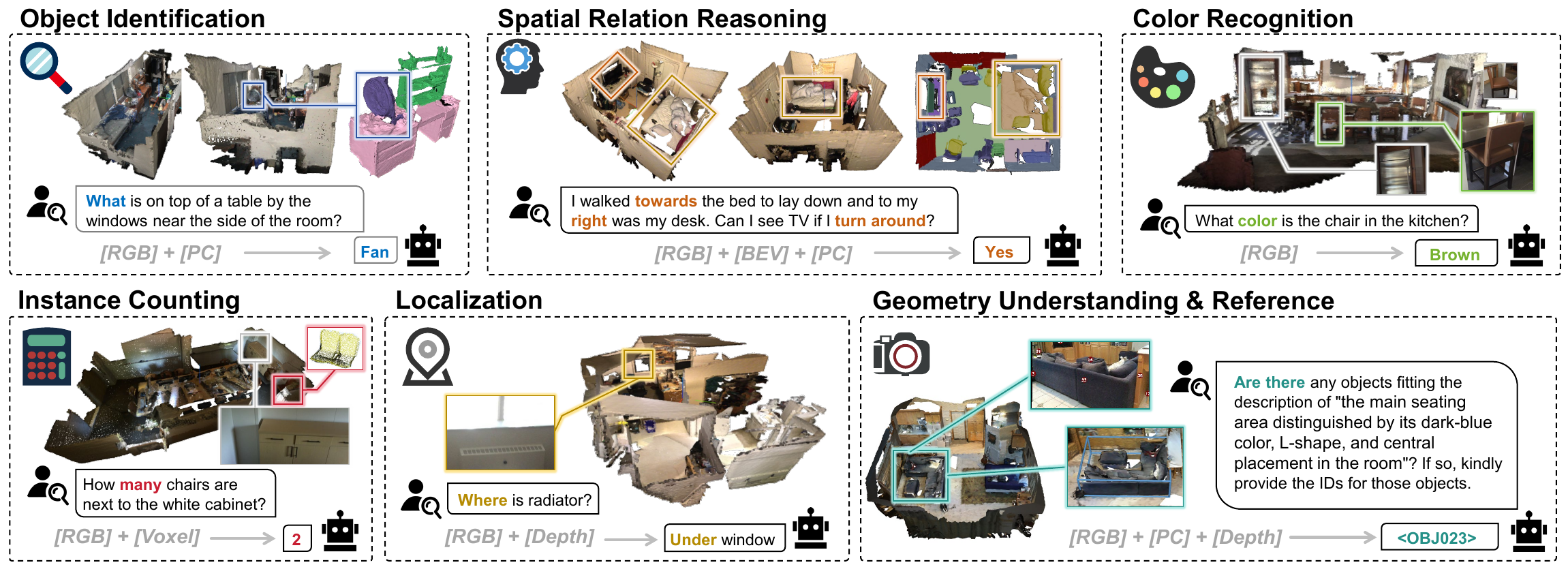}
\vspace{-2em}
    \caption{Qualitative visualization. The model adaptively selects task-relevant modalities for interpretable reasoning.} 
    \label{fig:visual}
    \vspace{-1em}
\end{figure*}

\subsection{Additional Analysis and Discussion}
We provide a comprehensive analysis of SmartMage from perspectives including efficiency, modality preference, expert specialization, interpretability, and robustness under limited modalities.

\noindent
\textit{\textbf{SmartMage improves training efficiency while maintaining competitive inference latency.}}
Table~\ref{tab:efficiency} shows that SmartMage reduces training time per iteration to 47.4s, compared to 91.6s and 125.2s for Video-3D LLM~\citep{zheng2025video} and Ross3D~\citep{wang2025ross3d}, respectively, achieving a $2.0\times$--$2.6\times$ speedup.
Data preprocessing is more costly due to complex 3D inputs (\textit{e.g.}, point clouds and voxels).
Despite this overhead, inference latency remains comparable, with Time-to-First-Token (TTFT) of 112.5ms and End-to-End (E2E) latency of 538.2ms, closely matching prior methods.
This is achieved through adaptive modality filtering in SMART and sparse expert activation in MAGE, which together reduce unnecessary computation.

\noindent
\textit{\textbf{SmartMage learns clear and semantically grounded modality preferences across different question types.}}
Fig.~\ref{fig:modality_selection} depicts the modality preference patterns learned by SMART for seven representative question types.
Color- and material-related questions favor RGB or RGB–Depth combinations, while spatial and counting ones rely more on geometry-aware modalities such as Depth and Voxel.
This indicates that SMART captures meaningful semantic–modality correlations rather than relying on uniform fusion.

\noindent
\textit{\textbf{MES induces modality-aware expert specialization with routing patterns aligned to question semantics.}}
Fig.~\ref{fig:expert_selection} shows that tokens are uniformly routed across experts without MES, whereas MES produces clear modality-specific expert preferences in the MoE layer.
Such structured specialization enables different experts to capture complementary modality characteristics.
Fig.~\ref{fig:q_m_e} further reveals distinct hierarchical question--modality--expert pathways across question categories, demonstrating that MES promotes meaningful expert differentiation and semantic-aware multimodal reasoning.

\noindent
\textit{\textbf{Qualitative visualization shows that SmartMage enables interpretable multimodal reasoning.}}
Fig.~\ref{fig:visual} demonstrates that our model adaptively routes task-relevant modalities under semantic guidance, achieving interpretable and task-aware multimodal reasoning across diverse 3D understanding tasks. 
More visualization results are provided in the Appendix.

\noindent
\textit{\textbf{SmartMage remains robust under limited modalities in RGB-only video settings.}}
We evaluate SmartMage zero-shot on RGB-only video benchmarks~\citep{yang2025thinking,yang2025cambrian,yang2025mmsi}, comparing it with both 2D video models~\citep{wang2024qwen2,lu2025internvl,an2025llava} and 3D scene understanding methods~\citep{qi2025gpt4scene,zheng2025video,wang2025ross3d}.
As shown in Table~\ref{tab:benchmark}, specialized 2D video models generally perform better, benefiting from longer temporal context and video-specific training.
Nevertheless, SmartMage remains competitive overall and performs well on relational reasoning, demonstrating stable generalization to RGB-only inputs under modality constraints.

\begin{table}[t!]\footnotesize
    \centering
    \caption{Comparison on RGB-only video benchmarks. Best results are in \textcolor{myred}{red}, and second-best results are in \textcolor{myblue}{blue}.}
    \label{tab:benchmark}
\vspace{-1em}
    \setlength{\tabcolsep}{1.25pt}{
    \begin{tabular}{l|cccc|cccc|c}
       \toprule
        \multirow{2}{*}{\textbf{Method}} & \multicolumn{4}{c|}{
        \makecell{\textbf{VSI-Bench} \\ \citep{yang2025thinking}}} &  \multicolumn{4}{c|}{\makecell{\textbf{VSI-SUPER-Recall} \\ \citep{yang2025cambrian}}} & 
        \makecell{\textbf{MMSI-Bench} \\ \citep{yang2025mmsi}}\\
        \cmidrule(lr){2-5} \cmidrule(lr){6-9} \cmidrule(lr){10-10}
        & \rotatebox{90}{Count}  & \rotatebox{90}{Rel. Dir.} & \rotatebox{90}{Route}  &  \rotatebox{90}{Avg.} & \rotatebox{90}{10min}  & \rotatebox{90}{30min}  & \rotatebox{90}{60min}  & \rotatebox{90}{240min}  & \rotatebox{90}{Avg.}  \\
         \midrule
        \rowcolor{gray!15!white}\multicolumn{10}{c}{\textit{Methods designed for 2D Video Understanding}} \\
         Qwen2-VL-7B~\citep{wang2024qwen2} & 27.3 & 35.9 & 22.2 & 14.7 & 26.7 & \textcolor{myblue}{28.3} & \textcolor{myblue}{28.3}  & \textcolor{myred}{\textbf{28.3}} & 24.5 \\
        InternVL2-8B~\citep{wang2025internvl3} &  23.1 & 30.7 & \textcolor{myblue}{29.9} & \textcolor{myblue}{34.6} &   - & - & - & - & 28.7 \\
        LLaVA-OneVision-7B~\citep{an2025llava} & \textcolor{myred}{\textbf{47.7}} & 35.2 & 29.4  & 32.4 & - & - & - & - & 24.5\\
        LLaVA-OneVision-72B~\citep{an2025llava}	& \textcolor{myblue}{43.5} & \textcolor{myblue}{39.9} & \textcolor{myred}{\textbf{32.5}} & \textcolor{myred}{\textbf{40.2}} &  - & - & - & -   & 28.4\\
         \midrule
         \rowcolor{gray!15!white}\multicolumn{10}{c}{\textit{Methods designed for 3D Scene Understanding}} \\
        GPT4Scene~\citep{qi2025gpt4scene} & 38.1  &  38.7 & 28.8 & 25.6 & \textcolor{myblue}{26.9}  & 26.6 & \textcolor{myblue}{28.3} & \textcolor{myblue}{26.6}  & 20.5 \\
        Video-3D LLM~\citep{zheng2025video}  & 23.4	& 35.1 &28.1 & 24.8 &   \textcolor{myblue}{26.9} & 26.9  & 25.0 & 25.0 &	27.8  \\
         Ross3D~\citep{wang2025ross3d} & 20.5 & 35.1 & 27.3 & 22.8 & 26.6 & 23.3 & 23.3 & 25.0 & \textcolor{myblue}{29.4} \\ 
        \textbf{SmartMage (ours)} & 26.5 &  \textcolor{myred}{\textbf{45.8}} & 29.4  & 27.8 &\textcolor{myred}{\textbf{31.8}} & \textcolor{myred}{\textbf{30.0}} & \textcolor{myred}{\textbf{30.0}} & \textcolor{myred}{\textbf{28.3}} & \textcolor{myred}{\textbf{30.3}} \\
         \bottomrule
    \end{tabular}
    }
    \vspace{-1em}
\end{table}

\section{Conclusion}
This paper introduces SmartMage, a novel MLLM for unified and adaptive 3D scene understanding. SmartMage decomposes multimodal reasoning into two stages: semantic-guided modality selection with SMART and modality-aware expert specialization with MAGE. Experiments show that SmartMage achieves state-of-the-art performance on five standard 3D scene understanding benchmarks, remains competitive under RGB-only inputs, and outperforms prior methods under matched settings. ScanFacet further reveals meaningful question-dependent modality preferences, underscoring the importance of dynamic modality orchestration for 3D scene understanding and future adaptive multimodal reasoning.

\begin{acks}
This work was supported by the National Natural Science Foundation of China (62472381, 62402432) and the Earth System Big Data Platform of the School of Earth Sciences, Zhejiang University.
\end{acks}

\bibliographystyle{ACM-Reference-Format}
\balance
\bibliography{samples/sample-base}

\clearpage
\nobalance
\appendix

\section*{\huge Appendix}

\startcontents[app]
\section*{Table of Contents}
\label{sec:overview}

\begingroup

\titlecontents{section}
  [2.5em]                 
  {\bfseries}
  {\contentslabel{2.2em}}
  {}
  {\hfill\contentspage}

\titlecontents{subsection}
  [4.5em]              
  {\normalfont}
  {\contentslabel{2.8em}}
  {}
  {\titlerule*[0.5em]{.}\contentspage}

\printcontents[app]{}{1}[2]{}

\endgroup

\section{Dataset and Preprocessing}\label{sec:suppl-data}
\subsection{3D Scene Datasets}
We construct a unified training corpus by aggregating multiple 3D scene understanding datasets built on ScanNet v2~\citep{dai2017scannet}. ScanNet is a seminal and richly annotated dataset of 3D indoor environments, crucial for advancing research in scene understanding. It comprises 1,513 RGB-D video sequences captured across 707 unique indoor spaces, including a diverse range of environments such as apartments, offices, hotels, and classrooms. The raw scans are processed into comprehensive 3D reconstructions, providing textured meshes for each scene. Beyond geometry and appearance, ScanNet is distinguished by its extensive and high-quality annotations: each scene is annotated with 3D camera poses, instance-level semantic segmentation, and axis-aligned bounding boxes for objects. This foundational work provides the essential 3D environmental data upon which many high-level reasoning tasks are built. Our unified corpus leverages several downstream datasets derived from ScanNet, including ScanQA~\citep{azuma2022scanqa}, SQA3D~\citep{ma2022sqa3d}, Scan2Cap~\citep{chen2021scan2cap}, ScanRefer~\citep{chen2020scanrefer}, and Multi3DRefer~\citep{zhang2023multi3drefer}.

\begin{itemize}
\item 
\textbf{ScanQA}~\cite{azuma2022scanqa} focuses on the task of 3D Visual Question Answering (VQA). It pairs ScanNet scenes with natural language questions and provides corresponding answers. ScanQA's questions often require complex reasoning about the 3D spatial relationships, object attributes, and commonsense knowledge within the scene. For example, a question like ``What is placed on top of the wooden desk next to the window?'' requires the model to first locate the window, then find the wooden desk adjacent to it, and finally identify the object on that desk's surface. This dataset is crucial for evaluating our model's ability to jointly comprehend language and 3D geometry.  

\item 
\textbf{SQA3D}~\cite{ma2022sqa3d} presents a more challenging scenario. It combines situational reasoning from embodied navigation with 3D visual reasoning. The dataset provides a base scene, a path an agent has taken within that scene, and a series of spatial questions about the environment from the agent's perspective. For instance, given a path, a question might be ``After you pass the kitchen, what color is the couch you see on your left?'' This requires our model to not only understand the static 3D scene but also to reason about the dynamic perspective changes along a path, making it an excellent testbed for spatial and temporal reasoning.

\item 
\textbf{Scan2Cap}~\cite{chen2021scan2cap} tackles the task of 3D dense captioning. The goal is to generate descriptive natural language captions for individual objects within a 3D scene. Given a scene and a specific object's location, the model must produce a phrase that describes that object, often including its attributes (e.g., color, shape) and its spatial relation to other objects. For example, an object might be captioned as ``a large green sofa placed against the wall facing a wooden coffee table.'' This task directly trains our model to generate accurate and context-aware textual descriptions grounded in 3D objects.

\item 
\textbf{ScanRefer}~\cite{chen2020scanrefer} and \textbf{Multi3DRefer}~\cite{zhang2023multi3drefer} are benchmarks for the 3D visual grounding task, which involves localizing a target object in a 3D scene based on a free-form natural language description. ScanRefer provides a single description for an object, while Multi3DRefer provides multiple diverse descriptions for the same object. This is critical for learning the variability in human language. A referring expression could be ``the tall black chair next to the bookshelf'' or alternatively ``the office chair with a high back near the wall of books.'' Training on these datasets enables our model to precisely associate complex linguistic expressions with their corresponding 3D object instances, a core capability for interactive 3D systems.
\end{itemize}

By integrating these diverse datasets—spanning question answering, situational reasoning, captioning, and grounding—we ensure our model develops a robust and generalizable understanding of the intricate connections between language and 3D visual scenes.

\begin{figure}[t!]
    \centering
    \includegraphics[width=\linewidth]{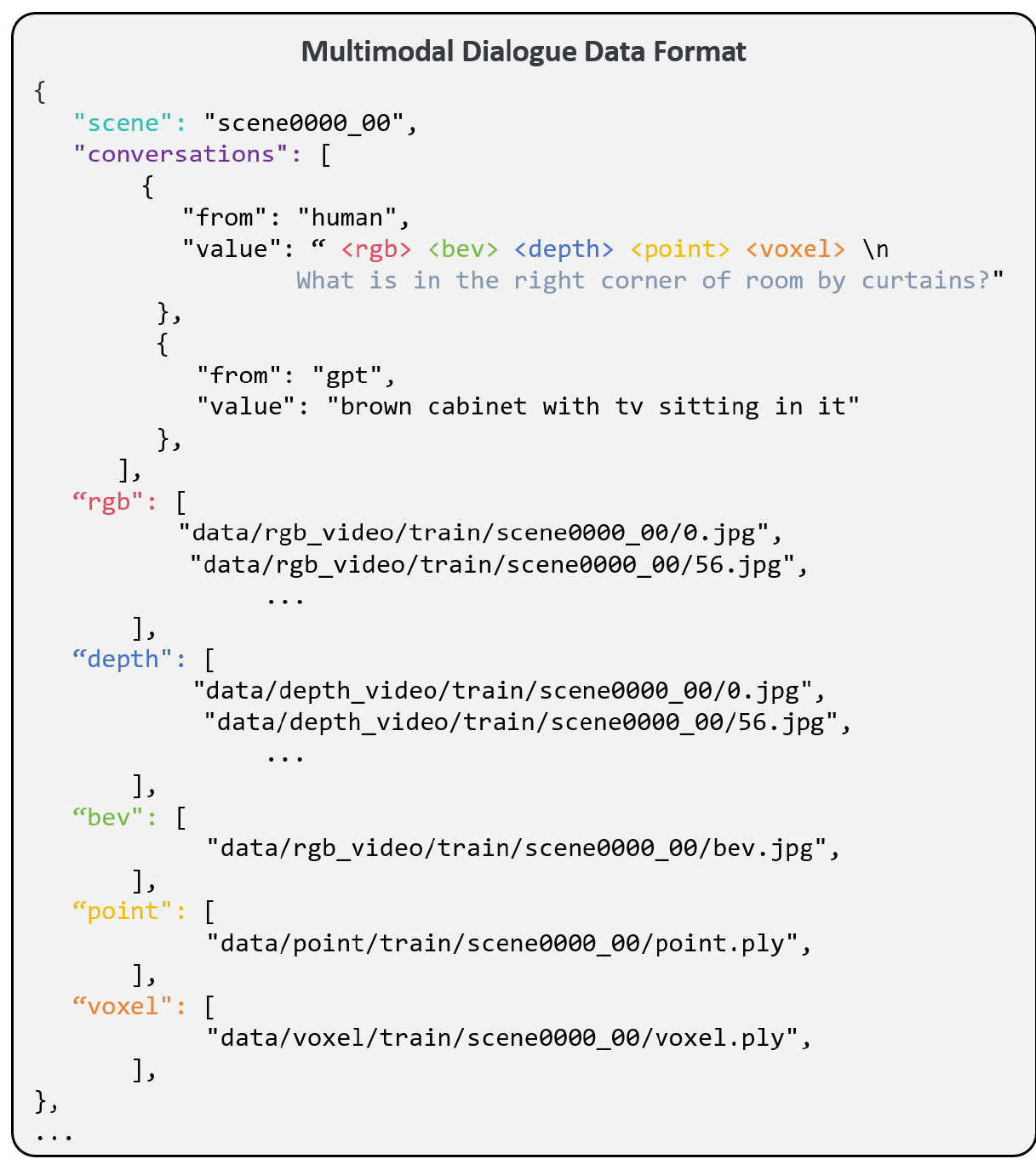}
    \vspace{-2em}
    \caption{Multimodal dialogue data format. Each instance links a dialogue to a 3D scene and provides aligned RGB, BEV, depth, point cloud, and voxel inputs together with human–model conversation turns.}
    \label{fig:dialogue}
\end{figure}

\begin{figure}[t!]
    \centering
    \includegraphics[width=\linewidth]{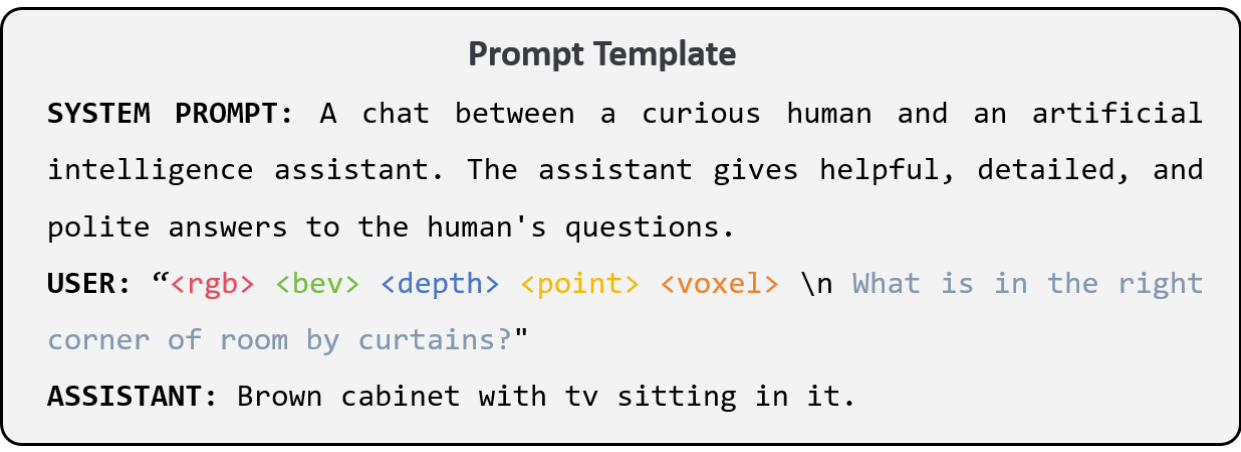}
    \vspace{-2em}
    \caption{Prompt template. A system message defines the conversational role of the assistant, followed by a user query containing modality placeholders and a natural-language question that guides the assistant’s response.}
    \label{fig:prompt}
\end{figure}

\subsection{Dialogue Data and Prompt Templates}

\begin{figure}[ht!]
    \centering
    \includegraphics[width=\linewidth]{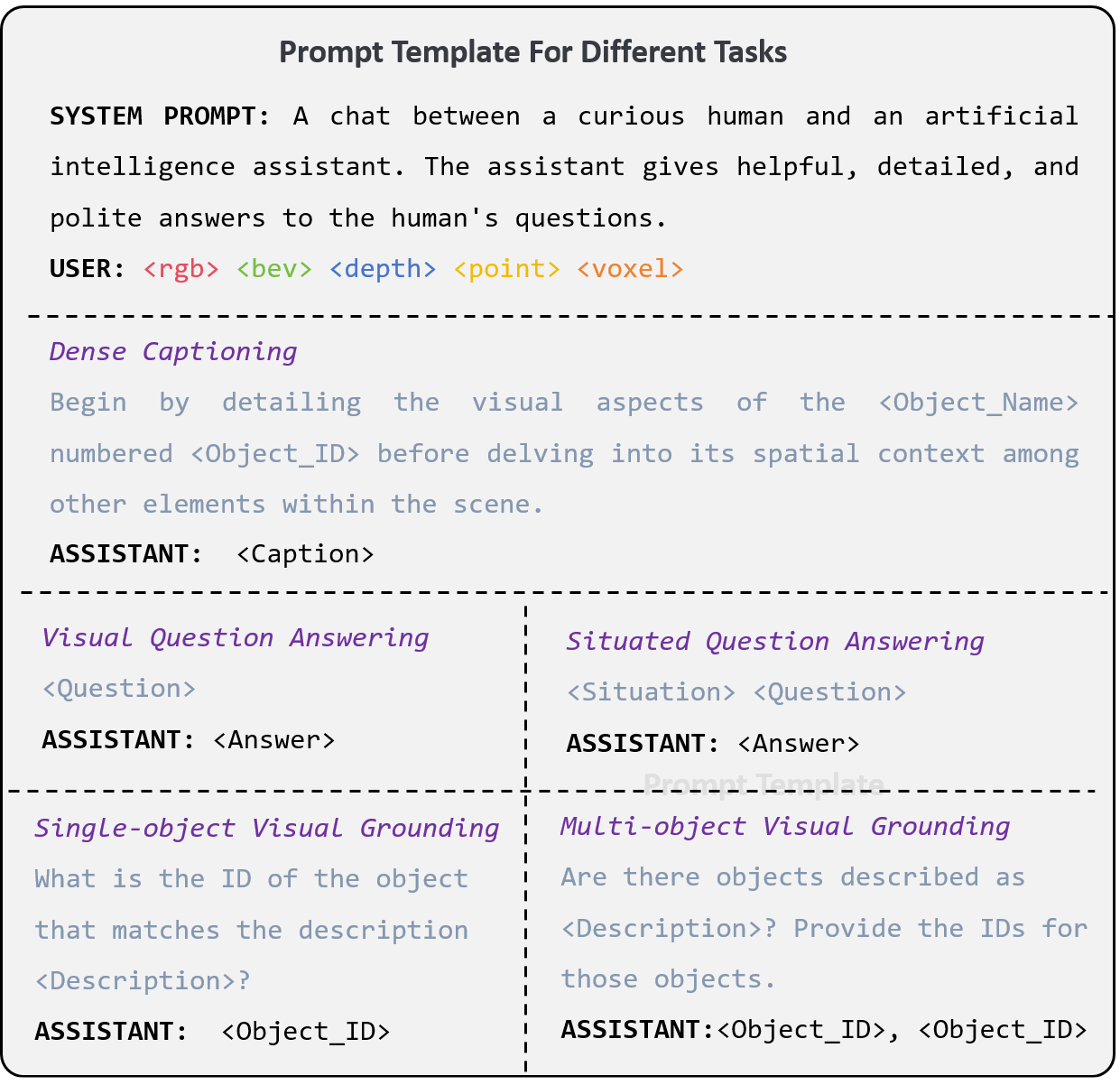}
    \vspace{-2em}
    \caption{Prompt templates for different tasks. Unified prompt structures for dense captioning, visual question answering, and visual grounding tasks, each specifying how the assistant should interpret modality inputs and produce task-specific outputs.}
    \label{fig:prompt-task}
\end{figure}

Figs.~\ref{fig:dialogue}--\ref{fig:prompt-task} illustrate the data organization and prompting approach used for multimodal dialogue tasks.

Fig.~\ref{fig:dialogue} illustrates the structure of our multimodal dialogue data format.
Each dialogue instance is grounded in a specific 3D scene, indicated by the ``scene'' field, which links the conversation to the corresponding ScanNet environment. The ``conversation'' field stores a sequence of interaction turns between the human user and the model. Each turn is annotated with a ``from'' field (either ``human'' or ``gpt'') and a ``value'' containing the natural-language message.
In the first user turn, multimodal placeholders such as ``<rgb>'', ``<bev>'', ``<depth>'', ``<point>'', and ``<voxel>'' explicitly specify which modalities are provided to the model.
The remaining fields list the concrete data sources for each modality—for example, multi-view RGB images, depth maps, BEV renderings, point clouds, and voxel grids—ensuring that every dialogue is consistently grounded in the same set of multimodal observations.
This unified format allows the model to jointly learn instruction following, multimodal reasoning, and 3D scene understanding within a single, coherent representation.

Figs.~\ref{fig:prompt} and~\ref{fig:prompt-task} illustrate the prompt templates used in our multimodal dialogue framework.
Each prompt begins with a system message that defines the behavior of the AI assistant.
The user turn then specifies available modalities using placeholder tokens (e.g., \texttt{<rgb>}, \texttt{<bev>}, \texttt{<depth>}, \texttt{<point>}, \texttt{<voxel>}), followed by the natural-language query.
These tokens indicate which modality features are supplied to the model and serve as routing cues within the multimodal architecture rather than representing raw data directly.

Fig.~\ref{fig:prompt-task} further organizes the prompt formats by task type.
For dense captioning, the prompt instructs the assistant to describe the appearance and spatial context of an object based on its name and instance ID.
For visual question answering, the user provides a natural-language question, and the assistant generates the corresponding answer grounded in the 3D scene.
For visual grounding, the assistant must identify which object instance(s) match the given description and return the corresponding instance ID(s).
The instance IDs referenced in these tasks are obtained from a 3D instance segmentation model such as Mask3D~\citep{schult2023mask3d}, ensuring consistent and scene-aligned object indexing across the dataset.

\subsection{Multi-Modal Preprocessing Pipeline}

\subsubsection{Keyframe Selection Using FoVSR}\label{sec:suppl-fovsr}
For each ScanNet scene we have tens to hundreds of RGB(-D) frames. Feeding all views into the multimodal encoder is computationally prohibitive and introduces large redundancy, since many frames share almost identical fields of view. We therefore select a compact set of keyframes that preserves high 3D coverage of the scene while discarding redundant or low-quality views.

\vspace{0.3em}\noindent\textit{\textbf{FoVSR Algorithm.}}
We propose a fast keyframe selection algorithm, named Field-of-view Sampling with Refinement (FoVSR), as shown in Algorithm~\ref{appendix:alg-enhanced-FoVSR}. 
FoVSR operates in three stages:
\begin{enumerate}
    \item 
    \emph{Voxel pruning and per-view coverage.}
Given the scene voxel set $V$ and camera parameters $\{\Pi_k\}$, we first remove uninformative structural voxels such as floor, ceiling, and wall regions, obtaining a reduced voxel set $V_{\text{scene}}$.
For each candidate view $f_k$, we then compute the subset $V_k \subseteq V_{\text{scene}}$ that lies within its field of view and satisfies the distance threshold $d_{\max}$.

    \item 
\emph{Greedy field-of-view sampling.}
Starting from an empty selected set $S$ and a covered-voxel set $U$, we iteratively choose views that contribute the largest marginal coverage. 
At each step, for every unselected view $f_k \notin S$ we compute $g_k = |V_k \setminus U|$ as the number of newly covered voxels, select $f^* = \arg\max_k g_k$, add it to $S$, and update $U \leftarrow U \cup V_{f^*}$. 
This process repeats until we obtain $K$ keyframes.

    \item 
\emph{Local refinement for view quality.}
 Finally, for each selected index $i$ (corresponding to frame $f_i$ in the original sequence), we consider a small temporal neighborhood 
$N_i = \{f_j \mid j \in [\max(0, i-2), \min(i+2, n-1)]\}$, 
where $n = |F|$ denotes the total number of frames in the sequence. 
For each neighbor $f_j \in N_i$, we compute a sharpness score $s_j$ using the variance of the Laplacian of the image, \eg, 
$s_j = \mathrm{Var}(\nabla^2 I_{f_j})$. 
We then replace $f_i$ with the sharpest frame 
$f_{\text{best}} = \arg\max_{f_j \in N_i} s_j$, 
yielding a final keyframe set $S$ that is both spatially diverse and visually clear.

\end{enumerate}

\begin{algorithm}[t!] 
  \caption{Field-of-view Sampling with Refinement (FoVSR)}
  \label{appendix:alg-enhanced-FoVSR}
  \begin{algorithmic}[1]

    \Require Candidate frames $F=\{f_k\}_{k=1}^{n}$, scene voxel set $V$, camera params $\{\Pi_k\}_{k=1}^{n}$, budget $K$, distance limit $d_{\max}$
\Ensure Selected view set $S$
\Function{Sampling}{$F$, $V$, $\{\Pi_k\}$, $K$, $d_{\max}$}
    
    \Statex \textcolor{Gray}{// 1. Voxel pruning}
    \State $V_{\mathrm{scene}} \gets \{\,v\in V \mid v.\text{type} \notin \{\text{floor}, \text{ceiling}, \text{wall}\} \}$ 
    \For{each view $f_k$}
      \State $V_k \gets \{\,v\in V_{\mathrm{scene}} \mid \text{visible}(\text{proj}(v, \Pi_k))\text{ and } \|X_v - X_{\Pi_k}\| \le d_{\max}\,\}$
    \EndFor
    
    \Statex \textcolor{Gray}{// 2. Perform greedy selection based on marginal coverage}
    \State $S \gets \varnothing$, \quad $U \gets \varnothing$  
    \While{$|S| < K$}
      \For{each view $f_k \notin S$}
        \State $g_k \gets |V_k \setminus U|$
      \EndFor
      \State $f^* \gets \arg\max_k g_k$
      \State $S \gets S \cup \{f^*\}$,\quad $U \gets U \cup V_{f^*}$
    \EndWhile

    \Statex \textcolor{Gray}{// 3. Selected clear views} 
    \For{each index $i$ such that $f_i \in S$}
      \State $N \gets \{\, f_j \mid j \in [\max(0, i{-}2), \min(i{+}2, n{-}1)] \,\}$
      \For{each $f_j \in N$}
        \State $s_j \gets \mathrm{Var}(\nabla^2 I_{f_j})$
      \EndFor
      \State $f_{\text{best}} \gets \arg\max_{f_j \in N} s_j$
      \State $S \gets (S \setminus \{f_i\}) \cup \{f_{\text{best}}\}$
    \EndFor
    \State \Return $S$
    \EndFunction
  \end{algorithmic}
\end{algorithm}

\vspace{0.3em}\noindent\textit{\textbf{Efficiency Analysis.}}
We assess the efficiency of our Field-of-view Sampling with Refinement (FoVSR) algorithm and compare it with the widely used Maximum Coverage Sampling (MC) in Video-3D LLM~\cite{zheng2025video}. This subsection presents both theoretical complexity analysis and empirical runtime comparison to illustrate FoVSR’s computational advantages.

\begin{itemize} 
\item \underline{\textit{Theoretical Analysis.}}
Let $F$ denote the complete frame set with size $|F|$, and $V$ the scene voxel set with size $|V|$.
The MC method repeatedly computes depth-based voxel coverage for all remaining frames during each greedy iteration, leading to a complexity of $\mathcal{O}(K|F||V|)$ for selecting $K$ keyframes.

FoVSR instead first removes uninformative structural voxels to obtain a reduced set $V_{\text{scene}}$, where $|V_{\text{scene}}|\ll|V|$.
Using camera poses, the visible voxel set $V_k$ of each frame is computed once with complexity $\mathcal{O}(|F||V_{\text{scene}}|)$.
Greedy sampling then operates on these precomputed coverage sets, requiring at most $\mathcal{O}(K|F||V_{\text{scene}}|)$, while the fixed-size local refinement introduces only $\mathcal{O}(K)$ additional cost.
Thus, the overall complexity is
$\mathcal{O}(K|F||V_{\text{scene}}|)$, compared with $\mathcal{O}(K|F||V|)$ for MC.
Since $|V_{\text{scene}}|\ll|V|$ and expensive depth-based coverage computation is avoided during iterative selection, FoVSR substantially reduces the practical computational cost.

\item \underline{\textit{Experimental Analysis.}}  
Table~\ref{tab:mc_vs_fovsr} reports the execution time of viewpoint selection using MC and FoVSR under varying numbers of input frames. Both methods operate at a fixed sampling rate of 1 frame per 20. As the number of frames increases, the runtime of MC grows rapidly, whereas FoVSR consistently maintains low and stable execution time. Compared to MC~\cite{zheng2025video}, our algorithm achieves up to 100× speed-up in computing coverage by using camera poses instead of depth maps. This demonstrates the superior efficiency of FoVSR in practice, especially for large-scale scenes.

\end{itemize}

\begin{table*}[t!]
\centering
\caption{Runtime comparison of MC and FoVSR as the number of frames increases across different settings.}
\label{tab:mc_vs_fovsr}
\vspace{-1.0em}
\begin{tabular}{c|cccccccccccccccc}
\toprule
\textbf{Setting} 
& 1 & 2 & 3 & 4 & 5 & 6 & 7 & 8 
& 9 & 10 & 11 & 12 & 13 & 14 & 15 & 16 \\
\midrule
\textbf{Frames} 
& 289 & 577 & 760 & 980 & 1215 & 1285 & 1747 & 1896 
& 1960 & 2155 & 2391 & 2513 & 2674 & 3362 & 4498 & 5336 \\
\textbf{MC (s)} 
& 0.99 & 2.28 & 2.62 & 4.23 & 4.62 & 5.54 & 6.32 & 7.20 
& 7.22 & 8.04 & 9.21 & 8.20 & 10.01 & 10.94 & 15.18 & 16.03 \\
\textbf{FoVSR (s)} 
& 0.02 & 0.09 & 0.04 & 0.07 & 0.10 & 0.13 & 0.09 & 0.12 
& 0.23 & 0.15 & 0.08 & 0.13 & 0.19 & 0.13 & 0.14 & 0.21 \\
\bottomrule
\end{tabular}
\vspace{-1.0em}
\end{table*}

\subsubsection{3D-to-BEV Rendering}
To encode global spatial structure, we generate a BEV representation by rendering the reconstructed 3D mesh from an overhead viewpoint.
We start by converting the mesh into a unified global point set and filtering out ceiling and unstable regions.
A virtual orthographic camera is placed above the scene with a fixed height and orientation. Through a lightweight rendering pass, the full 3D geometry is projected onto the ground plane, producing a BEV image that captures structural boundaries, major surfaces, and object footprints.

We additionally incorporate semantic information by projecting mesh-level instance segmentation onto the BEV plane.
Each object instance produces a 2D region corresponding to its contact area with the ground plane, from which we derive a representative center.
These centers serve as object anchors that are drawn on the BEV map to highlight object locations and facilitate multimodal alignment across views.

This top-down map condenses complex 3D structure into a compact global modality, complementing the fine-grained geometry provided by point clouds and voxels, and enabling the model to reason over long-range spatial context.

\subsubsection{Depth Normalization and Completion}
To ensure consistent geometric inputs across scenes, we preprocess the raw depth maps obtained from the ScanNet sensor pipeline.
Depth values are first normalized to metric scale and clipped to a fixed valid range to suppress sensor noise and unstable far-range estimates.
Missing or invalid regions—often caused by reflective surfaces or occlusions—are filled using a lightweight depth completion module based on bilateral filtering and spatial consistency constraints.
We additionally apply per-frame smoothing to reduce temporal fluctuations across neighboring keyframes selected by FoVSR.
The processed depth maps provide stable geometric cues for depth-aware feature encoding and multimodal reasoning together with the aligned RGB observations.

\subsubsection{Point and Voxel Construction}
We extract point cloud inputs directly from the reconstructed mesh of each ScanNet scene.
We sample 8,192 points from the mesh surface using Farthest Point Sampling (FPS), providing broad spatial coverage of the scene.
Each sampled point includes its 3D coordinate, vertex color, surface normal, and semantic label, providing both geometric detail and appearance cues.

To obtain a complementary volumetric representation, we voxelize the scene using a fixed voxel size of 0.02~m.
Each occupied voxel stores aggregated geometric occupancy and normalized color statistics.
The resulting sparse voxel grid captures the global spatial structure of the environment, while the sampled points preserve fine-grained surface information.

Together, the point cloud and voxel grid provide a dual-granularity representation that supports robust multimodal encoding across different spatial scales in 3D scenes.

\section{Model Architecture}\label{sec:suppl-model}
\subsection{Omni-modal Feature Extractor}
To obtain a comprehensive and geometry-aware representation of indoor 3D scenes, SmartMage incorporates five complementary modalities, including multi-view RGB, depth, BEV, point cloud, and voxel. Each modality captures distinct aspects of the environment: appearance cues from RGB, metric geometry from depth, global spatial layout from BEV, fine-grained 3D structure from point clouds, and volumetric context from voxels. We design a unified extractor–adapter pipeline that (1) encodes heterogeneous inputs with modality-specific backbones, and (2) projects them into a shared 4096-dimensional embedding space for seamless fusion in the language model. Formally, the multimodal feature set is denoted as $\mF=\{\mathbf{f}_{m}\in\mathbb{R}^{N_m \times d_v}\}_{m\in\mathcal{M}}$, where $\mathcal{M}=\{\mathrm{rgb},\mathrm{dpt},\mathrm{bev},\mathrm{pc},\mathrm{vox}\}$.

\vspace{0.3em}\noindent\textit{\textbf{RGB Feature Extractor.}}
RGB frames provide rich appearance and texture cues that are essential for recognizing objects, materials, and attributes. To avoid processing long RGB-D videos, we apply the FoVSR keyframe selection algorithm (detailed in \S~\ref{sec:suppl-fovsr}) to obtain a set of $K$ multi-view images $\mathcal{R}=\{\mathcal{R}_1,\ldots,\mathcal{R}_K\}$ with high coverage and visual clarity. Each frame $\mathcal{R}_k$ is fed into the \texttt{Qwen3-VL-8B-Instruct} vision encoder~\cite{bai2025qwen3}, which divides the image into $N^p_{\text{rgb}}$ patches of size $16\times 16$ and produces patch embeddings $\mZ^{(0)}_{\text{rgb}} \in \mathbb{R}^{N^p_{\text{rgb}} \times d_{\text{rgb}}}$,
where $d_{\text{rgb}}=1152$ is the hidden size of the vision backbone. After 27-layer Transformer blocks, a PatchMerger module aggregates the patches and maps them into the vision–language representation:
\begin{equation}
   \mathbf{f}_{\text{rgb}}=\text{PatchMerger}(\mZ^{(L)}_{\text{rgb}}) \in \mathbb{R}^{N_{\text{rgb}} \times d_v} 
\end{equation}

\vspace{0.3em}\noindent\textit{\textbf{Depth Feature Extractor.}}
Depth maps provide metric geometry complementary to the RGB appearance. 
Given depth video $\mathcal{D}=\{\mathcal{D}_1,\ldots,\mathcal{D}_K\}$, each depth value is back-projected to a 3D coordinate using camera intrinsics.  
We follow~\citep{zhu2025llava} and augment patch embeddings with their corresponding 3D coordinates:
\begin{equation}
\tilde{\mZ}^{(0)}_{\text{dpt},k}
=
\mZ^{(0)}_{\text{rgb},k}
+
\mathrm{PE}_{3\text{D}}(\mathcal{D}_k),
\end{equation}
where $\mZ^{(0)}_{\text{rgb},k}$ denotes the initial patch embeddings of the $k$-th RGB frame, and $\mathrm{PE}_{3\text{D}}(\mathcal{D}_k)$ denotes the geometry-aware positional encoding derived from its aligned depth map.
The enriched tokens are processed by the same \texttt{Qwen3-VL-8B-Instruct}~\cite{bai2025qwen3} vision tower and PatchMerger, yielding $\mathbf{f}_{\text{dpt}} \in \mathbb{R}^{N_{\text{dpt}} \times d_v}$.
These depth-aware embeddings provide explicit spatial information crucial for 3D reasoning.

\vspace{0.3em}\noindent\textit{\textbf{BEV Feature Extractor.}}
Egocentric multi-view images lack global scene coverage. To introduce global structural priors, we render the reconstructed mesh into a semantic BEV map $\mathcal{B} \in \mathbb{R}^{H \times W \times 3}$ using orthographic projection. The BEV contains planar layout, room boundaries, object footprints, and instance-level masks.  
The BEV image is processed by the same \texttt{Qwen3-VL-8B-Instruct}~\cite{bai2025qwen3} vision encoder and PatchMerger, generating $\mathbf{f}_{\text{bev}} \in \mathbb{R}^{N_{\text{bev}} \times d_v}$.
These tokens encode holistic spatial layout and object relationships complementing the egocentric RGB–depth observations.

\vspace{0.3em}\noindent\textit{\textbf{Point Cloud Feature Extractor.}}
The point cloud branch operates on the FPS-sampled point set obtained from the reconstructed mesh, where each point already contains 3D coordinates, color, normals, and semantic labels. 
Let the sampled point cloud be denoted as $\mathcal{P} = \{p_1, p_2, \ldots, p_{N^s_{\text{pc}}}\}
\in \mathbb{R}^{N^{s}_{\text{pc}} \times d^{s}_{\text{pc}}}$, where $N^{s}_{\text{pc}}=8192$ is the number of sampled points and $d^{s}_{\text{pc}}$ is the per-point feature dimension. 
In our implementation, each point includes its 3D coordinates, color, normals, 
and a semantic label, resulting in $d^{s}_{\text{pc}}=10$.
We employ a pre-trained PointNet++~\cite{qi2017pointnet++} backbone to encode local geometric neighborhoods and hierarchical surface patterns, producing point-level features $\mZ_{\text{pc}} \in \mathbb{R}^{N_{\text{pc}} \times d_{\text{pc}}}$, where $d_{\text{pc}}$ is the output dimension of the point encoder (256 in our implementation).  
A modality-specific adapter then maps these features into the unified multimodal embedding space:
\begin{equation}
\mathbf{f}_{\text{pc}}
=
\text{Adapter}_{\text{pc}}(\mZ_{\text{pc}})
\in \mathbb{R}^{N_{\text{pc}} \times d_v},  
\end{equation}
where $d_v=4096$ is the shared vision–language embedding dimension.  
This branch supplies fine-grained geometric cues at the object and surface level, complementing the global layout captured by the 2D modalities and the voxel grid.

\vspace{0.3em}\noindent\textit{\textbf{Voxel Feature Extractor.}}
The voxel branch operates on the sparse voxel representation obtained from the reconstructed scene. 
We denote the voxel-level input feature set as $\mX \in \mathbb{R}^{N^{s}_{\text{vox}} \times d^{s}_{\text{vox}}}$,
where $N^{s}_{\text{vox}}$ is the number of occupied voxels and $d^{s}_{\text{vox}}$ is the per-voxel feature dimension provided by the preprocessed sparse grid.

We employ Mask3D~\citep{schult2023mask3d}, a sparse UNet designed for large-scale 3D scenes, to extract hierarchical volumetric features from $\mX$.  
This produces voxel-level embeddings $\mZ_{\text{vox}} \in \mathbb{R}^{N_{\text{vox}} \times d_{\text{vox}}}$,
where $N_{\text{vox}}$ is the number of voxel tokens after instance/segment aggregation and $d_{\text{vox}}$ is the feature dimension of the Mask3D output.

A modality-specific voxel adapter projects these embeddings into the unified multimodal embedding space:
\begin{equation}
\mathbf{f}_{\text{vox}}
=
\text{Adapter}_{\text{vox}}(\mZ_{\text{vox}})
\in \mathbb{R}^{N_{\text{vox}} \times d_v},  
\end{equation}
where $d_v=4096$ is the shared vision-language embedding dimension across modalities.
This branch provides coarse-to-fine volumetric structure that complements the surface-level point cloud features and global cues from the 2D modalities.

\begin{table*}[t!]
\centering
\caption{Token counts and feature dimensions for each modality.}
\label{tab:token_counts}
\vspace{-1.0em}
\begin{tabular}{lccccc}
\toprule
\textbf{Stage} & \textbf{RGB} & \textbf{Depth} & \textbf{PC} & \textbf{BEV} & \textbf{Voxel} \\
\midrule
Original Input 
& (32, 128, 123, 3)  
& (32, 128, 123, 1) 
& (8192, 10) 
& (128, 123, 3) 
& (16384, 10) \\
\midrule
After Encoder 
& (32$\times$56, 1152) 
& (32$\times$56, 1152) 
& (64, 256) 
& (56, 1152)
& (10k$\sim$200k, 96) \\
\midrule
Aggregation 
& Token Merging 
& Token Merging
& - 
& Token Merging
& Instance-based pooling \\
\midrule
After Agg. 
& (32$\times$14, 1152) 
& (32$\times$14, 1152)
& (64, 256) 
& (15, 1152) 
& (100, 96) \\
\midrule
After Projector 
& (32$\times$15, 4096) 
& (32$\times$15, 4096) 
& (64, 4096) 
& (15, 4096) 
& (100, 4096) \\
\midrule
Token Counts 
& $N_{\text{rgb}} = 480$ 
& $N_{\text{dpt}} = 480$ 
& $N_{\text{pc}} = 64$ 
& $N_{\text{bev}} = 15$ 
& $N_{\text{vox}} = 100$ \\
\bottomrule
\end{tabular}
\end{table*}

\vspace{0.3em}\noindent\textit{\textbf{Unified Multimodal Representation.}}
All modality features are now aligned in the same embedding space $d_v=4096$ and fused with text tokens for multimodal reasoning.  
The omni-modal feature sequence is expressed as $\mF_{\text{uni}}
=
\{\,\mathbf{f}_{txt},\,\mathbf{f}_{\text{rgb}},\,\mathbf{f}_{\text{dpt}},\,\mathbf{f}_{\text{bev}},\,\mathbf{f}_{\text{pc}},\,\mathbf{f}_{\text{vox}}\,\}$,
which provides SmartMage with holistic information spanning local appearance, metric depth, global layout, geometric surfaces, and volumetric context.
This unified representation serves as the foundation for semantic-guided modality routing and MoE reasoning in subsequent modules.

\vspace{0.3em}\noindent\textit{\textbf{Token Count Summary.}}
For completeness, we also summarize the token statistics of all modalities in Table~\ref{tab:token_counts}.  
The table lists the original input shapes, the feature dimensions after the encoder, the aggregation strategies applied to each modality (such as token merging for RGB/Depth/BEV or instance-based pooling for voxels), and the resulting token counts.  
These details make the processing pipeline of each modality more transparent and help readers understand how the unified multimodal representation is formed before entering the later routing and MoE modules.

\subsection{SMART Module} 
The Semantic-guided Modality Adaptive RouTing (SMART) module integrates semantic priors, calibrated text–visual similarity, modality reliability estimation, and RGB evidence gating into a unified routing mechanism. 
Given the instruction tokens, a lightweight text summarizer extracts both a global instruction embedding $\mathbf{f}_{\text{txt}}$ and $M$ modality-specific queries $\{\boldsymbol{\phi}_m\}$. 
The summarizer is implemented as a single multi-head attention layer (4 heads, hidden dimension 4096) applied to the language embeddings of the Qwen3-VL-8B-Instruct~\cite{bai2025qwen3} backbone, using a text-only attention mask to exclude multimodal placeholder tokens. 
These queries are used to pool each modality’s visual tokens through attention, producing modality summaries $\{\mathbf{f}_m\}$ that form a matrix $\mF_{\text{modal}} \in \mathbb{R}^{B\times M\times D}$. 
In parallel, the Modality Quality Evaluator (MQE) estimates modality reliability from feature activation statistics.
For modality $m$, we compute the activation strength $\mu_m$, activation sparsity $\rho_m$, and activation stability $\sigma_m^2$ from its feature norms.
These statistics are concatenated and mapped to a scalar quality score:
\begin{equation}
q_m = \mathbf{w}_q^\top[\mu_m,\rho_m,\sigma_m^2] + b_q,
\end{equation}
where $\mathbf{w}_q$ and $b_q$ are learnable parameters.

The Semantic Prior Estimator (SPE) computes a prior distribution via:
\begin{equation}
\mathbf{p}=\operatorname{Softmax}\!\big(\mW_{\text{prior}}\,\operatorname{LN}(\mathbf{f}_{\text{txt}})/\tau_{\text{prior}}\big),
\end{equation}
with temperature $\tau_{\text{prior}}=0.7$.
The Semantic Similarity Scorer (SSS) projects $\mathbf{f}_{\text{txt}}$ and each $\mathbf{f}_m$ into a shared space using $\mW_t$ and $\mW_m$, applies $\ell_2$ normalization, and computes cosine similarities. 
A per-sample normalization is followed by modality-specific affine calibration:
\begin{equation}
\hat{s}_m = \frac{s_m^{\text{raw}}-\mu}{\sigma+10^{-5}},
\qquad
s_m = \gamma_m \hat{s}_m+\beta_m,
\end{equation}
with $\gamma_m\!=\!1$ and $\beta_m\!=\!0$ at initialization.

The three modality-wise signals (semantic similarity $\mathbf{s}$, semantic prior $\mathbf{p}$, and quality $\mathbf{q}$) are fused into routing logits:
\begin{equation}
    \mathbf{z} = 
\alpha_s \mathbf{s} + \alpha_p \mathbf{p} + \alpha_q \mathbf{q} + \mathbf{b},
\end{equation}
with routing weights $\alpha_s=1.0$, $\alpha_p=0.7$, and $\alpha_q=0.5$.

To further modulate the primary RGB modality, SMART introduces an RGB evidence gate:
\begin{equation}
g_{\text{rgb}}
=
\sigma(w_g s_{\text{rgb}} + b_g),
\qquad
\Delta
=
\lambda_{\text{rgb}}(g_{\text{rgb}}-0.5),
\end{equation}
where $\lambda_{\text{rgb}}=1.0$.
The gate produces a modality-specific logit adjustment
$\boldsymbol{\delta}_{\text{rgb}}\in\mathbb{R}^{M}$:
\begin{equation}
\delta_{\text{rgb},m}
=
\begin{cases}
\Delta, & m=\mathrm{rgb},\\[2pt]
-\dfrac{\Delta}{M-1}, & m\neq\mathrm{rgb},
\end{cases}
\end{equation}
which strengthens RGB when text--RGB alignment is reliable while suppressing it otherwise, with the opposite adjustment distributed across auxiliary modalities.

During training, SMART adopts a Gumbel--Softmax relaxation for differentiable modality selection.
For each modality $m$, we sample
\begin{equation}
\xi_m=-\log\!\big(-\log u_m\big),
\qquad
u_m\sim\operatorname{Uniform}(0,1),
\end{equation}
and compute the routing distribution as
\begin{equation}
\tilde{\pi}_m
=
\frac{
\exp\!\left((z_m+\delta_{\text{rgb},m}+\xi_m)/T\right)
}{
\sum_{j=1}^{M}
\exp\!\left((z_j+\delta_{\text{rgb},j}+\xi_j)/T\right)
},
\end{equation}
where $T=0.5$ is the Gumbel--Softmax temperature.
At inference, the Gumbel noise is removed and deterministic routing is obtained as
\begin{equation}
\pi_m
=
\frac{
\exp\!\left((z_m+\delta_{\text{rgb},m})/T\right)
}{
\sum_{j=1}^{M}
\exp\!\left((z_j+\delta_{\text{rgb},j})/T\right)
}.
\end{equation}
Here, $\mathbf{g}_{\text{rgb}}$ denotes the broadcasted version of the scalar gate $g_{\text{rgb}}$, aligned with the modality dimension of $\mathbf{z}$.
The final auxiliary modalities are selected adaptively: probabilities of non-primary modalities are sorted and the smallest number whose cumulative mass exceeds $\rho=0.8$ is chosen, clamped between $k_{\min}=0$ and $k_{\max}=3$.
This produces a modality keep mask used to perform token replacement and cross-modal token pruning.


\subsection{MAGE Module}
The proposed Modality-Aware Gating Expert (MAGE) module extends the sparse MoE paradigm by explicitly coupling expert routing with modality-aware representations.
While SMART performs input-level modality selection, MAGE injects modality cues into the internal expert routing process of the decoder, encouraging experts to specialize along modality-consistent dimensions and transforming stochastic expert competition into structured multimodal cooperation.
In this section, we provide additional architectural and implementation details of MAGE, including the design of modality-aware expert specialization, the integration with sparse MoE layers, and the concrete loss terms and hyperparameters used in our final model.

\vspace{0.3em}\noindent\textit{\textbf{Modality-aware Expert Specialization.}}
For each token representation $\mathbf{h}_i^{(\ell)}$, the Modality-aware Expert Speculation (MES) module predicts a modality distribution $\mathbf{r}_i \in \Delta^{M-1}$ over $M$ modalities.
The component $r_{i,m}$ represents the probability that token $i$ is associated with modality $m$.
The MES prediction head is implemented as a lightweight linear layer followed by a softmax.
In practice, ground-truth modality labels $\tilde{\mathbf{r}}_i$ are derived from the token type (e.g., RGB, BEV, point cloud, voxel, depth, or text), and the modality attribution loss in regularizes the instructor to align with these labels.
This explicit supervision encourages the hidden states of tokens from the same modality to cluster in a modality-aware subspace.

To link modality cues with expert specialization, MAGE introduces a learnable modality–expert affinity matrix $\mathbf{A} \in \mathbb{R}^{M \times E}$, where $E$ is the number of experts per MoE layer.
The $m$-th row $\mathbf{a}_m^{\top}$ represents the affinity profile of modality $m$ over the expert ensemble.
A modality-conditioned expert prior is derived via a temperature-scaled softmax,
\begin{equation}
\tilde{\pi}_{m,e}
=
\frac{\exp(a_{m,e} / \tau_e)}{\sum_{j=1}^{E} \exp(a_{m,j} / \tau_e)},
\qquad
\tilde{\boldsymbol{\pi}}_{m} \in \Delta^{E-1},
\end{equation}
where $\tau_e$ is the gating temperature.
For a token with modality distribution $\mathbf{r}_i$, its token-level expert prior is the mixture
\begin{equation}
\tilde{\pi}_{i,e} = \sum_{m=1}^{M} r_{i,m}\,\tilde{\pi}_{m,e},
\end{equation}
providing a soft modality-guided preference over experts.
These priors do not replace the intrinsic MoE gate; instead, they serve as structural guidance that regularizes the gate towards modality-consistent activation patterns through the expert calibration loss.
As a result, different experts tend to specialize in complementary, modality-related functions (e.g., RGB-centric appearance reasoning or geometry-centric spatial reasoning).

\vspace{0.3em}\noindent\textit{\textbf{Integration with Sparse MoE Layers.}}
MAGE is implemented on top of Qwen-LLM by replacing the feed-forward network of selected decoder layers with sparse MoE blocks.
In our final configuration, we insert MAGE at layers 8, 12, 16, 20, 24, and 28 of the LLM, following a uniform spacing strategy.
Each selected layer contains $E=8$ experts with top-$k=2$ routing.
The experts are initialized from the pretrained FFNs of the corresponding \texttt{Qwen3-VL-8B-Instruct}~\citep{bai2025qwen3} layers, so that MAGE starts from a meaningful functional decomposition instead of random experts.
The gating network is a linear projection from the token representation $\mathbf{h}_i^{(\ell)}$ to $E$ logits,
\begin{equation}
s_{i,e}^{(\ell)} = \mathbf{w}_e^{(\ell)\top}\mathbf{h}_i^{(\ell)},
\end{equation}
followed by a temperature-scaled softmax
\begin{equation}
\pi_{i,e}^{(\ell)}
=
\frac{\exp(s_{i,e}^{(\ell)} / \tau_e)}{\sum_{j=1}^{E} \exp(s_{i,j}^{(\ell)} / \tau_e)},
\end{equation}
where $\tau_e$ is the same gating temperature as used in the expert prior.
We set $\tau_e = 1.0$ in all experiments.
The top-$k$ experts with the largest $\pi_{i,e}^{(\ell)}$ are selected for each token, and the token is updated by a weighted sum of the selected expert outputs, as in Eq.~(6) of the main paper.

\begin{figure*}[t!]
    \centering
    \includegraphics[width=\linewidth]{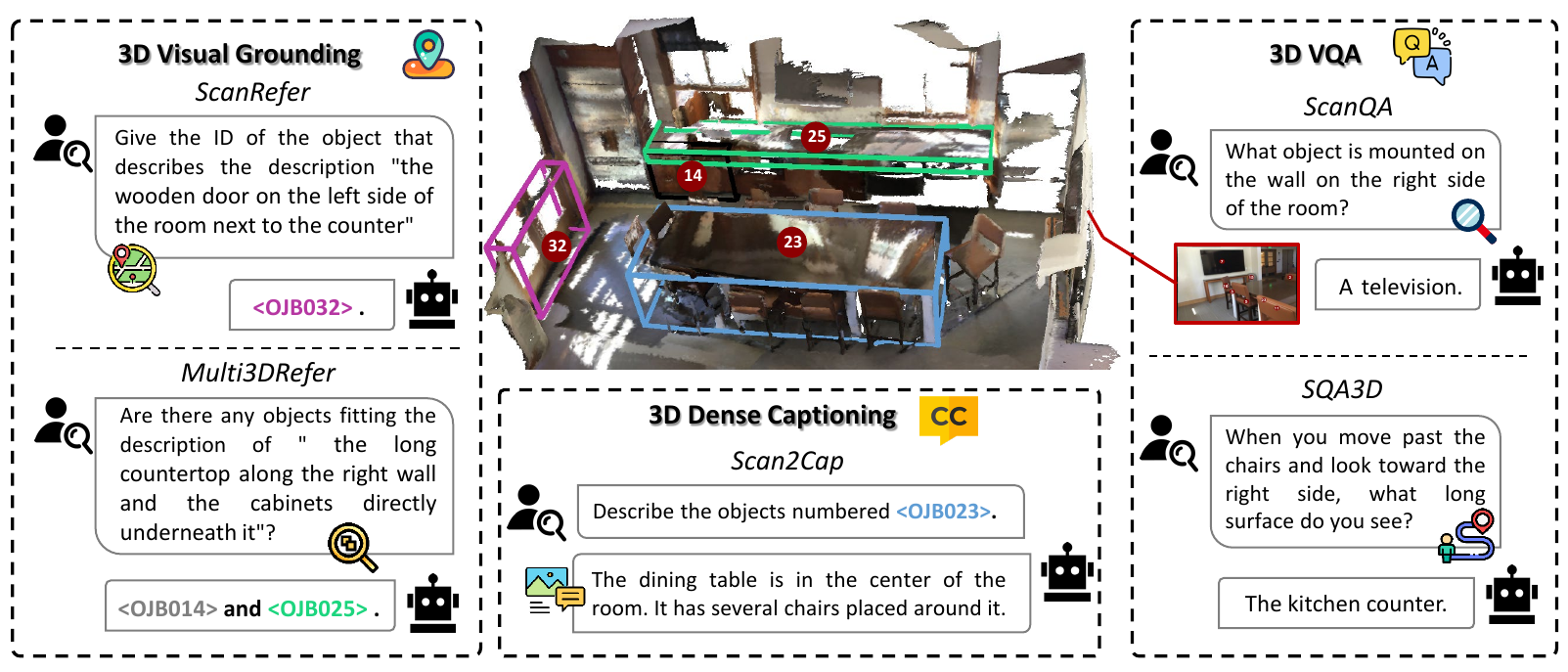}
    \vspace{-2.5em}
    \caption{Examples of 3D scene understanding tasks, including 3D visual grounding (ScanRefer~\cite{chen2020scanrefer}, Multi3DRefer~\cite{zhang2023multi3drefer}), 3D dense captioning (Scan2Cap~\cite{chen2021scan2cap}), and 3D visual question answering (ScanQA~\cite{azuma2022scanqa}, SQA3D~\cite{ma2022sqa3d}).}
    \label{fig:task-overview}
\end{figure*}

\subsection{Loss Design}
\textit{\textbf{Semantic Alignment Loss.}} 
$\mathcal{L}_{\text{SMART}}$ encourages the router to assign higher probabilities to modalities aligned with the textual semantics. It consists of:
(1) semantic correlation loss $\mathcal{L}_{\text{sem}}$, a temperature-scaled contrastive objective encouraging higher semantic similarity between the query embedding and relevant modalities;
and (2) margin discrimination loss~\citep{schroff2015facenet} $\mathcal{L}_{\text{dis}}$, which pushes the gap between relevant and irrelevant modalities. 
\begin{equation}\label{eq:l_smart}
\mathcal{L}_{\text{SMART}} = \lambda_\text{sem}\mathcal{L}_{\text{sem}} + \lambda_\text{dis} \mathcal{L}_{\text{dis}},
\end{equation}
where coefficients $\lambda_\text{sem}$ and $\lambda_\text{dis}$ balance the two terms.
The detailed formulations of $\mathcal{L}_{\text{sem}}$ and $\mathcal{L}_{\text{dis}}$ are:
\begin{equation}\label{eq:l_smart_detail}
\begin{aligned}
\mathcal{L}_{\text{SMART}}
&=
\lambda_{\text{sem}}
\underbrace{
-\mathbb{E}_{b}\!\left[
\frac{1}{|\mathcal{M}_b^{+}|}
\sum_{m\in\mathcal{M}_b^{+}}
\log
\frac{\exp(s_{b,m}/\tau_s)}
{\sum_{j=1}^{M}\exp(s_{b,j}/\tau_s)}
\right]
}_{\mathcal{L}_{\text{sem}}}
\\
&\quad+
\lambda_{\text{dis}}
\underbrace{
\mathbb{E}_{b}\!\left[
\max\bigl(0,\delta_s-(\bar{\ell}_b^{+}-\ell_b^{-})\bigr)
\right]
}_{\mathcal{L}_{\text{dis}}}.
\end{aligned}
\end{equation}
where $\mathcal{M}_b^{+}$ denotes the set of semantically relevant modalities for sample $b$.

\vspace{0.3em}\noindent\textit{\textbf{Expert Assignment Loss.}}
The overall MAGE loss $\mathcal{L}_{\text{MAGE}}$ comprises three terms: the modality attribution loss $\mathcal{L}_{\text{ma}}$, the expert calibration loss $\mathcal{L}_{\text{ec}}$, and the expert balancing loss $\mathcal{L}_{\text{bal}}$.
The modality attribution loss is a cross-entropy objective between the predicted modality distribution $\mathbf{r}_i$ and the ground-truth label $\tilde{\mathbf{r}}_i$, encouraging accurate and discriminative modality assignments at the token level.
The expert calibration loss is a divergence between the intrinsic gating distribution $\pi_{i,e}^{(\ell)}$ and the token-level expert prior $\tilde{\pi}_{i,e}$ produced by the modality–expert affinity matrix, enforcing semantic consistency between modality cues and expert activations.
The expert balancing loss penalizes the deviation of each expert’s average utilization $\pi_e$ from the uniform value $1/E$, preventing expert collapse and promoting more even load distribution.
The expert assignment loss is composed of the above three items:
\begin{equation}
    \mathcal{L}_{\text{MAGE}} = \mathcal{L}_{\text{ma}} +  \mathcal{L}_{\text{ec}} + \lambda_{\text{bal}} \mathcal{L}_{\text{bal}},
\end{equation}
The detailed formulations are:
\begin{equation}\label{eq:l_mage}
\begin{aligned}
\mathcal{L}_{\text{MAGE}}
&=
\,
\underbrace{
\mathbb{E}_{i}\!\Big[
-\!\sum_{m=1}^{M}\tilde{r}_{i,m}\log r_{i,m}
\Big]
}_{\mathcal{L}_{\text{ma}}} \\
&+
\,
\underbrace{
\mathbb{E}_{i}\!\Big[
-\!\sum_{e=1}^{E} \tilde{\pi}_{i,e}\log \pi_{i,e}
\Big]
}_{\mathcal{L}_{\text{ec}}}
+
\lambda_{\text{bal}}\,
\underbrace{
    \mathbb{E}_{e}\!\left[
        \Big( \pi_e - \tfrac{1}{E} \Big)^{2}
    \right]
    }_{\mathcal{L}_{\text{bal}}},
\end{aligned}
\end{equation}
where $r_{i,m}$ is the predicted probability that token $i$ belongs to modality $m$,
$\tilde{r}_{i,m}$ is the ground-truth modality label,
$\pi_{i,e}^{(\ell)}$ is the intrinsic expert gating at layer $\ell$,
$\tilde{\pi}_{i,e}$ is the modality-informed expert prior,
$\pi_e = \mathbb{E}_{i}[\pi_{i,e}]$ is the average activation of expert $e$,
and $E$ the total number of experts.

The total training objective is defined as:
\begin{equation}
    \mathcal{L} = \mathcal{L}_{\text{CE}} +  \mathcal{L}_{\text{SMART}} + \mathcal{L}_{\text{MAGE}}.
\end{equation}

\section{Task Setup}\label{sec:suppl-task}
We systematically evaluate the model's core competencies in 3D scene understanding.
As shown in Fig.~\ref{fig:task-overview}, we consider three representative task families evaluated on five benchmarks, covering question answering, dense captioning, and visual grounding.
We formalize a 3D scene as a multimodal input
\[
\mathcal{S}=
\big(
\{\mathcal{R}_k\}_{k=1}^{K},
\{\mathcal{D}_k\}_{k=1}^{K},
\{\Pi_k\}_{k=1}^{K},
\mathcal{B},
\mathcal{P},
\mathcal{X}
\big),
\]
where $\mathcal{R}_k$ and $\mathcal{D}_k$ denote multi-view RGB and depth observations, $\Pi_k$ denotes the corresponding camera parameters, $\mathcal{B}$ is the BEV map, $\mathcal{P}$ is the point cloud, and $\mathcal{X}$ is the voxel representation.

\subsection{3D Visual Question Answering}
    \textit{\textbf{Objective.}}
    Given a scene $\mathcal{S}$ and a natural-language question $Q$, the model generates an answer $A$:
    \begin{equation}
          f_{\text{VQA}}: (\mathcal{S}, Q) \rightarrow A.
    \end{equation}
    This task evaluates holistic reasoning over objects, attributes, spatial relations, and commonsense.

    \vspace{0.3em}\noindent\textit{\textbf{Task Benchmarks.}}
    ScanQA~\cite{azuma2022scanqa} pairs ScanNet scenes with natural-language questions that demand complex reasoning over 3D spatial relations, object attributes, and commonsense cues (e.g., localizing a wooden desk next to a window and identifying the item on top). 
    SQA3D~\cite{ma2022sqa3d} further introduces sequential, egocentric observations, where questions rely on temporal transitions along an agent’s path, requiring memory of viewpoint changes and dynamic spatial understanding.

\subsection{3D Dense Captioning}
    \vspace{0.3em}\noindent\textit{\textbf{Objective.}}
    Given a scene $\mathcal{S}$, the model detects objects and outputs a caption ( $C_i$ ) for each instance:
    \begin{equation}
        f_{\text{DC}}: \mathcal{S} \rightarrow {(B_i, C_i)}_{i=1}^{N},    
    \end{equation}
    where $B_i$ is a 3D bounding box and $C_i$ is description.   
    This task measures instance-level perception and free-form language generation.

    \vspace{0.3em}\noindent\textit{\textbf{Task Benchmarks.}} 
    Scan2Cap~\cite{chen2021scan2cap} formulates 3D dense captioning by requiring the model to generate natural-language descriptions for individual objects in a scene, incorporating category, appearance, material, and spatial relations (\eg, ``a large green sofa placed against the wall facing a wooden coffee table'').

    \vspace{0.3em}\noindent\textit{\textbf{Proposal-Based Captioning.}}
    For dense caption generation, we employ a proposal-based formulation consistent with prior work.
    We run Mask3D on each scene to obtain instance-level segmentations, where each detected object is associated with a 3D bounding box $\mathbf{b}_i$ and semantic label $c_i$.
    For each instance, we extract geometric descriptors and visual features to form an object embedding $\mathbf{v}_i$.
    The model then conditions on $\mathbf{v}_i$ to produce a free-form description, while optionally incorporating broader scene context to enrich spatial and relational cues.
    This instance-conditioned decoding allows the model to describe objects individually while maintaining consistency with the global 3D structure.

\subsection{3D Visual Grounding}
    \vspace{0.3em}\noindent\textit{\textbf{Objective.}}
    Given a scene $\mathcal{S}$ and a referring expression $E$, the model identifies the target object set:
    \begin{equation}
    f_{\text{VG}}:(\mathcal{S},E)\rightarrow\mathcal{B}_{\text{target}},
    \end{equation}
    where $\mathcal{B}_{\text{target}}$ may contain zero, one, or multiple target objects depending on the grounding setting.

    \vspace{0.3em}\noindent\textit{\textbf{Task Benchmarks.}}
    ScanRefer~\cite{chen2020scanrefer} localizes the target object in a 3D scene based on a free-form referring expression, linking fine-grained linguistic cues to geometric structure. 
    Multi3DRefer~\cite{zhang2023multi3drefer} extends this setting to multi-object referring, requiring the model to jointly ground multiple described instances and resolve ambiguities among similar or closely arranged objects.

    \vspace{0.3em}\noindent\textit{\textbf{Proposal-based Grounding.}}
    Instead of regressing 3D bounding boxes directly, we apply Mask3D~\cite{schult2023mask3d} to each scene to obtain detected all object instances:
    \begin{equation}
            \mathcal{O}_{\text{proposal}} = \{(\mathbf{b}_i, s_i)\}_{i=1}^{N},
    \end{equation}
    where $\mathbf{b}_i$ is the predicted 3D box of the $i$-th instance and $s_i$ its semantic label.
    These proposals are aligned with the dataset annotations, allowing the referred object to be represented as a ground-truth index within $\mathcal{O}_{proposal}$.
    For grounding, each proposal is encoded into a geometric–visual embedding $\mathbf{v}_i$, while the referring expression $E$ is encoded into a text embedding $\mathbf{t}$. 
    We then compute a joint representation:
    \begin{equation}
    \mathbf{h}_i = \phi([\mathbf{t}; \mathbf{v}_i]),
   \end{equation}
    where $\phi(\cdot)$ denotes our multimodal fusion module.
    The model outputs a matching score for every proposal, and the proposal with the highest score is selected as the predicted referred object; its corresponding Mask3D bounding box serves as the final grounded 3D localization.

Together, these three tasks span global scene comprehension (3D-VQA), instance-level caption generation (3D Dense Captioning), and precise object localization (3D Visual Grounding).
Evaluating across this spectrum enables a comprehensive assessment of the proposed model’s capability in multimodal 3D understanding, covering reasoning, generation, and grounding under diverse input conditions and reasoning complexities.

\begin{table}[t!]
\centering
\caption{Category distribution of the ScanFacet benchmark.}
\label{tab:scanfacet-counts}
\vspace{-1.0em}
\resizebox{\linewidth}{!}{
\begin{tabular}{lrrrrrr}
\toprule
\multirow{2}{*}{\textbf{Category}} & 
\multicolumn{2}{c}{\textbf{ScanQA}} &
\multicolumn{2}{c}{\textbf{SQA3D}} &
\multicolumn{2}{c}{\textbf{ScanFacet}} \\
\cmidrule(lr){2-3} \cmidrule(lr){4-5} \cmidrule(lr){6-7}
 & Count & \% & Count & \% & Count & \% \\
\midrule
Color            & 862  & 18.44 & 263  & 8.07  & 1125 & 14.18 \\
Location         & 1170 & 25.03 & 425  & 13.03 & 1595 & 20.10 \\
Material         & 24   & 0.51  & 31   & 0.95  & 55   & 0.69  \\
Number           & 226  & 4.83  & 728  & 22.32 & 954  & 12.02 \\
Object shape     & 140  & 2.99  & 199  & 6.10  & 339  & 4.27  \\
Object type      & 791  & 16.92 & 66   & 2.02  & 857  & 10.80 \\
Spatial relation & 1178 & 25.20 & 1373 & 42.10 & 2551 & 32.14 \\
Other            & 284  & 6.07  & 176  & 5.40  & 460  & 5.80  \\
\rowcolor{gray!15!white}
\textbf{Total}   & \textbf{4675} & \textbf{100} &
                   \textbf{3261} & \textbf{100} &
                   \textbf{7936} & \textbf{100} \\
\bottomrule
\end{tabular}
}
\vspace{-1.5em}
\end{table}

\section{Evaluation Protocol}\label{sec:suppl-eval}
\subsection{ScanFacet Benchmark}
\label{sec:scanfacet-details}
To enable fine-grained analysis of modality--semantics interactions, we construct a diagnostic benchmark named \texttt{ScanFacet}. 
ScanFacet restructures the question--answer pairs from ScanQA~\citep{azuma2022scanqa} and SQA3D~\citep{ma2022sqa3d} into eight semantic categories:
\emph{color}, \emph{location}, \emph{material}, \emph{number}, \emph{object shape}, \emph{object type}, \emph{spatial relation}, and \emph{other}. 
Below we describe the full construction pipeline and dataset composition.

\vspace{0.3em}\noindent\textit{\textbf{LLM-assisted Semantic Taxonomy Pipeline.}}
Each question is categorized using a three-stage LLM-driven procedure:
(1)~\textit{Semantic parsing} extracts the core intent by removing scene-dependent phrasing and normalizing synonyms (e.g., ``how many chairs'' and ``number of chairs'' are mapped to a unified representation);
(2)~\textit{Intent normalization} maps parsed expressions to one of the eight predefined semantic types using rule-based templates and LLM scoring;
(3)~\textit{Self-consistency filtering} applies majority voting across multiple LLM passes to improve robustness. 
All auto-generated labels are further reviewed by a human annotator with minimal corrections (less than 3\% of samples), ensuring the reliability of semantic categories while keeping the process scalable.

\vspace{0.3em}\noindent\textit{\textbf{Dataset Composition.}}
Table~\ref{tab:scanfacet-counts} summarizes the final distribution of semantic categories across the two source datasets.
ScanQA contributes more samples to \textit{color}, \textit{location}, and \textit{object type}, while SQA3D provides a larger proportion of \textit{number} and \textit{spatial relation} questions.  
The combination yields a semantically diverse benchmark suitable for evaluating modality preferences.

\vspace{0.3em}\noindent\textit{\textbf{Benchmark Characteristics.}}
ScanFacet offers:
(1)~fine-grained semantic grouping aligned with human reasoning patterns;
(2)~rich coverage of both attribute-centric queries (\eg color) and relation-centric queries (\eg spatial relation);
(3)~a unified interface that allows detailed analysis of modality contributions across semantic types.
ScanFacet is designed not as another training dataset, but as a controlled diagnostic tool to reveal modality–semantic dependencies in multimodal 3D understanding models.

\subsection{Evaluation Metric}\label{sec:suppl-eval-metric}
Following previous work~\citep{zhu2025llava, huang2024chat, wang2025ross3d}, we comprehensively evaluate our method using standard metrics across multiple tasks in 3D scene understanding. Specifically:
\begin{itemize}
    \item 
    For the Scan2Cap~\citep{chen2021scan2cap} task, we assess the quality of generated scene descriptions using widely adopted captioning metrics, including BLEU-4, METEOR, ROUGE, and CIDEr, computed specifically at Intersection-over-Union (IoU) thresholds of 0.25 and 0.5 between predicted and ground-truth bounding boxes.
    \item 
    For the ScanQA~\citep{azuma2022scanqa} question-answering task, besides captioning metrics, we utilize metrics tailored for answer accuracy and completeness: Exact Match accuracy (EM@1) measures strict correctness of top-1 answers, Relaxed Exact Match (EM-R@1) allows minor acceptable variations, and F1 scores evaluate token-level overlaps.
    \item 
    For referring expression grounding tasks, ScanRefer~\citep{chen2020scanrefer} and Multi3DRefer~\citep{zhang2023multi3drefer}, we evaluate localization accuracy of predicted bounding boxes against ground-truth annotations. Specifically, we report accuracy (Acc@0.25, Acc@0.5) at IoU thresholds of 0.25 and 0.5 for ScanRefer, and F1 scores (F1@0.25, F1@0.5) at the same IoU thresholds for Multi3DRefer.
\end{itemize}

The metrics can be grouped into three categories: text similarity metrics (BLEU, METEOR, ROUGE, CIDEr) for assessing the quality and fluency of generated descriptions; accuracy metrics (EM@1, EM-R@1, F1) for evaluating exactness and completeness in question-answering tasks; and spatial localization metrics (Acc@IoU, F1@IoU, captioning metrics at IoU thresholds) to assess bounding-box predictions in scene grounding tasks.

\subsection{Compared Baselines}\label{sec:suppl-eval-comparison}
We compare SmartMage against three groups of state-of-the-art baselines categorized by their primary modality dependencies. This organization reflects how existing approaches utilize appearance cues, geometric structure, or heterogeneous multimodal signals.

\vspace{0.3em}\noindent\textit{\textbf{RGB-based Models.}}
These methods rely mainly on RGB inputs without explicit 3D geometry.
They include strong 2D vision–language systems such as InternVL2-8B~\citep{lu2025internvl}, Qwen2-VL-7B~\citep{wang2024qwen2}, LLaVA-Video~\citep{zhang2024video}, and LEO~\citep{huang2024embodied}.
While effective for appearance reasoning, their lack of 3D structural cues limits spatial understanding.

\vspace{0.3em}\noindent\textit{\textbf{Geometry-based Models.}}
These systems operate primarily on point clouds or mesh-derived geometry, sometimes with colors or normals.
These methods include 3D-VLP~\citep{jin2023context}, 3D-VisTA~\citep{zhu20233d}, ScanRefer~\citep{chen2020scanrefer}, MVT~\citep{chen2021mvt}, 3DVG-Trans~\citep{zhao20213dvg}, ViL3DRel~\citep{chen2022language}, Scan2Cap~\citep{chen2021scan2cap}, 3DJCG~\citep{cai20223djcg}, Vote2Cap-DETR~\citep{chen2023end}, X-Trans2Cap~\citep{yuan2022x} and M3DRef-CLIP~\citep{zhang2023multi3drefer}.
Although strong in geometric reasoning, they lack complementary appearance cues and multimodal synergy.

\vspace{0.3em}\noindent\textit{\textbf{Multimodal 3D Models.}}
This category integrates heterogeneous modalities such as RGB, depth, point clouds, and voxels.
Representative methods include PQ3D~\citep{zhu2024unifying}, LAMM~\citep{yin2023lamm}, 3D-LLM~\citep{hong20233d}, Chat-3D~\citep{wang2023chat}, Chat-3D V2~\citep{huang2023chat}, LL3DA~\citep{chen2024ll3da}, GPT4Scene~\citep{qi2025gpt4scene}, LLaVA-3D~\citep{zhu2025llava}, Scene-LLM~\citep{fu2024scene}, Chat-Scene~\citep{huang2024chat}, and Grounded 3D-LLM~\citep{chen2024grounded}. 
While these approaches benefit from richer inputs, they typically rely on fixed modality fusion, lacking the semantic-adaptive routing and expert specialization introduced in SmartMage.

\section{Additional Results}\label{suppl-results}

\subsection{More Quantitative Comparisons}\label{sec:suppl-results-quantitative}

\begin{table*}[t!]
        \centering
        \small
        \caption{Evaluation results of 3D question answering across different question types on the test set of SQA3D~\citep{ma2022sqa3d}. $\dagger$ marks the results obtained without high-resolution settings. The best performance is highlighted in \textbf{bold}.}
        \vspace{-1.0em}
        \begin{tabular}{lccccccccc}
            \toprule
                \multirow{2}{*}{\textbf{Method}} & \multicolumn{6}{c}{\textbf{Question Type}}  & \multicolumn{2}{c}{\textbf{Total}}\\
                 \cmidrule(lr){2-7}  \cmidrule(lr){8-9}
               &  What & Is & How & Can & Which & Others & EM@1 & EM-R@1 \\
            \midrule
     \rowcolor{gray!15!white}\multicolumn{9}{c}{\textit{Task-specific Models}} \\
    SQA3D~\citep{ma2022sqa3d} & 31.6  & 63.8  & 46.0  &69.5 &43.9  & 45.3 &46.6 & - \\
    3D-VisTA~\citep{zhu20233d} & 34.8 & 63.3  & 45.4 & 69.8  & 47.2  & 48.1  &48.5 & - \\
    ClipBERT~\citep{lei2021less} &  30.2 & 60.1 & 38.7 & 63.3 & 42.5 & 42.7 & 43.3 & – \\
    \midrule
     \rowcolor{gray!15!white}\multicolumn{9}{c}{\textit{2D MLLMs}} \\
    InternVL2-8B~\citep{lu2025internvl} &  30.5  & 53.8 & 5.5 & 47.3  & 25.8  & 36.3 & 33.0 & 45.3  \\
    Qwen2-VL-7B~\citep{wang2024qwen2} & 29.0 & 59.2 & 33.4 & 50.5 & 44.2 & 43.2 & 40.7 & 46.7 \\
    LLaVA-Video-7B~\citep{zhang2024video} & 42.7 & 56.3 & 47.5 & 55.3 & 50.1 & 47.2 & 48.5 & – \\
    \midrule
     \rowcolor{gray!15!white}\multicolumn{9}{c}{\textit{3D MLLMs}} \\
    LEO~\citep{huang2024embodied} & – & – & – & – & – & – & 50.0 & 52.4 \\
    Scene-LLM~\citep{fu2024scene} & 40.9 & 69.1 & 45.0 & 70.8 & 47.2 & 52.3 & 54.2 & – \\
    ChatScene~\citep{huang2024chat}  & 45.4 & 67.0 & 52.0 & 69.5 & 49.9 & 55.0 & 54.6 & 57.5 \\
    LLaVA-3D~\citep{zhu2025llava} & – & – & – & –  &– &– & 55.6 & – \\
    GPT4Scene$^{\dagger}$ ~\citep{qi2025gpt4scene} & 50.7 & 70.9 & 48.0 & 70.5 & 52.9 & 59.3 & - & 60.7 \\ 

    Video-3D LLM~\cite{zheng2025video} & 50.0 & 70.7 & 57.9 & 69.8 & 50.1 & 55.8 & 57.7 & - \\
    Ross3D~\cite{wang2025ross3d} & 56.0 & 79.8 & 60.6 & 70.4 & 55.3 & 60.1 & 63.0 & 65.7 \\
    \rowcolor{gray!15!white}
    SmartMage (ours) & \textbf{59.4} & \textbf{82.1} & \textbf{66.8} & \textbf{71.5} & \textbf{60.6} & \textbf{60.2}  & \textbf{66.8} & \textbf{71.8} \\

 \bottomrule
    \end{tabular}
    \label{tab:appendix-qt-sqa3d}
    \vspace{-0.5em}
\end{table*}

\textit{\textbf{3D Visual Question Answering.}}
Table~\ref{tab:appendix-qt-sqa3d} provides a comprehensive evaluation of various models on the SQA3D benchmark across different 3D question-answering tasks. The tasks are categorized by question types, including ``What'', ``Is'', ``How'', ``Can'', ``Which'', and ``Others'', alongside aggregated metrics of Exact Match accuracy (EM@1) and Relaxed Exact Match accuracy (EM-R@1). 
SmartMage demonstrates superior performance across all six question types, achieving 59.4 on ``What'', 82.1 on ``Is'', 66.8 on ``How'', 71.5 on ``Can'', 60.6 on ``Which'', and 60.2 on ``Others''. Our model further achieves 66.8 EM@1 and 71.8 EM-R@1, outperforming prior state-of-the-art methods.
These results suggest that SmartMage possesses strong cross-type generalization and robust semantic reasoning capabilities, effectively handling diverse question forms that require appearance understanding, spatial reasoning, object-centric inference, and commonsense grounding. The consistent improvements across question categories further validate the benefit of semantic-adaptive modality selection and modality-aware expert specialization in complex 3D VQA scenarios.

Table~\ref{tab:scanfacet-f1-cider} presents a comparison across seven fine-grained semantic categories on the ScanFacet benchmark, which measures how different modality configurations contribute to semantic reasoning. Across all categories, our semantic-adaptive multimodal routing consistently achieves the highest F1 and CIDEr scores, outperforming single-modality inputs and static fusion. 
Improvements are observed across both appearance- and geometry-related categories: color and material benefit from RGB-centric combinations, while shape and spatial reasoning benefit more from geometry-aware modalities.

These consistent gains demonstrate that SmartMage improves both categorical accuracy and semantic alignment. While fixed fusion is stronger than using any single modality, it still introduces semantic-irrelevant redundancy. In contrast, our adaptive approach composes the most meaningful modality subset per query, achieving robust modality–semantic matching. This fine-grained advantage highlights the effectiveness of semantic-adaptive routing and its strong generalization across diverse reasoning skills in 3D scene understanding.

\begin{table*}[t!]
\centering
\small
\caption{Comparison on the ScanFacet benchmark across different modality configurations, where RGB uses image-only inputs; RGB+DPT adds monocular depth; RGB+PC incorporates point clouds; RGB+VOX uses voxelized geometry; RGB+BEV includes BEV-rendered views; All Modal. (fixed) fuses all modalities with a static fusion pipeline; and All Modal. (ours) applies our semantic-adaptive multimodal routing.}
\vspace{-1.0em}
\label{tab:scanfacet-f1-cider}
\resizebox{\textwidth}{!}{
\begin{tabular}{l*{14}{c}}
\toprule
\multirow{2}{*}{\textbf{Method}} &
\multicolumn{2}{c}{\textbf{Color}} &
\multicolumn{2}{c}{\textbf{Location}} &
\multicolumn{2}{c}{\textbf{Material}} &
\multicolumn{2}{c}{\textbf{Number}} &
\multicolumn{2}{c}{\textbf{Shape}} &
\multicolumn{2}{c}{\textbf{Type}} &
\multicolumn{2}{c}{\textbf{Spatial}} \\
\cmidrule(lr){2-3}
\cmidrule(lr){4-5}
\cmidrule(lr){6-7}
\cmidrule(lr){8-9}
\cmidrule(lr){10-11}
\cmidrule(lr){12-13}
\cmidrule(lr){14-15}
& F1 & CIDEr & F1 & CIDEr & F1 & CIDEr & F1 & CIDEr & F1 & CIDEr & F1 & CIDEr & F1 & CIDEr \\
\midrule
RGB               & 64.7 & 100.0 & 51.3 &  60.0 & 60.8 &  85.0 & 62.5 &  88.0 & 56.4 &  75.0 & 60.2 &  90.0 & 54.9 &  66.0 \\
RGB+DPT           & 59.3 &  80.0 & 54.8 &  68.0 & 62.1 &  88.0 & 57.6 &  78.0 & 55.2 &  72.0 & 58.7 &  85.0 & 57.4 &  77.0 \\
RGB+PC            & 63.1 &  90.0 & 56.9 &  70.0 & 60.3 &  86.0 & 63.1 &  88.0 & 60.5 &  88.0 & 58.2 &  81.0 & 58.9 &  76.0 \\
RGB+VOX           & 61.8 &  85.0 & 50.7 &  59.0 & 58.3 &  77.0 & 63.8 &  92.0 & 57.1 &  77.0 & 58.9 &  82.0 & 55.7 &  69.0 \\
RGB+BEV           & 66.9 & 107.0 & 59.2 &  78.0 & 60.4 &  84.0 & 56.8 &  76.0 & 55.7 &  70.0 & 57.5 &  79.0 & 58.1 &  81.0 \\
All Modal. (fixed) & 65.4 &  99.0 & 61.7 &  83.0 & 66.3 &  99.0 & 64.2 &  89.0 & 66.8 &  99.0 & 64.1 &  95.0 & 60.5 &  82.0 \\
\rowcolor{gray!15!white}
All Modal. (ours)
& \textbf{70.8} & \textbf{115.9}
& \textbf{64.9} & \textbf{87.0}
& \textbf{71.6} & \textbf{112.1}
& \textbf{68.7} & \textbf{99.0}
& \textbf{69.3} & \textbf{104.0}
& \textbf{69.5} & \textbf{108.0}
& \textbf{63.4} & \textbf{86.0} \\
\bottomrule
\end{tabular}
}
\end{table*}

\vspace{0.3em}\noindent\textit{\textbf{3D Visual Grounding.}}
Table~\ref{tab:appendix-scanrefer+multi3drefer} summarizes evaluation results for 3D visual grounding tasks on ScanRefer~\citep{chen2020scanrefer} and Multi3DRefer~\citep{zhang2023multi3drefer}, comparing task-specific models and general 3D Large Language Models (3D LLMs). Our method achieves state-of-the-art results, outperforming existing approaches on both benchmarks. 
Specifically, our model achieves 65.9\% and 59.5\% Acc@0.25/0.5 on ScanRefer and 65.4\% and 60.7\% F1@0.25/0.5 on Multi3DRefer, consistently outperforming prior methods across both benchmarks.
These gains highlight SmartMage’s ability to precisely localize target objects in complex indoor scenes, even when descriptions involve subtle appearance cues, fine-grained geometry, or intricate spatial relations. The consistent improvements across both single-reference (ScanRefer) and multi-reference (Multi3DRefer) settings further validate the effectiveness of our semantic-adaptive modality routing, which enables the model to select the most informative modalities for grounding under varying linguistic instructions.


Table~\ref{tab:appendix-scanrefer-detail} presents a comprehensive comparison between our method and other state-of-the-art approaches on the ScanRefer~\citep{chen2020scanrefer}.
Performance is assessed separately across ``Unique'', ``Multiple'', and combined ``Overall'' subsets. The ``Unique'' subset involves unambiguous samples, each with only a single instance per object category, whereas the ``Multiple'' subset includes ambiguous samples containing multiple instances from the same category. Metrics used are accuracy measured at IoU thresholds of 0.25 and 0.5.
The proposed method achieves superior performance, especially in handling ambiguous cases within the ``Multiple'' subset, obtaining promising accuracy scores at 56.7\% (Acc@0.25) and 53.5\% (Acc@0.5). It also demonstrates outstanding overall capabilities, achieving state-of-the-art results on the ``Overall'' subset with accuracies of 65.9\% and 59.5\%, closely surpassing the previously best-performing model GPT4Scene-HDM~\cite{qi2025gpt4scene}. 
In the ``Unique'' subset, SmartMage remains highly competitive, achieving 89.6\% Acc@0.25 and the best 84.5\% Acc@0.5, reflecting strong capability in handling clear, well-defined visual grounding scenarios.
Overall, these results highlight SmartMage’s robustness across both clear and ambiguous grounding conditions and demonstrate the advantage of semantic-adaptive modality routing in capturing subtle object distinctions and resolving referential ambiguity in complex 3D scenes.

Table~\ref{tab:appendix-multi3drefer-detail} illustrates the comprehensive evaluation results for 3D visual grounding performance on the Multi3DRefer~\citep{zhang2023multi3drefer} across five distinct scenarios: Zero Target without Distractors (ZT w/o D), Zero Target with Distractors (ZT w/ D), Single Target without Distractors (ST w/o D), Single Target with Distractors (ST w/ D), and Multi-Target (MT). Performance is assessed through F1 scores at IoU thresholds of 0.25 and 0.5, emphasizing precision in object localization under varying complexity and distractor presence conditions.
The proposed approach demonstrates competitive performance across multiple scenarios, achieving notable results especially in scenarios involving distractors. 
For the challenging Single Target with Distractors (ST w/ D) setting, SmartMage achieves the best F1@0.25 of 60.0 and ties for the best F1@0.5 of 55.1.
Similarly, in the comprehensive evaluation across all scenarios (denoted ``ALL''), our method attains leading performance (F1@0.25: 65.4, F1@0.5: 60.7), indicating its broad effectiveness in diverse grounding contexts.
Task-specific methods, such as M3DRef-CLIP~\citep{zhang2023multi3drefer} and 3DJCG (Grounding)~\citep{cai20223djcg}, exhibit strong performance in simpler settings (e.g., ZT w/o D and ST w/o D), though their results show noticeable declines when encountering scenarios with distractors or multiple targets. 
In contrast, the proposed approach demonstrates enhanced robustness and flexibility in addressing increased task complexity. 
This observation suggests that our semantic-adaptive modality routing and expert specialization mechanisms enable the model to better handle ambiguous descriptions, reduce distractor interference, and maintain stable grounding performance even in heavily cluttered or multi-target environments.

\begin{table}[t!]
    \centering
    \caption{Evaluation results of 3D visual grounding on ScanRefer~\citep{chen2020scanrefer} and Multi3DRefer~\citep{zhang2023multi3drefer}. $\dagger$ marks results obtained without high-resolution.}
    \vspace{-1.0em}
    \resizebox{\linewidth}{!}{
    \begin{tabular}{ccccccccc}
    \toprule
     \multirow{2}{*}{\textbf{Method}} &  \multicolumn{2}{c}{\textbf{ScanRefer}} &  \multicolumn{2}{c}{\textbf{Multi3DRefer}}    \\
     \cmidrule(lr){2-3} \cmidrule(lr){4-5} 
       & Acc@0.25 & Acc@0.5 & F1@0.25 & F1@0.5  \\
     \midrule
     \rowcolor{gray!15!white}\multicolumn{5}{c}{\textit{Task-specific Models}} \\
      ScanRefer~\citep{chen2020scanrefer}& 37.3 & 24.3 & – & – \\
     MVT~\citep{chen2021mvt} & 40.8 & 33.3 & – & – \\
     3DVG-Trans~\citep{zhao20213dvg} & 47.6 & 34.7 & – & 25.5 \\
     ViL3DRel~\citep{chen2022language} & 47.9 & 37.7 & – & – \\
     3DJCG~\citep{cai20223djcg} & 49.6 & 37.3 & – & 26.6 \\
      M3DRef-CLIP~\citep{zhang2023multi3drefer} & 51.9 & 44.7 & 42.8 & 38.4 \\
      \midrule
    \rowcolor{gray!15!white}\multicolumn{5}{c}{\textit{3D MLLMs}} \\
    3D-LLM~\citep{hong20233d} & 30.3 & – &  – & – \\
    Grounded 3D-LLM~\citep{chen2024grounded} & 47.9 & 44.1 & 45.2 & 40.6 \\
    Chat-Scene~\citep{huang2024chat} & 55.5 & 50.2 & 57.1 & 52.4 \\
    LLaVA-3D~\citep{zhu2025llava} & 50.1 & 42.7  & – & – \\
    GPT4Scene$\dagger$~\citep{qi2025gpt4scene} & 40.5 & 36.7 & 45.4 & 42.1 \\
    GPT4Scene-HD~\citep{qi2025gpt4scene} & 50.9 &  46.4 & 53.7 & 50.0  \\
    GPT4Scene-HDM~\citep{qi2025gpt4scene} & 62.6 & 57.0 & 64.5 & 59.8 \\
    
    Video-3D LLM~\cite{zheng2025video}   & 58.1  & 51.7  & 58.0   & 52.7 \\
    Ross3D~\cite{wang2025ross3d}  & 61.1 & 54.4  & 59.6   & 54.3         \\
\rowcolor{gray!15!white}
  SmartMage (ours)  & \textbf{65.9} &  \textbf{59.5}   &  \textbf{65.4}         &  \textbf{60.7}  \\ %
      \bottomrule
    \end{tabular}
    }
    \label{tab:appendix-scanrefer+multi3drefer}
    \vspace{-2em}
\end{table}

\begin{table*}[t!]
    \centering
    \small
    \caption{Full Evaluation of 3D visual grounding on ScanRefer~\citep{chen2020scanrefer}. The ``Unique'' subset contains samples in which the described object corresponds to exactly one unique instance within a given object category, whereas the ``Multiple'' subset includes ambiguous cases with multiple instances belonging to the same object category. The ``Overall'' category aggregates performance across both unique and multiple-instance subsets. Accuracy is measured using IoU thresholds of 0.25 and 0.5 between predicted and ground-truth bounding boxes.  $\dagger$ indicates using a low-resolution setting. The best performance is highlighted in \textbf{bold}.}
    \vspace{-0.5em}
    \begin{tabular}{ccccccccccc}
    \toprule
     \multirow{2}{*}{\textbf{Method}} &  \multicolumn{2}{c}{\textbf{Unique}} &  \multicolumn{2}{c}{\textbf{Multiple}}  & \multicolumn{2}{c}{\textbf{Overall}}   \\
     \cmidrule(lr){2-3} \cmidrule(lr){4-5} \cmidrule(lr){6-7} 
       & Acc@0.25 & Acc@0.5 & Acc@0.25 & Acc@0.5 & Acc@0.25 & Acc@0.5  \\
     \midrule
    \rowcolor{gray!15!white}\multicolumn{7}{c}{\textit{Task-specific Models}} \\
    ScanRefer~\citep{chen2020scanrefer} & 76.3 & 53.5 & 32.7 & 21.1 & 41.2 & 27.4 \\
    TGNN~\citep{huang2021text} & 68.6  & 56.8 & 29.8 & 23.2 & 37.4 & 29.7 \\
    X-Trans2Cap~\citep{yuan2022x} & 73.2 & 50.8 & 37.6 & 25.2 & 44.5 & 30.1 \\
    InstanceRefer~\citep{yuan2021instancerefer} &  75.7 & 64.7 & 29.4 & 23.0 & 38.4 & 31.1 \\
    3DVG-Trans~\citep{zhao20213dvg} & 81.9 & 60.6 & 39.3 & 28.4 & 47.6 & 34.7 \\
    MVT~\citep{chen2021mvt} & 77.7 & 66.4 & 31.9 & 25.3 & 40.8 & 33.3 \\
    3D-SPS~\citep{luo20223d} & 84.1 & 66.7 & 40.3 & 29.8 & 48.8 & 37.0 \\
    ViL3DRel~\citep{chen2022language} & 81.6 & 68.6 & 40.3 & 30.7 & 47.9 & 37.7 \\
    3DJCG~\citep{cai20223djcg} & 83.5 & 64.3 & 41.4 & 30.8 & 49.6 & 37.3 \\
    D3Net~\citep{chen2021d3net} & – & 72.0 & – & 30.1 & – & 37.9 \\
    BUTD-DETR~\citep{jain2022bottom} &  84.2 & 66.3 & 46.6 & 35.1 & 52.2 & 39.8 \\
    HAM~\citep{chen2022ham} & 79.2 & 67.9 & 41.5 & 34.0 & 48.8 & 40.6 \\
    3DRP-Net~\citep{wang20233drp} & 83.1 & 67.7 & 42.1 & 32.0 & 50.1 & 38.9 \\
    3D-VLP~\citep{jin2023context} & 84.2 & 64.6 & 43.5 & 33.4 & 51.4 & 39.5 \\
    EDA~\citep{wu2023eda} & 85.8 & 68.6 & 49.1 & 37.6 & 54.6 & 42.3 \\
    M3DRef-CLIP~\citep{zhang2023multi3drefer} & 85.3 & 77.2 & 43.8 & 36.8 & 51.9 & 44.7 \\
    3D-VisTA~\citep{zhu20233d} & 81.6 & 75.1 & 43.7 & 39.1 & 50.6 & 45.8 \\
    ConcreteNet~\citep{unal2024four} & 86.4 & 82.1 & 42.4 & 38.4 & 50.6 & 46.5 \\
    \midrule

    \rowcolor{gray!15!white}\multicolumn{7}{c}{\textit{3D MLLMs}} \\
    Chat-Scene~\citep{huang2024chat} & 89.6 & 82.5 & 47.8 & 42.9 & 55.5 & 50.2 \\
    Video-3D-LLM~\citep{zheng2025video} & 88.0 & 78.3 & 50.9 & 45.3 & 58.1 & 51.7 \\
    GPT4Scene$^{\dagger}$~\citep{qi2025gpt4scene} & 65.5 & 61.2 & 34.8 & 31.1 & 40.5 & 36.7 \\
    GPT4Scene-HD~\citep{qi2025gpt4scene} & 77.5 & 71.9 & 44.9 & 40.6 & 50.9 & 46.4 \\
    GPT4Scene-HDM~\citep{qi2025gpt4scene} & \textbf{90.3} & 83.7 & 56.4 & 50.9 & 62.6 & 57.0 \\
    Ross3D~\cite{wang2025ross3d} & 87.2 & 77.4 & 54.8 & 48.9 & 61.1 & 54.4 \\
    \rowcolor{gray!15!white}
    SmartMage (ours)  &  89.6 &  \textbf{84.5}  &\textbf{56.7} & \textbf{53.5} &  \textbf{65.9}& \textbf{59.5} \\ 
      \bottomrule
    \end{tabular}
    \label{tab:appendix-scanrefer-detail}
\end{table*}

\begin{table*}[h!]
    \centering
    \caption{Full evaluation results on Multi3DRefer~\citep{zhang2023multi3drefer} for 3D visual grounding. We report results for five scenarios: Zero Target without Distractors (ZT w/o D), Zero Target with Distractors (ZT w/ D), Single Target without Distractors (ST w/o D), Single Target with Distractors (ST w/ D), and Multi-Target (MT). Performance is measured by F1 at IoU thresholds of 0.25 and 0.5 (F1@0.25, F1@0.5), and ``ALL'' denotes the overall results across all scenarios. $\dagger$ indicates that high-resolution settings are not used. The best results are highlighted in \textbf{bold}.
    }  
    \vspace{-0.5em}
    \resizebox{\textwidth}{!}{
    \begin{tabular}{ccccccccccc}
    \toprule
     \multirow{2}{*}{\textbf{Method}} &  \textbf{ZT w/o D} & \textbf{ZT w/ D}  & \multicolumn{2}{c}{\textbf{ST w/o D}}   & \multicolumn{2}{c}{\textbf{ST w/ D}} & \multicolumn{2}{c}{\textbf{MT}}  & \multicolumn{2}{c}{\textbf{ALL}}  \\
     \cmidrule(lr){2-2} \cmidrule(lr){3-3} \cmidrule(lr){4-5} \cmidrule(lr){6-7}  \cmidrule(lr){8-9}  \cmidrule(lr){10-11} 
       & F1 & F1 & F1@0.25 & F1@0.5 & F1@0.25 & F1@0.5 & F1@0.25 & F1@0.5 & F1@0.25 & F1@0.5 \\
     \midrule
    \rowcolor{gray!15!white}\multicolumn{11}{c}{\textit{Task-specific Models}} \\
    3DVG-Trans~\citep{zhao20213dvg} & 87.1 & 45.8 & – & 27.5 &  – & 16.7 & – & 26.5 & – & 25.5 \\
    D3Net (Grounding)~\citep{chen2021d3net} &  81.6 & 32.5 & – & 38.6 & – & 23.3 & – & 35.0 &  – &  32.2 \\
    3DJCG (Grounding)~\citep{cai20223djcg} & 94.1 & 66.9 & – & 26.0 & – & 16.7 & – & 26.2 & – & 26.6 \\
    M3DRef-CLIP~\citep{zhang2023multi3drefer} & 81.8 & 39.4 & 53.5 & 47.8 & 34.6 & 30.6 & 43.6 &  37.9 &  42.8 &  38.4 \\
    \midrule
   \rowcolor{gray!15!white}\multicolumn{11}{c}{\textit{3D MLLMs}} \\
    Chat-Scene~\citep{huang2024chat} & 90.3 & 62.6 & 82.9 & 75.9 & 49.1 & 44.5 & 45.7 & 41.1 & 57.1 & 52.4 \\
    GPT4Scene$^{\dagger}$~\citep{qi2025gpt4scene} & 85.2 & 61.4 & 60.1 & 55.1 & 37.7 & 34.4 & 39.4 & 36.3 & 45.4 & 42.1 \\
    GPT4Scene-HD~\citep{qi2025gpt4scene} & 93.6 & 81.8 & 72.5&  66.2 & 46.6 &  42.9 &  41.8 & 38.9 & 53.7 &  50.0 \\
    GPT4Scene-HDM~\citep{qi2025gpt4scene} &  \textbf{97.4} & 84.4 &  \textbf{85.0} & \textbf{77.7} &  59.9 & 55.1 & 48.6 &  44.6 & 64.5 & 59.8 \\
    Video-3D LLM~\cite{zheng2025video} &  94.1 & 76.7 & 81.2 & 72.6 & 52.7 & 47.4 & 40.6 & 35.3 & 57.9 & 52.4 \\
    Ross3D~\cite{wang2025ross3d} & 93.6 & 77.8 & 80.2 & 72.1 & 54.7 & 49.6 & 44.3 & 39.1 & 59.6 & 54.3 \\
    \rowcolor{gray!15!white}
    SmartMage (ours) &   96.8 & \textbf{85.7} & 84.9 & 77.3 & \textbf{60.0} & \textbf{55.1} & \textbf{51.4} & \textbf{47.7} &  \textbf{65.4} &  \textbf{60.7} \\
      \bottomrule
    \end{tabular}
    }
    \label{tab:appendix-multi3drefer-detail}
\end{table*}

\begin{table*}[t!]
    \centering
    \small
    \caption{Evaluation results of 3D dense captioning on Scan2Cap~\citep{chen2021scan2cap}. BLEU-4, METEOR, ROUGE and CIDEr denote text similarity scores between the predicted answer and the ground-truth answer. Metrics are computed under IoU thresholds of 0.25 and 0.5 between the predicted and reference bounding boxes. $\dagger$ indicates that high-resolution settings are not used. The best performance is highlighted in \textbf{bold}.}
    \vspace{-0.5em}
    \begin{tabular}{ccccccccc}
    \toprule
     \multirow{2}{*}{\textbf{Method}} &  \multicolumn{4}{c}{\textbf{Scan2Cap} (IoU@0.25)}   & \multicolumn{4}{c}{\textbf{Scan2Cap} (IoU@0.5)} \\
     \cmidrule(lr){2-5} \cmidrule(lr){6-9}  
       & BLEU-4 & METEOR & ROUGE & CIDEr & BLEU-4 & METEOR & ROUGE & CIDEr \\
     \midrule
    \rowcolor{gray!15!white}\multicolumn{9}{c}{\textit{Task-specific Models}} \\
    Scan2Cap~\citep{chen2021scan2cap} & 34.2 & 26.3  & 55.3 & 56.8 & 22.4 & 21.4 & 43.5 & 35.2 \\
    3DJCG~\citep{cai20223djcg} & 40.2 & 27.7 & 59.2 & 64.7 &  31.5 & 24.3 & 51.8 &  47.7 \\
    3D-VLP~\citep{jin2023context} & 41.0 & 28.1 & 59.7 & 70.7 & 32.3 &  24.8 & 51.5  & 54.9 \\
    3D-VisTA~\citep{zhu20233d} & 36.5 & 28.4 & 57.6 &  71.0 & 34.0 &  26.8 &  54.3 & 61.6 \\
    Vote2Cap-DETR~\citep{chen2023end} & 39.3  & 28.3 & 59.3 & 71.5 & 34.5 & 26.2 & 54.4 & 61.8 \\
    X-Trans2Cap~\citep{yuan2022x} & 35.7 & 26.6 & 54.7 & 61.8 & 25.1 & 22.5 & 45.3 & 43.9 \\
    
      \midrule
     \rowcolor{gray!15!white}\multicolumn{9}{c}{\textit{3D MLLMs}} \\
    LEO~\citep{huang2024embodied} & - & - & - & - & 38.2 & 27.9 & 58.1  & 72.4\\
    LL3DA~\citep{chen2024ll3da} &  41.4 & 27.8 &  59.5 & 74.2 & 36.8 & 26.0 & 55.1 & 65.2 \\
    Chat-Scene~\citep{huang2024chat} & 38.2 & 29.0 & 60.6 & 81.9 & 36.3 & 28.0 & 58.1 & 77.1 \\
    LLaVA-3D~\citep{zhu2025llava} & - & - & - & - & 41.1 & 30.2 & 63.4 & 79.2\\
    GPT4Scene$^{\dagger}$~\citep{qi2025gpt4scene} & 36.3 & 26.5 & 57.6 & 63.8 & 34.2 & 25.6 & 55.2 & 60.6\\
    GPT4Scene-HD~\citep{qi2025gpt4scene} & 40.4 & 28.3 & 60.2 & 79.1 & 37.9 & 27.3 & 57.7 & 74.4 \\
    GPT4Scene-HDM~\citep{qi2025gpt4scene} &  43.1 & 29.3 & \textbf{61.9} & 91.7 & 40.6 & 28.2 & 59.3 & 86.3 \\
    Video-3D LLM~\cite{zheng2025video} & - & - & - & - & 40.1 & 28.4 & 61.6  &  80.0 \\
     Ross3D~\citep{wang2025ross3d} & - & - & - & - & 43.4 & 30.3 & 66.9 & 81.3 \\
     \rowcolor{gray!15!white}
    SmartMage (ours) & \textbf{44.4} & \textbf{30.9} & 60.7 & \textbf{93.8} &  \textbf{43.6} & \textbf{32.9} & \textbf{69.3} &  \textbf{88.7} \\

      \bottomrule
    \end{tabular}
    \label{tab:appendix-scan2cap}
\end{table*}

\vspace{0.3em}\noindent\textit{\textbf{3D Dense Captioning.}}
Table~\ref{tab:appendix-scan2cap} presents the evaluation results of 3D dense captioning on the Scan2Cap~\citep{chen2021scan2cap} benchmark, comparing our model against several SOTA methods. Performance is measured using widely adopted captioning metrics—BLEU-4, METEOR, ROUGE, and CIDEr—at IoU thresholds of 0.25 and 0.5, indicating the quality and spatial accuracy of generated captions.

SmartMage achieves superior performance compared to advanced methods, including task-specific models and recent 3D LLMs. Specifically, at IoU=0.25, it attains the competitive BLEU-4 (44.4), METEOR (30.9), and CIDEr (93.8) scores, indicating strong fluency, semantic alignment, and relevance of the captions. 
At a stricter threshold of IoU=0.5, our model also demonstrates leading performance with top results in BLEU-4 (43.6) and METEOR (32.9), highlighting its robustness in precise localization conditions. 
These improvements indicate that SmartMage not only generates coherent and informative textual descriptions but also maintains accurate spatial grounding of target objects. The consistent gains over prior models further highlight the benefits of adaptive modality selection in capturing fine-grained geometric details and scene semantics essential for high-quality 3D caption generation.



\begin{figure*}[t!]
    \centering
    \includegraphics[width=1.0\linewidth]{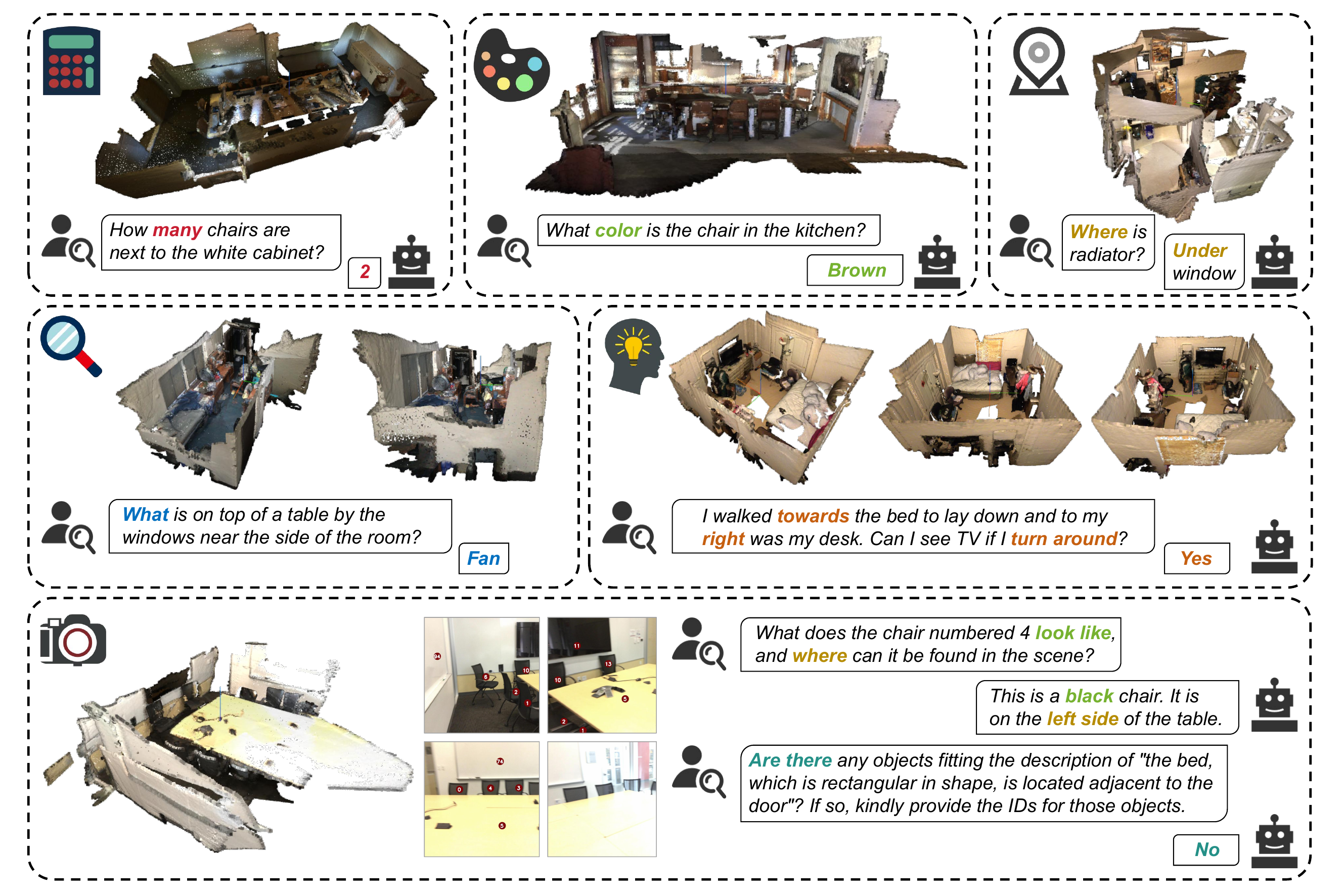}
    \vspace{-2.0em}
    \caption{Visualization of SmartMage performing diverse 3D scene understanding tasks on ScanFacet. Examples highlight the model's adaptive multimodal reasoning capabilities across different query types including counting, color recognition, object localization, spatial reasoning, and semantic identification.}
    \label{fig:chat}
    \vspace{-1em}
\end{figure*}

\begin{figure*}[t!]
    \centering
    \includegraphics[width=0.9\linewidth]{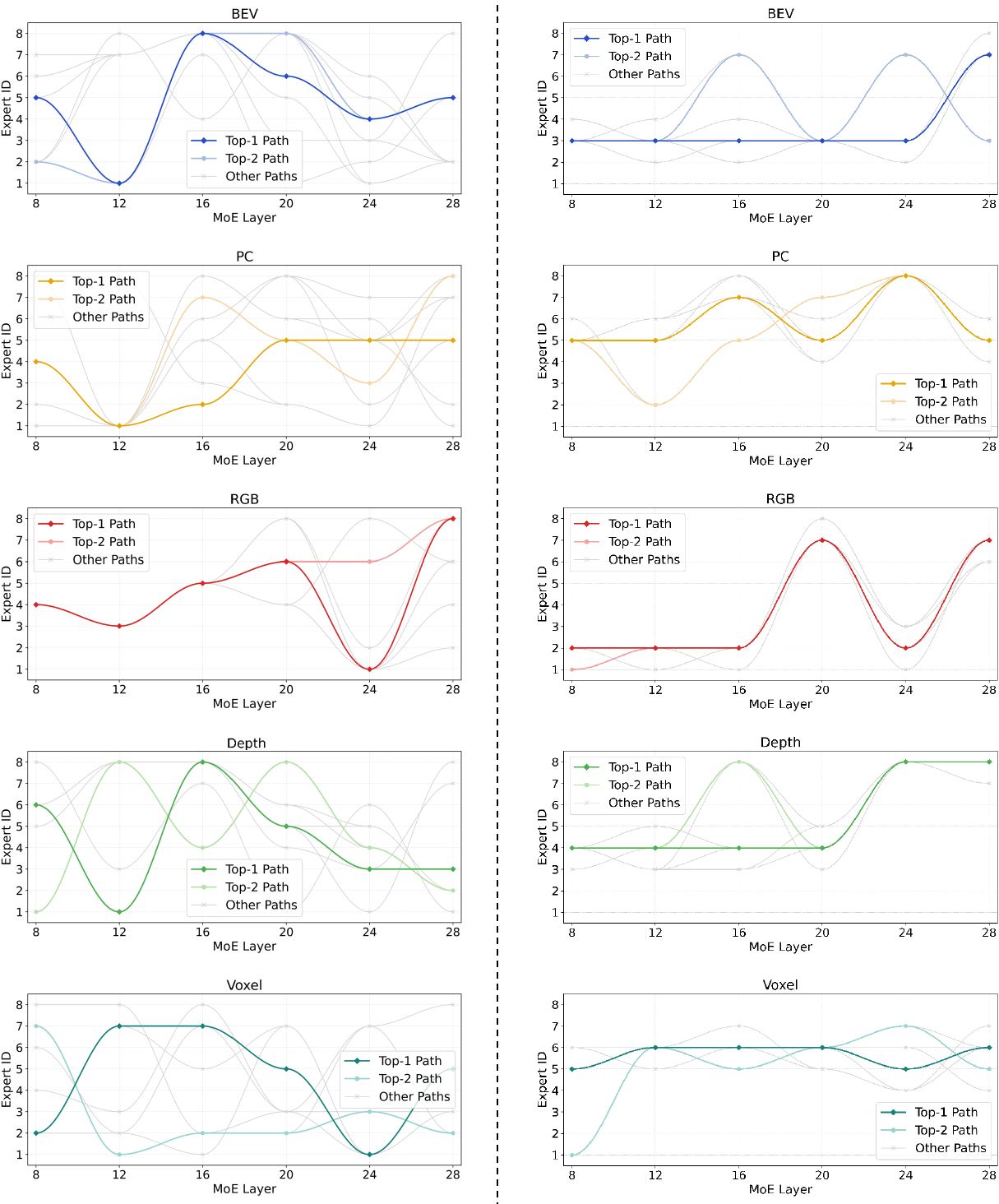}
    \vspace{-0.5em}
    \caption{Visualization of modality-specific routing trajectories across MoE layers.
The left column shows the top-10 activated routing pathways without MES, where different modalities exhibit diffuse and weakly-specialized expert usage.
The right column presents routing with the Modality-aware Expert Speculation (MES) module, which induces clearer modality-guided expert preferences and more stable, specialized activation patterns.
For each modality, colored curves denote its top-1 and top-2 routing paths, while gray curves represent the remaining candidate pathways.}
    \label{fig:suppl-path}
    \vspace{-1em}
\end{figure*}

\begin{table}[t!]
\centering
\caption{Exploring expert scaling in MoE layers. 
Here, \#Exp. denotes the total number of experts in each MoE layer. 
To enable training with larger expert counts under practical GPU memory limits, the hidden dimension of each expert is proportionally reduced (11008→1408). 
Results show that increasing the number of experts steadily improves performance in the low-to-mid range (4–20 experts), while the gains saturate or slightly regress when scaling to 30 experts.}
\label{tab:expert_scaling}
\vspace{-0.5em}
\resizebox{\linewidth}{!}{
\begin{tabular}{ccccccc}
\toprule
\multirow{2}{*}{\textbf{Setting}} & \multirow{2}{*}{\textbf{\#Exp.}} & \textbf{ScanQA} & \textbf{SQA3D} & \textbf{Scan2Cap} & \textbf{ScanRefer}   \\
\cmidrule(lr){3-3}
\cmidrule(lr){4-4}
\cmidrule(lr){5-5}
\cmidrule(lr){6-6}
& & EM & EM & BLEU-1@0.25 & Acc@0.25 \\ 
\midrule
1 & 4  & 24.4 & 51.6 & 70.1 & 31.7 \\
2 & 6  & 24.6 & 53.2 & 70.2 & 34.8 \\
3 & 8  & 24.7 & 53.3 & 71.8 & 37.4 \\
4 & 10 & 24.9 & 53.7 & 72.2 & 38.0 \\
5 & 15 & 25.3 & 54.3 & 72.8 & 38.5 \\
6 & 20 & 25.6 & 54.8 & 73.1 & 38.7 \\
7 & 30 & 25.9 & 54.9 & 72.9 & 38.2 \\
\bottomrule
\end{tabular}
}
\vspace{-1.0em}
\end{table}

\subsection{Additional Exploration of MoE}

\noindent\textit{\textbf{Exploring the Benefits of Expert Scaling.}}
Table~\ref{tab:expert_scaling} shows that performance generally improves as the number of experts increases from 4 to 20.
This trend suggests that a larger expert pool provides more flexible specialization, allowing the router to assign tokens to more suitable computation paths. 
We did not experiment with larger expert counts under the default setting due to GPU memory limitations.
To further investigate this trend, we reduced the expert hidden dimension from 11008 to 1408, allowing us to fit more experts within the same memory budget. Although this modification slightly lowers the absolute performance, it enables a cleaner analysis of the effect of expert number itself. In this way, we can better distinguish the benefit of scaling the number of experts from that of increasing per-expert capacity. 
As shown in Table~\ref{tab:expert_scaling}, performance improves noticeably in the low-expert regime (\(<15\)), but gains diminish in the mid-range (15--20). This suggests that the main benefit comes from introducing a moderate level of expert diversity, while excessively increasing the number of experts brings limited additional gains. 
At 30 experts, the gains saturate or slightly regress on some metrics, suggesting diminishing benefits from further increasing expert diversity.
Overall, these results indicate that effective MoE design should balance expert diversity and routing stability, rather than simply increasing the number of experts.

\vspace{0.3em}\noindent\textit{\textbf{Exploring MoE Layer Placement.}}
Table~\ref{tab:ab-moe-layer} presents the results of integrating the MoE module at different layers on the ScanQA~\citep{azuma2022scanqa} benchmark. We observe a clear performance improvement across all evaluation metrics when employing the MoE module at deeper layers. This trend suggests that higher transformer layers are more suitable for expert routing, since they operate on more semantically refined multimodal representations. In contrast, lower layers mainly capture generic perceptual patterns, where strong expert specialization may be less effective.
Specifically, incorporating MoE layers at depths [8,12,16,20,24,28] achieves the best results, significantly enhancing EM@1 accuracy from 29.8\% (without MoE) to 32.6\%. Similar improvements are evident across other metrics, such as CIDEr, which rises notably from 88.4 to 107.6. These results indicate that deeper MoE insertion improves not only answer correctness but also the quality of generated responses. This is likely because late-stage routing is better aligned with high-level reasoning and multimodal semantic fusion.
These results highlight that deeper integration of the MoE mechanism facilitates richer feature extraction, thereby enhancing the model's ability to accurately answer complex questions. 
These results highlight that placing MoE blocks at deeper stages leads to stronger performance, suggesting that expert routing benefits from semantically richer representations in later transformer layers.

\begin{table*}[t!]
    \centering
    \small
    \caption{Exploration of MoE layer placement on ScanQA~\cite{azuma2022scanqa}.
This table reports the performance of models with MoE layers inserted at different depths of the transformer.} 
    \vspace{-0.5em}
    \setlength{\tabcolsep}{10.0pt}{
    \begin{tabular}{lllllllllll}
        \toprule
        \textbf{Method} & \textbf{MoE layers} & \textbf{EM@1} & \textbf{EM-R@1} & \textbf{F1} &  \textbf{BLEU-1} & \textbf{BLEU-4} & \textbf{METEOR} & \textbf{ROUGE} & \textbf{CIDEr} \\
        \midrule
         w/o MoE & - & 29.8 & 45.1 & 45.5 & 41.9 & 13.9 & 17.1 & 43.8 &  88.4  \\ 
   
         w/ MoE & [0,2,4,6,8,10] & 30.7 & 45.3 & 46.2 & 41.5 & 14.1 & 17.6 & 44.0 & 90.9  \\
         w/ MoE & [0,4,8,12,16,20] & 31.7 & 47.1 & 47.8 & 42.2 & 15.7 & 17.9 & 46.0 & 104.3  \\

          w/  MoE & [8,12,16,20,24,28] & 32.6 & 49.0 & 48.8 & 43.7 & 17.5 & 19.0 & 47.1 & 107.6 \\
         \bottomrule
    \end{tabular}
    }
    \label{tab:ab-moe-layer}
    \vspace{-1.0em}
\end{table*}

\subsection{Qualitative Results on ScanFacet}\label{sec:suppl-results-visualization}
As illustrated in Fig.~\ref{fig:chat}, the proposed SmartMage model adeptly handles various categories of 3D scene understanding tasks, effectively demonstrating its versatility across distinct question types. For instance, it accurately identifies numeric details in counting tasks (e.g., ``How many chairs are next to the white cabinet?''), utilizes color recognition to specify attributes (e.g., the ``brown'' chair in the kitchen), and spatially localizes objects by contextual information (e.g., finding a radiator "under the window"). Moreover, the model is capable of interpreting spatial orientation and viewpoint-dependent questions, successfully answering queries related to turning around to view specific objects. Conversely, it can clearly recognize when queried objects or conditions are absent in the scene, indicating robust negative reasoning capability. These diverse examples underscore SmartMage's effectiveness in dynamically leveraging multimodal data representations, showcasing its capability for nuanced and contextually adaptive responses.

\subsection{Routing Trajectory Visualization}\label{sec:suppl-moe-path}
As shown in Fig.~\ref{fig:suppl-path}, we visualize the top-10 activated routing trajectories for each modality across the Mixture-of-Experts (MoE) layers, comparing the baseline routing without the Modality-aware Expert Speculation (MES) module (left) and the MES-enhanced routing (right). Each modality is visualized with a distinct color, with its Top-1 and Top-2 routing trajectories highlighted, while the remaining paths are shown in gray.

Without MES, routing behaviors are diffuse and lack clear specialization, with experts activated inconsistently across layers. In contrast, incorporating MES yields clearer and more modality-aligned routing patterns. Specifically, RGB tokens exhibit strong activations at $\mathcal{E}_2$ and $\mathcal{E}_6$, while still maintaining moderate routing to other experts. BEV tokens primarily rely on $\mathcal{E}_3$ and $\mathcal{E}_7$, showing more focused expert engagement. 
Point cloud (PC) tokens show dominant routing to $\mathcal{E}_5$, while voxel tokens favor $\mathcal{E}_6$.

These MES-induced routing trajectories exhibit clearer specialization, reduced dispersion, and more consistent modality-specific expert preferences. This structured behavior highlights MES's effectiveness in guiding heterogeneous modalities toward the most suitable experts across representation depths, thereby improving routing robustness and multimodal representation quality.

\begin{figure*}[t!]
    \centering
    \includegraphics[width=\linewidth]{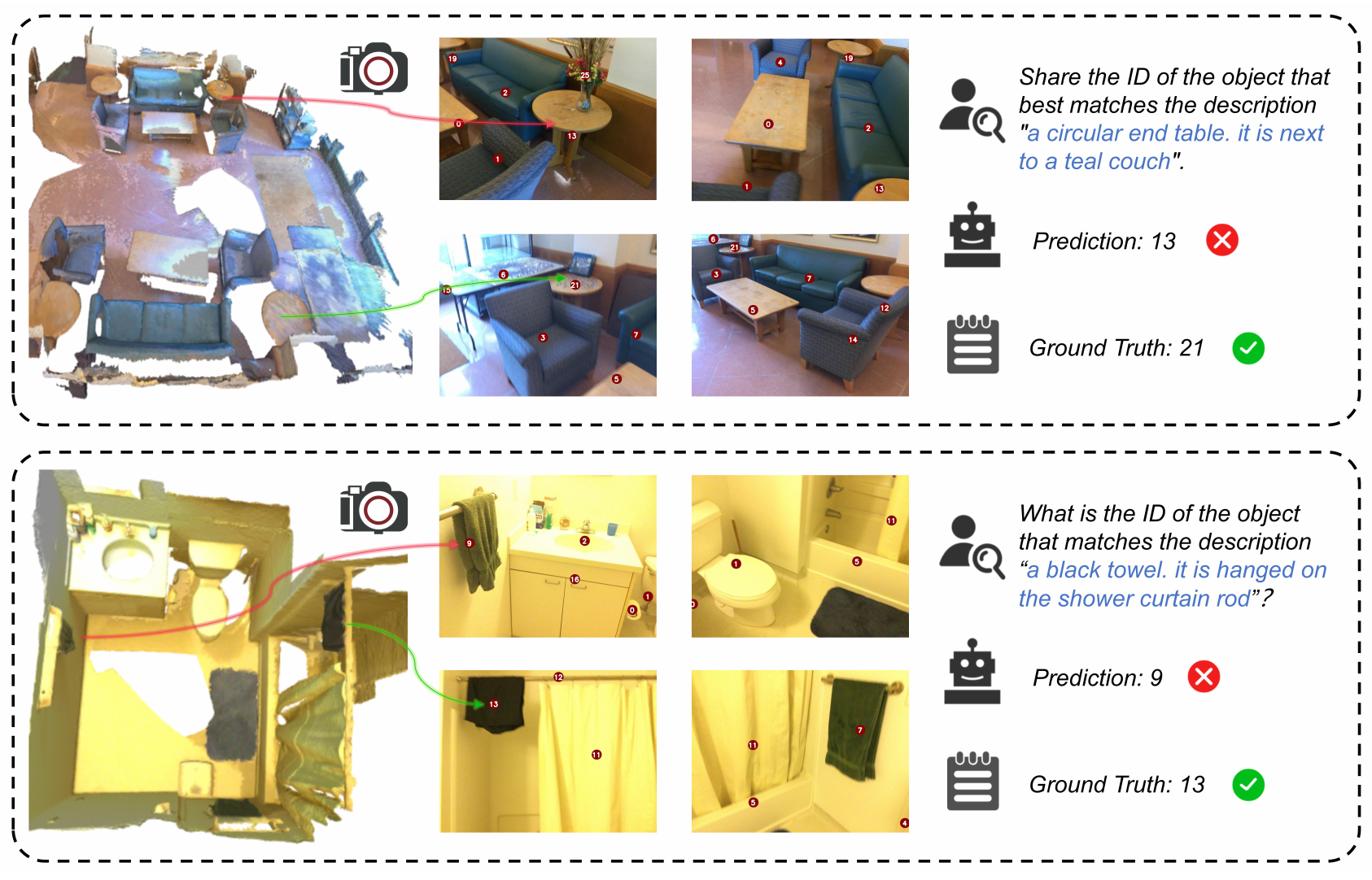}
    \vspace{-2.5em}
    \caption{Failure cases of SmartMage. Top: an annotation-ambiguity example where the model selects object 13 for the description 
“a circular end table next to a teal couch,” while the ground truth is object 21—both objects 
reasonably match the query. 
Bottom: a perception-related failure where lighting and color inconsistencies across views lead 
the model to misidentify the black towel, predicting object 9 instead of the correct object 13. 
These cases illustrate two key challenges: ambiguous annotations and sensitivity to visual 
appearance variations.}
    \label{fig:suppl-failure}
    \vspace{-1em}
\end{figure*}

\subsection{Failure Case Analysis}\label{sec:suppl-results-failure}
Fig.~\ref{fig:suppl-failure} illustrates the failure cases of our method.
In the first example, given the query ``a circular end table, it is next to a teal couch'' the model predicts object 13, while the ground truth is object 21.
Notably, object 13 also appears to satisfy the description, as it is similarly positioned next to a teal couch and matches the specified shape.
This indicates that the error arises primarily from inherent annotation ambiguity, rather than from a fundamental shortcoming of the model’s visual grounding capability. 
In the second example, for the query ``a black towel, it is hung on the shower curtain rod'' the model incorrectly selects object 9 instead of the correct object 13. 
This failure may be attributed to inconsistent lighting conditions between this scene and others, resulting in color deviations in multi-view RGB images, thus impairing the model's ability to accurately interpret visual cues and distinguish subtle color differences.
Additionally, the incorrect prediction might stem from the higher occurrence frequency of object 9 across multiple frames, potentially biasing the model’s attention toward it over the less prominently featured yet correct object 13. 
These cases highlight the importance of addressing both annotation ambiguity and robustness to visual variations in future model improvements.

\section{Limitations and Broader Impacts}\label{sec:suppl-limitation}
\noindent\textit{\textbf{Limitations.}}
Despite achieving promising results across various tasks, SmartMage still exhibits several limitations. 
First, the token budget constraints of large language models necessitate strict control over modality-specific inputs. 
To this end, multi-view images are selected using the fast keyframe selection algorithm \texttt{FoVSR}. 
While effective, this method may overlook critical viewpoints, leading to an incomplete spatial context.
Similarly, point clouds are downsampled using Farthest Point Sampling (FPS), reducing point density and limiting the representation of fine-grained object details—particularly for small-scale structures. 
These input reductions might degrade model performance, especially when other modalities fail to provide sufficient complementary information.
Second, the model’s effectiveness is partially constrained by the quality of the training dataset. Blurry multi-view images and annotation inaccuracies introduce noise and ambiguity, which can hinder performance in tasks that demand precise spatial understanding and accurate object localization.

\noindent\textit{\textbf{Broader Impacts.}}
This work advances 3D perception in vision--language models through adaptive and semantically grounded multimodal reasoning.
By dynamically selecting informative visual and geometric modalities, SmartMage may improve spatial understanding in applications such as assistive human--computer interaction, embodied AI, indoor robotics, and intelligent navigation.
More reliable reasoning over object attributes, spatial relations, and scene layouts could support context-aware interfaces, safer navigation, and more effective task execution in complex indoor environments.
The adaptive use of heterogeneous modalities may also reduce unnecessary reliance on redundant inputs and encourage more efficient multimodal reasoning.

At the same time, stronger 3D scene understanding introduces potential risks.
Detailed spatial perception and multimodal reasoning could be misused in privacy-sensitive scenarios, including surveillance or military applications.
In embodied systems, inaccurate modality selection or erroneous spatial reasoning may also lead to inappropriate actions when predictions are deployed without sufficient verification.
Moreover, model behavior can still be affected by imperfect visual observations, geometric reconstruction errors, and noisy annotations, which may propagate to downstream decisions.
These considerations call for responsible development, transparent reporting of model limitations, and careful human oversight when deploying such systems in safety- or privacy-sensitive environments.
Overall, we believe the potential benefits of adaptive 3D multimodal understanding outweigh the associated risks when the technology is developed and deployed responsibly.

\stopcontents[app]

\end{document}